\pdfoutput=1
\documentclass{article}

\usepackage{preprint,times}
\finalcopy   %

\usepackage{booktabs}
\usepackage{comment}
\usepackage{tabularx}
\usepackage{multirow}
\usepackage{subcaption}
\usepackage{amsmath,amssymb}
\usepackage{graphicx}
\usepackage{wrapfig}
\usepackage{needspace}                         
\usepackage{url}
\usepackage{xcolor}
\usepackage{colortbl}                          
\usepackage{float}                             
\definecolor{rowband}{gray}{0.945}             

\usepackage{xspace}
\usepackage{hyperref}                          %
\newcommand{\methodName}{BudgetPix\xspace}   

\usepackage{amsmath,amsfonts,bm}

\def\eqref#1{equation~\ref{#1}}

\def\1{\bm{1}}

\DeclareMathAlphabet{\mathsfit}{\encodingdefault}{\sfdefault}{m}{sl}
\SetMathAlphabet{\mathsfit}{bold}{\encodingdefault}{\sfdefault}{bx}{n}

\usepackage{paralist}

\title{\parbox{\textwidth}{\centering \methodName{}:~Compute-Adaptive Tokenization for Pixel-Space Image Diffusion}}

\author{%
\parbox[t]{\dimexpr\textwidth-2\tabcolsep\relax}{\centering
Ozgur Kara$^{1,\dagger}$\quad
Yujia Chen$^{2}$\quad
Daniel Watson$^{2}$\quad
David Forsyth$^{1}$ \\[2pt]
James Matthew Rehg$^{1}$\quad
Wen-Sheng Chu$^{2,\ddagger}$\quad
Du Tran$^{2,\ddagger}$ \\[6pt]
{\normalfont\normalsize $^{1}$University of Illinois Urbana-Champaign \quad
$^{2}$Google}
}%
}

\makeatletter
\renewcommand\paragraph{\@startsection{paragraph}{4}{\z@}{0pt}{-1em}{\normalsize\bf}}
\makeatother
\begin{document}
\maketitle
\lhead{}\renewcommand{\headrulewidth}{0pt}   
{\renewcommand{\thefootnote}{\fnsymbol{footnote}}
\footnotetext[2]{Work done during an internship at Google.}
\footnotetext[3]{Joint last authors.}
}
\addtocontents{toc}{\protect\setcounter{tocdepth}{-1}}

\begin{figure}[h!]
\centering
\input{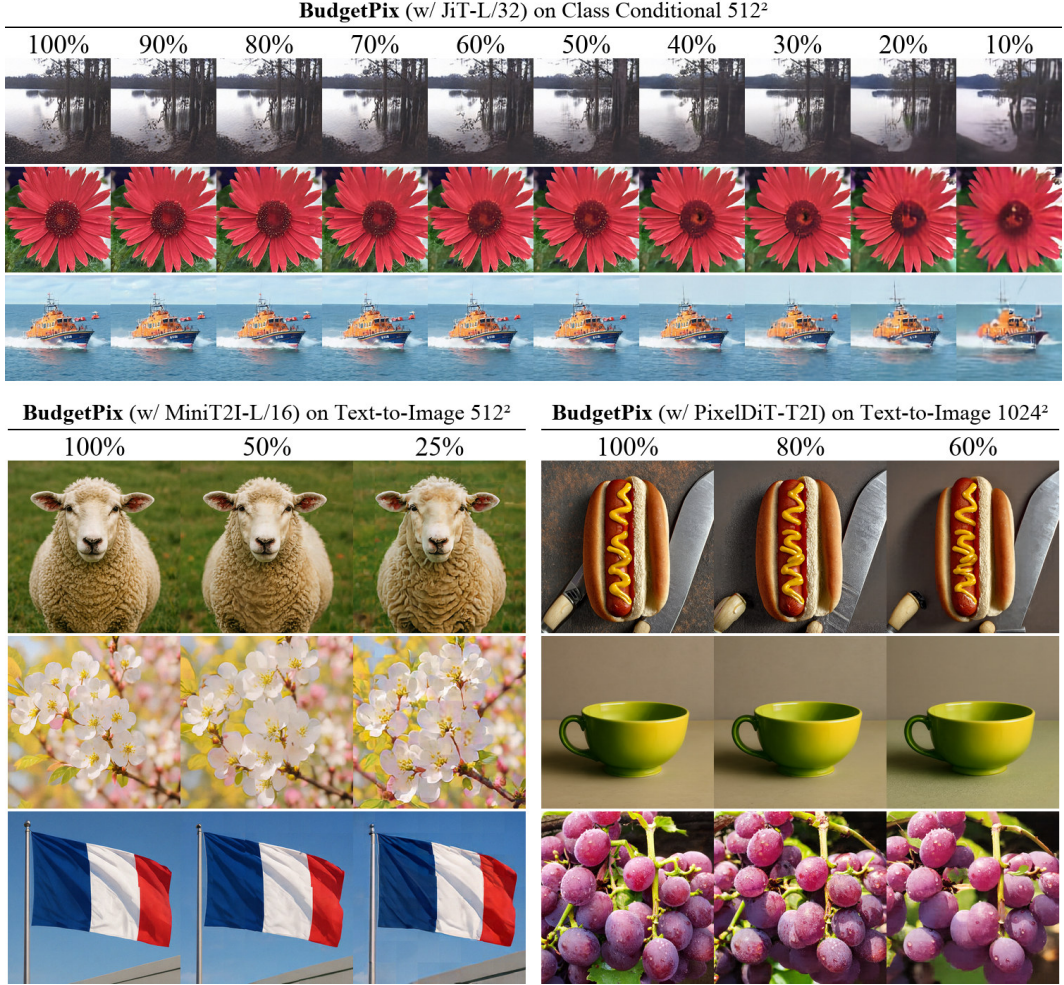}%
\vspace{-18pt}   
\caption{\textbf{Test-time compute adjustment.} \methodName{} dynamically scales inference costs, achieving a strong quality-efficiency tradeoff across JiT (top), MiniT2I (bottom left), PixelDiT (bottom right), 3 popular architectures for pixel-space image diffusion. Percentages show the fraction of tokens used.}
\label{fig:teaser}
\end{figure}

\begin{abstract}
Most image generation models rely on uniform tokenization, allocating the exact same computational budget to equally-sized image patches. This static paradigm cannot adapt to different resource constraints at inference time, and yields suboptimal quality-cost tradeoff by devoting the same effort to both plain backgrounds and intricate details. We propose \methodName{}, an adaptive tokenization framework that dynamically allocates compute based on visual complexity and spatial layout, enabling flexible computational budgeting at inference time. \methodName{} comprises three key components: (1) an adaptive encoder that maps a fixed-size image to a variable-length token sequence using an entropy-guided quadtree alongside a multi-scale patch embedder; (2) a scale-aware decoder reconstructs fixed-resolution images from multi-scale token sets; and (3) a flexible training and sampling schedule that enables pixel-space denoisers to operate across variable token counts. \methodName{} seamlessly integrates with existing pixel-space diffusion architectures, enabling a single checkpoint to be operated at a wide range of compute budgets. Evaluated on text-to-image generation, \methodName{} matches the fidelity of MiniT2I-L at $512^2$ 
and PixelDiT at $1024^2$ 
using just 25\% of the original compute budget. In class-conditional using a MeanFlow backbone, \methodName{} requires merely 60\% of the full compute budget to produce images with near-zero quality degradation, observing a marginal 0.8-point increase in FID. 
Comprehensive assessments by human and VLM judges confirm that \methodName{} establishes a significantly improved quality-efficiency tradeoff over prior budget-adaptive baselines. More details are available at our project page: \url{https://karaozgur.com/BudgetPix}
\end{abstract}

\begin{figure}[b!]
    \centering
    \includegraphics[width=\linewidth]{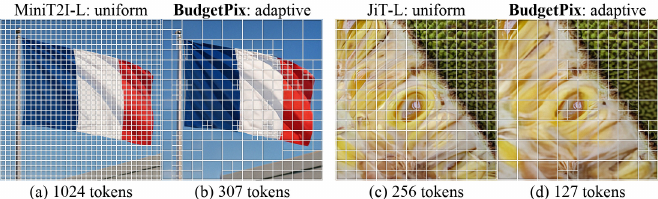}
    \vspace{-18pt}
    \caption{\textbf{Uniform vs.\ content-adaptive tokenization.} For a $512\times512$ image, uniform tokenization requires 1,024 tokens with $16\times16$ patches (a) and 256 tokens with $32\times32$ patches (c). \methodName{} adapts to visual content, requiring only 307 (b) and 127 (d) tokens. Baselines in (a) and (c) represent full-budget MiniT2I-L and JiT-L; \methodName{} in (b) and (d) operates at 30\% and 50\% budgets, respectively.}
    \label{fig:motivation}
\end{figure}

\section{Introduction}
\label{sec:intro}

Diffusion Transformers (DiTs) \citep{peebles2023scalable} have established a new standard for image generation. While most architectures operate in the latent space of a pretrained autoencoder \citep{rombach2022high}, emerging pixel-space DiTs \citep{li2026back,yu2026pixeldit} process raw pixels directly, bypassing the reconstruction ceiling and eliminating the reliance on a separate tokenizer. However, this architectural purity introduces a severe computational bottleneck: the uniform tokenization. Because self-attention scales quadratically with the number of tokens, pixel-space models incur massive computational costs. Furthermore, because this grid resolution is hardcoded during training, the inference cost becomes rigidly fixed. A deployed model cannot scale down to accommodate resource-constrained edge devices, nor can it dynamically adjust to varying user latency requirements, without undergoing entirely separate retraining.


This static uniform tokenization is fundamentally misaligned with how visual data is structured. 
Information density varies significantly within images: a cloudless sky offers minimal visual entropy, while heavily textured objects demand the capture of high-frequency details. Forcing a uniform grid to blindly spend identical compute across both regions creates severe computational waste. As shown in Figure~\ref{fig:motivation}a, standard MiniT2I \citep{minit2i2026} wastes 1,024 tokens rendering a simple flag against an open sky, whereas our content-adaptive framework reconstructs the scene perfectly with just 307 tokens (Figure~\ref{fig:motivation}b). Even for complex, edge-to-edge textures like a jackfruit--which strictly requires 256 tokens in standard JiT \citep{li2026back} (Figure~\ref{fig:motivation}c)--\methodName{} matches the baseline visual quality while cutting the required token budget by 50\% (Figure~\ref{fig:motivation}d).

Content-adaptive tokenization is well established in discriminative vision models via token merging \citep{bolya2023token}, token pruning \citep{rao2021dynamicvit}, or dynamic multi-scale representations \citep{beyer2023flexivit,dehghani2023patch,kim2025sapiensid,choudhury2026faster}.
However, extending this paradigm to generative models introduces a fundamental challenge. While discriminative models dynamically allocate tokens over an \emph{already-provided} input image to predict a class label \citep{choudhury2026faster}, generative denoisers must dynamically allocate compute over an image that is \emph{actively being generated}. This creates a chicken-and-egg dilemma: the denoiser must gauge the local spatial complexity of the image before the image content itself has fully materialized.


In this work, we propose \methodName{}, a framework that adapts any pretrained pixel-space diffusion transformer into a compute-adaptive image generator. \methodName{} introduces three novel components. First, an entropy-driven encoder maps images into variable-length sequences of multi-scale tokens. Second, A complementary decoder reconstructs these variable-length token sequences into predetermined, fixed-resolution images. Finally, a training and inference strategy, coupled with content-layout augmentation, enables standard checkpoints--originally trained on fixed token budgets--to become dynamically adjustable during inference. Our extensive experiments (Section~\ref{sec:analysis}) demonstrate that \methodName{} can be integrated seamlessly with established pixel-space architectures--including JiT \citep{li2026back}, MiniT2I \citep{minit2i2026}, PixelDiT \citep{yu2026pixeldit}, and MeanFlow \citep{geng2026mean,lu2026onestep}--and provide a significantly-better quality-efficiency tradeoff (see plots in Section~\ref{sec:results}).

We evaluate \methodName{} on class-conditional and text-to-image pixel-space diffusion models from $256^2$ to $1024^2$ resolution. Under strict low-compute budgets where prior budget-adaptive methods undergo catastrophic collapse, \methodName{} degrades gracefully, preserving high perceptual quality. For class-conditional generation, \methodName{} synthesizes coherent, high-quality images using as little as 10\% of the original compute budget (Figure~\ref{fig:teaser}, top 3 rows)--a regime where competing methods fail entirely (cf. Figure~\ref{fig:qual_minit2i_l}). This robustness mirrors phenomena observed in lossy image compression, where visual data can be heavily down-sampled without significant perceptual degradation.


Quantitatively, for text-to-image generation, \methodName{} matches the fidelity of full-compute baselines while utilizing only 25\% of the original token budget, achieving GenEval scores of $0.874$ (vs. $0.882$) on MiniT2I-L at $512^2$, and $0.725$ (vs. $0.721$) on PixelDiT at $1024^2$. For class-conditional generation, integrating \methodName{} into a MeanFlow \citep{lu2026onestep} backbone allows the model to operate at merely 60\% of the original compute budget with negligible degradation in visual quality, observing a minimal FID increase of only $0.81$ points (from $2.65$ to $3.46$). Finally, comprehensive evaluations by both human and Vision-Language Model (VLM) judges confirm that \methodName{} achieves a better quality-efficiency tradeoff compared to existing budget-adaptive methods.





Our primary contributions are:
\begin{compactitem}
\item  We propose \methodName{}, a novel compute-adaptive tokenization framework for pixel-space image diffusion, designed to integrate smoothly into diverse pixel-based diffusion architectures such as JiT, MiniT2I, PixelDiT, and MeanFlow.

\item \methodName{} introduces three novel components: (1) an entropy-based encoder mapping images into variable-length sequences of multi-scale tokens; (2) a specialized decoder reconstructing fixed-resolution images from these tokens; and (3) a flexible training and sampling schedule supporting denoisers to operate across continuously variable token counts.

\item Comprehensive experiments show that \methodName{} achieves an effective quality-efficiency tradeoff compared with alternative adaptive denoising methods.
\end{compactitem}

\section{Related work}

\paragraph{Pixel-space diffusion models.} For several years the cost of resolution was handled by
not paying it: latent diffusion generates inside a pretrained autoencoder
\citep{rombach2022high}, at the price of a reconstruction ceiling and a second training stage.
To manage this, early pixel-space models relied on U-Nets that omitted attention except at the lowest resolutions, before advancing to cascaded models \citep{ho2022cascaded}, single models with resolution-shifting noise schedules \citep{hoogeboom2023simple,hoogeboom2025simpler}, and finally, hierarchical transformers \citep{pmlr-v235-crowson24a,chen2025pixelflow}. The most recent line drops the
hierarchy altogether: JiT \citep{li2026back} and PixelDiT \citep{yu2026pixeldit} denoise large patches
with a plain transformer and a narrow patch embedding, matching latent-model quality at the same sequence length.
While this approach offers simplicity, it relies on a single uniform patch grid where size rigidly fixes the computational cost of every step. It also introduces a known weakness in decoding large patches, which PixelDiT addresses
with a second pixel-level stage and PixNerd resolves with a per-patch neural field \citep{wang2026pixnerd}.
We adapt a released model of exactly same architecture with AdaLN conditioning \citep{peebles2023scalable} trained via flow matching
\citep{lipman2023flow} settings and classifier-free guidance \citep{ho2021classifierfree} utilized during inference .

\paragraph{Content-adaptive tokenization.} Non-uniform, layout-based token reduction first emerged for discriminative vision tasks within pretrained transformers
\citep{bolya2023token,rao2021dynamicvit,liang2022evit,yin2022vit,wang2021not}, and later as
tokenization proper, through flexible patch sizes and content-following tokenizers
\citep{beyer2023flexivit,dehghani2023patch,lew2024superpixel,ronen2023vision,havtorn2023msvit,kim2025sapiensid}. Adapting such strategies remains difficult for generation, because a denoiser must write an image \emph{through} the layout and no image exists when
the layout is chosen. Adaptive work here has therefore stayed off the input grid, varying what is
updated or cached along the trajectory \citep{liu2026region,bolya2023tokensd,wu2025importance,zou2025accelerating},
varying patch size uniformly over the image \citep{anagnostidis2025flexidit,li2026pyramid}, or
retraining a latent tokenizer \citep{shen2026cat}. 
Concurrent with us, RTI \citep{zamfir2026elastic} adapts a frozen pixel-space model using LoRA to consume region tokens. In contrast, our method, \methodName{}, plugs in our adaptive encoder and decoder, fully fine-tunes the entire checkpoint to accept variable-sized inputs starting from the first block, and derives the layout from the model's own estimate of the clean image rather than its internal features.

\section{\methodName{}}
\label{sec:method}

\subsection{Overview}
\label{sec:overview}
\begin{wrapfigure}{r}{0.57\linewidth}
    \vspace{-5.4\baselineskip}
    \centering
    \includegraphics[width=\linewidth]{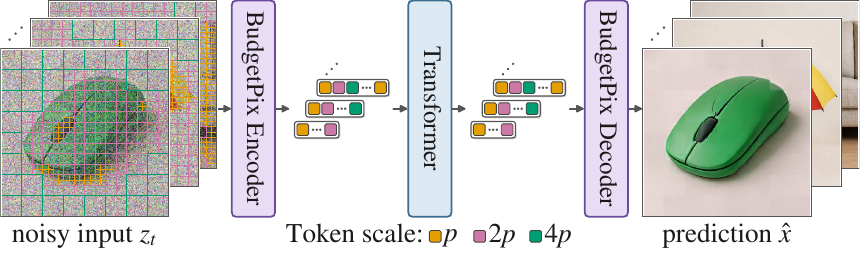}
    \vspace{-16pt}
\caption{\textbf{Overview.} 
\methodName{} enables compute-adaptive pixel-space diffusion with three components: an encoder, a decoder, and flexible training and inference across token counts. The Encoder and Decoder (purple) are trained from scratch; the Transformer (blue) is initialized from existing baselines.
\vspace{0.5em}
}



\label{fig:budgets}
\vspace{-1.0\baselineskip}
\end{wrapfigure}

\methodName{} is designed to adapt existing pixel-space diffusion transformers into compute-adaptive image generators. As shown in Figure \ref{fig:budgets}, it consists of three core components: an encoder that compresses images into variable-length, multi-scale token sequences (Sec.~\ref{sec:encoder}), a complementary decoder that reconstructs these token sequences back to fixed-resolution images (Sec.~\ref{sec:decoder}), and a unified training and inference strategy (Sec.~\ref{sec:traininf}).
By replacing the uniform tokenization of standard baselines with this content-adaptive pipeline, \methodName{} enables existing diffusion backbones to seamlessly process flexible token counts and adjust their computational budget based on visual complexity. 


\subsection{\methodName{} Encoder: Embed Non-Uniform Patches}
\label{sec:encoder}



{\bf Non-Uniform Layout Construction:}
The image layout determines the spatial partition of the image--specifically, which regions maintain the base $p\!\times\! p$ patch resolution and which are merged into larger patches to be encoded as a single token. Because visual complexity is spatially non-uniform, fine details concentrate on specific regions while homogeneous areas ({\em e.g.}, sky or a plain wall) contain minimal information. Inspired by APT \citep{choudhury2026faster}, we construct this non-uniform layout based on local intensity entropy. The layout is computed from the image during both training and inference. During training, it is derived directly from the ground-truth image $x$. At inference, because the clean image is unavailable, the model initially runs a few warm-up denoising steps on a uniform $p\times p$ grid to generate a predicted image $\hat{x}$, which is then used to estimate the content-adaptive layout.

\begin{figure}[t]
    \centering
    \includegraphics[width=\linewidth]{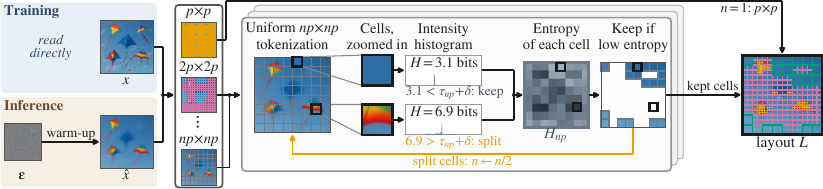}
    \vspace{-12pt}
    \caption{\textbf{Non-uniform layout construction.} Given an image (the original image $x$ during training, or the denoising image $\hat{x}$ during inference), the objective is to partition it into cells across multiple scales: $p\times p$, $2p\times 2p$, \dots, $np\times np$, where $n=2^j$ for $j \ge 1$. The process begins by dividing the image into $np\times np$ cells, which is the coarsest scale. For each cell, we construct a pixel intensity histogram and compute its entropy $H$. If $H < \tau_{np} + \delta$ (where $\tau_{np}$ is the scale-specific threshold and $\delta$ is a global offset), the cell remains a single token. Otherwise, it is subdivided into four cells of the next finer scale ($n \leftarrow n/2$). This recursive subdivision terminates at $n = 1$, corresponding to the minimum cell size of $p\times p$.}
    \label{fig:layout}
\end{figure}

Given $N$ different patch scales $s \!=\! 2^j p$, ${j \!\in\! \{0, 1, \dots, N-1\}}$, with $N\!-\!1$ the coarsest level, corresponding to the largest patch size. For each spatial cell $R$ at this coarsest level, we compute a $K$-bin histogram of its internal pixel intensities and calculate the corresponding Shannon entropy: $H(R) \!=\! -\sum_{k=1}^{K} q_k(R) \log_2 q_k(R)$, where $q_k(R)$ denotes the fraction of pixels falling into bin $k$. 
Token allocation starts from the coarsest level---a cell is retained as one token if $H(R)\le\tau_s+\delta$, where $\tau_s$ is a scale-specific threshold and $\delta$ a global offset; otherwise, it is recursively split into four quadrants and evaluated at the next finer scale. 
This recursive quadtree splitting continues to the finest $p \times p$ scale, ensuring the resulting multi-scale tokens perfectly tile the image (Figure~\ref{fig:layout}).

{\bf \methodName{} Encoder:}
The \methodName{} encoder receives non-uniform image patches defined by the entropy-guided quadtree and encodes them into a set of tokens with the same embedding size. Specifically, a patch $P_i$ at scale $s = np$ (where $n = 2^j$ and $j \in \{0, 1, \dots, N-1\}$) is mapped to a single token vector $\mathbf{x}_i$.

\begin{figure}[t]
\centering
\input{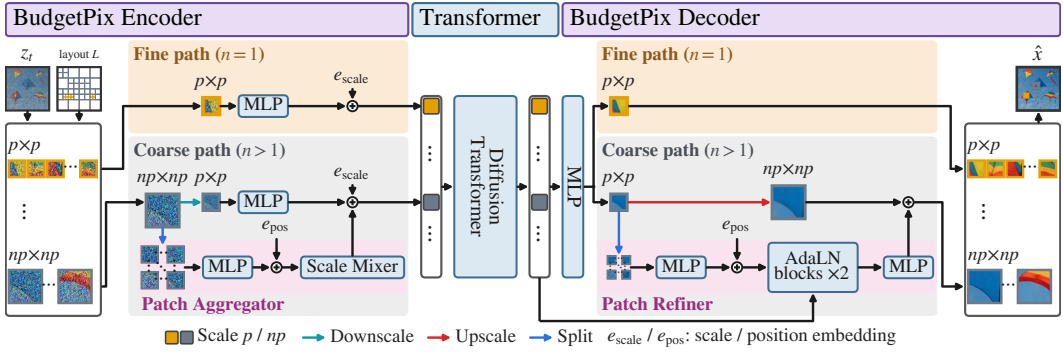}%
\vspace{-16pt}
\caption{\textbf{\methodName{} Architecture.} The encoder receives the noisy input $z_t$ and its corresponding spatial layout $L$, converts the image into a sequence of multi-scale patches. The finest $p \times p$ tokens route directly through the fine path, while coarser patches navigate a dedicated coarse path in both the encoder and decoder. In the encoder, each coarse patch undergoes simultaneous downscaling (to $p \times p$) and splitting via a Patch Aggregator (into $n^2$ sub-patches). These streams are processed through MLP layers and a scale mixer before being fused into a single coarse token. In the decoder, each denoised coarse patch is upscaled back to its target resolution ($np \times np$). Concurrently, a Patch Refiner splits the patch into $4 \times 4$ sub-patches, processes them through an MLP and a lightweight transformer, and fuses them with the upscaled patch to restore local texture. The central diffusion transformer is directly initialized from existing pixel-space models, modified only to operate across variable, dynamically determined token counts rather than a fixed count.}
\label{fig:model}
\end{figure}

To process these variable-sized patches, the encoder utilizes a dual-path architecture (Figure~\ref{fig:model}, left). The \emph{fine path} processes base patches ($n = 1$): each $p \times p$ patch is embedded directly via the pretrained patch embedder $E$ inherited from the base model. The \emph{coarse path} processes larger patches ($n \!>\! 1$). These patches are bilinearly downsampled to $p \times p$, denoted as $\mathrm{resize}_p(P_i)$, and subsequently embedded via $E$. While this main stream preserves macro-level semantics, it inherently discards high-frequency internal details. To recover this lost detail, a localized secondary stream, the \emph{Patch Aggregator} ($\mathrm{PA}_s$), is introduced. $\mathrm{PA}_s$ partitions $P_i$ into $n^2$ sub-patches, denoted as $\mathrm{split}(P_i)$. Each sub-patch is independently embedded with $E$, augmented with a learned relative positional encoding $\mathbf{e}_{\mathrm{pos}}$ to specify its spatial location within $P_i$, and finally fused into a single correction vector using a scale-specific mixer. This scale-mixer comprises a stack of convolutional layers followed by an MLP layer (architectural details are provided in supplementary). Both paths add a learned scale embedding $\mathbf{e}_{\mathrm{scale}}(s)$, so the token is
\begin{equation}
\mathbf{x}_i =
\begin{cases}
E(P_i) + \mathbf{e}_{\mathrm{scale}}(s), & n = 1,\\[2pt]
E\big(\mathrm{resize}_{p}(P_i)\big) + \mathbf{e}_{\mathrm{scale}}(s) + \mathrm{PA}_s\big(\mathrm{split}(P_i)\big), & n > 1.
\end{cases}
\label{eq:enc}
\end{equation}

\subsection{\methodName{} Decoder: Decoding Multi-scale Image Patches}
\label{sec:decoder}

The \methodName{} decoder receives the output token sequence from the central transformer--where each token corresponds to a specific patch in the non-uniform spatial layout--and reconstructs them into pixel-space patches at their designated scales. These reconstructed patches subsequently tile together to form the final predicted image $\hat{x}$. Specifically, the decoder maps the latent token $\mathbf{h}_i$, corresponding to a patch $P_i$ of scale $s = np$, back into an $np \times np$ pixel space patch $\hat{P}^{(s)}_i$.

To reconstruct tokens of varying scale back into pixel space, the decoder (Figure~\ref{fig:model}, right) mirrors the dual-path architecture of the encoder. The \emph{fine path} processes base tokens ($n \!=\! 1$): a pretrained decoding head maps the latent token $\mathbf{h}_i$ directly into a $p \times p$ pixel patch $\hat{P}_i \in \mathbb{R}^{p \times p \times C}$, where $C=3$ represents the RGB image channels. The \emph{coarse path} handles larger tokens ($n > 1$). Initially, the main stream decodes the token and bilinearly upsamples the output $\hat{P}_i$ to the target resolution of $np \times np$, denoted as $\mathrm{resize}_{np}(\hat{P}_i)$. While this main stream successfully recovers the region's low-frequency structural layout, the resulting patch is inherently over-smoothed, lacking internal high-frequency textures. To resolve this, a localized secondary stream--the \emph{Patch Refiner} ($\mathrm{PR}_s$)--generates and injects the missing high-frequency details as an additive residual:
\begin{equation}
    \hat{P}^{(s)}_i = \mathrm{resize}_{np}\big(\hat{P}_i\big) + \mathrm{PR}_s\big(\mathrm{split}(\hat{P}_i), \mathbf{h}_i\big),
    \label{eq:dec}
\end{equation}
where $\hat{P}_i$ is the base patch from the pretrained head; on the fine path the resize is the identity and the residual is absent. The refiner is a lightweight transformer that stays inside one coarse token: it embeds $4\times4$ sub-patches $\mathrm{split}(\hat{P}_i)$, adds local positional embeddings $\mathbf{e}_{\mathrm{pos}}$, processes the sequence with two self-attention layers under adaptive layer normalization (AdaLN) conditioned on $\mathbf{h}_i$, which brings the token's global context into the local details, and projects the result linearly to the $np \times np$ residual. Because it runs on short, local token sequences, its cost per token does not grow with image resolution, and it adds only ${\sim}1\%$ to the parameter count.

The central diffusion transformer subsequently adds its own position embeddings and evaluates the tokens as usual. Note that the position embeddings applied within the Patch Aggregator and Patch Refiner operate in a localized, patch-relative coordinate system, whereas the embeddings utilized by the central diffusion transformer operate in a global, image-relative coordinate system.

\subsection{Training and Inference}
\label{sec:traininf}


{\bf Training Objective:}\label{sec:loss}
Any pretrained pixel-space diffusion transformer can be integrated into the \methodName{} framework by encapsulating it with our trainable encoder and decoder modules. The entire pipeline is then fine-tuned utilizing the base model's flow-matching objective for full-image $x$-prediction. Let the noisy latent state be $z_t = t\,x + (1 - t)\,\boldsymbol{\epsilon}$, and the network prediction be $\hat{x} = f_\theta(z_t, t, y, L)$, where $t$ is the timestep, $y$ is the conditioning signal, and $L$ is the spatial layout. The fine-tuning loss is formalized as: $\mathcal{L}(\theta) = \mathbb{E}_{x, y, L, t, \boldsymbol{\epsilon}} \big[ w(t) \lVert x - \hat{x} \rVert_2^2 \big]$ where $w(t)$ represents the noise-schedule weighting derived from the base model.


{\bf Layout Augmentation:}\label{sec:layoutdist}
To ensure the transformer can reliably process sequences of varying lengths, we dynamically assign one of three distinct layout types to each image during training. Let $Q(\cdot, \tau + \delta)$ represent the entropy quadtree defined by thresholds $\tau_s + \delta$, and let $H(x)$ denote the true entropy map of the training image $x$. {\bf {\em Uniform layout}} (probability $a < 1$): The model uses $L_{\mathrm{uniform}}$, the standard dense grid of the base architecture. This anchors the training process and prevents catastrophic drift from the pretrained weights. {\bf {\em Random layout}} (probability $b$, where $a+b < 1$): The quadtree $Q(\tilde{H}, \tau + \delta)$ is applied to a synthetic entropy map $\tilde{H}$, where values are drawn uniformly between $0$ and $\log_2 K$ bits per cell. Crucially, this forces coarse tokens to randomly encompass highly detailed image regions. This simulates the constraints of low-budget inference, preventing the model from assuming that coarse tokens only represent flat backgrounds. {\bf {\em Content-adaptive layout}} (probability $1 - a - b$): The standard quadtree $Q(H(x), \tau + \delta)$ is applied using the actual image entropy map, optimally routing tokens based on true visual complexity. For both the random and content-adaptive quadtree layouts, the global offset $\delta$ is randomly sampled for each training image. This ensures the model learns to operate across a continuous range of token counts, scaling from roughly 25\% up to 100\% of the original dense grid.

{\bf Sampling with Dynamic Layout Estimation:}
During inference, the image's entropy is unknown before it is generated, so the layout is estimated as the image forms. Let $i$ index the sampling steps and $\hat{x}_i$ be the model's prediction of the clean image at step $i$. The first $n_w$ warm-up steps run on the uniform grid, $L_i = L_{\mathrm{uniform}}$, which lets the model set the global composition of the image. After that, the layout is the entropy quadtree of the current prediction, $L_i = Q\big(H(\hat{x}_{i'}), \tau + \delta^*(\hat{x}_{i'}, B)\big)$, with the thresholds shifted by the offset $\delta^*$ that budget mode finds for this prediction and the budget $B$. The layout is rebuilt only every $m$ steps: $i' = n_w + m\lfloor (i - n_w)/m \rfloor$ is the last rebuild step at or before $i$, so the $m$ steps from $i'$ on share one layout.

\begin{figure}[t]
\centering
\resizebox{0.995\linewidth}{!}{\input{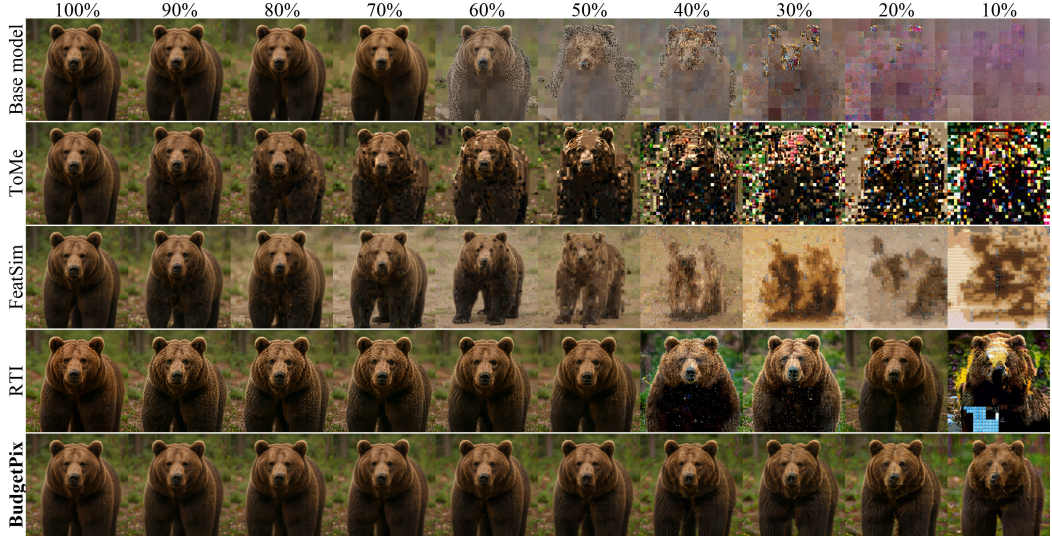}}%
\vspace{-5pt}
\caption{\textbf{Qualitative comparison across budgets, MiniT2I-L/16 at $512^2$.} Same prompt and noise were used for all methods. \methodName{} generates image with high quality even at 10\%.}
\label{fig:qual_minit2i_l}
\end{figure}

\section{Experiments}
\label{sec:setup}

\subsection{Implementation details}
\label{sec:impl}

We adapt seven released pixel-space diffusion transformers:
JiT-B/16 and JiT-L/16 at $256^2$ and JiT-B/32 and JiT-L/32 at $512^2$ \citep{li2026back}, MiniT2I-B/16 and
MiniT2I-L/16 \citep{minit2i2026} at $512^2$, and PixelDiT-T2I \citep{yu2026pixeldit} at $1024^2$, plus the
one-step pixel MeanFlow pMF-L/16 \citep{lu2026onestep} at $256^2$.
Each is fine-tuned once with the layout augmentation (with $a=0.15$, $b=0.1$) using the objective described in Section~\ref{sec:traininf}. 
The added modules are zero-initialized, so training starts exactly from the released model (see App.~\ref{app:loss}--\ref{app:settings}).

\subsection{Evaluation settings}
\label{sec:evalsettings}

\paragraph{Baselines.}

We compare \methodName{} against the released checkpoints driven to the same budget through our sampler 
and three token-reduction methods on those weights. The three token-reduction baselines are: (i) ToMe \citep{bolya2023token},
the feature-similarity merge of
ToMeSD \citep{bolya2023tokensd} applied once, (ii) down to the budget \citep{bolya2023tokensd} (\emph{FeatSim}), and (iii) RTI \citep{zamfir2026elastic}, a
concurrent adapter with region tokens, run from its released checkpoint. Every
method shares prompts or labels, initial noise, solver, step count and attention
arithmetic; only the budget and the reduction rule differ. See App.~\ref{app:protocol} for the
protocol.

\paragraph{Dataset and metrics.}

We use FID-50k \citep{heusel2017gans} and Inception score \citep{salimans2016improved} for
class-conditional quality, GenEval \citep{ghosh2023geneval} and DPG-Bench \citep{hu2024ella} for
prompt adherence, and CLIP \citep{radford2021learning}, PickScore \citep{kirstain2023pick} and
ImageReward \citep{xu2023imagereward} for preference on PartiPrompts \citep{yu2022scaling}. 
We additionally measure faithfulness,
whether a reduced-budget output preserves \emph{the same image}, using paired PSNR, SSIM \citep{wang2004image}, LPIPS \citep{zhang2018unreasonable}, CLIP and
DINOv2 \citep{oquab2024dinov} similarity against each method's own output at full 100\% budget. 
Cost is measured by wall-clock time; see App.~\ref{app:timing} for details.

\subsection{Analysis}
\label{sec:analysis}
\label{sec:results}

\begin{figure}[t]
    \centering
    \includegraphics[width=\linewidth]{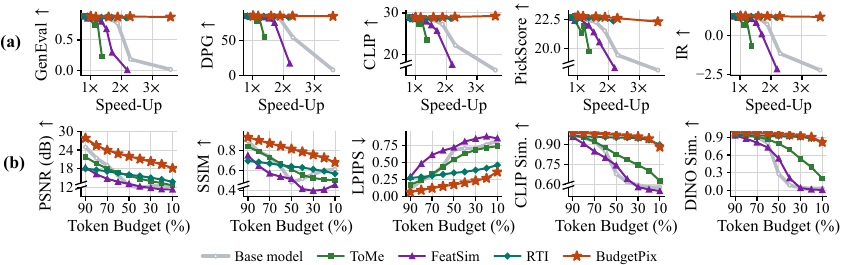}
    \vspace{-18pt}
    \caption{\textbf{Text-to-image results, MiniT2I-L/16 at $512^2$.} \textbf{(a)} Each metric plotted against the wall-clock
    speed-up over the released model (DPG: DPG-Bench; IR: ImageReward denote metrics). \textbf{(b)} Faithfulness to the method's own
    full-budget image plotted against the token budget (Sim.: CLIP or DINOv2 feature similarity).}
    \label{fig:t2i}
\end{figure}

\begin{table}
    \centering\footnotesize
    \caption{\textbf{Text-to-image generation} with PixelDiT-T2I at $1024^2$. \methodName{} consistently outperform the base model across different compute budgets.}
    \vspace{-2ex}
    \label{tab:main_1024}
\begin{tabular*}{\linewidth}{@{\extracolsep{\fill}}ll rrrrrrr@{}}
    \toprule
    Metric & Method & 100\% & 90\% & 80\% & 70\% & 60\% & 50\% & 25\% \\
    \midrule
    GenEval $\uparrow$ & Base model & 0.721 & 0.722 & 0.724 & 0.725 & 0.708 & 0.674 & 0.323 \\
     & \methodName{} & \textbf{0.741} & \textbf{0.740} & \textbf{0.741} & \textbf{0.738} & \textbf{0.746} & \textbf{0.743} & \textbf{0.725} \\
    \midrule
    DPG-Bench $\uparrow$ & Base model & 84.8 & 84.6 & 84.2 & 84.0 & 83.3 & 81.8 & 55.1 \\
     & \methodName{} & \textbf{85.2} & \textbf{85.2} & \textbf{85.0} & \textbf{85.0} & \textbf{84.6} & \textbf{84.6} & \textbf{83.7} \\
    \bottomrule
\end{tabular*}

\end{table}

\begin{wrapfigure}{R}{0.45\linewidth}
    \vspace{-2ex}
    \centering
    \includegraphics[width=\linewidth]{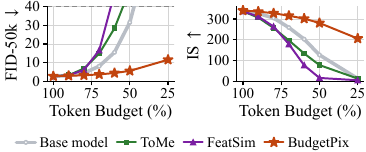}
    \caption{\textbf{FID and Inception Score (IS) on class-conditional generation across token budgets on ImageNet, JiT-L/32 at $512^2$.} FID is cut at 40.}
    \label{fig:jit_l}
    \vspace{-1.0\baselineskip}
\end{wrapfigure}

\paragraph{Class-conditional generation.}

On $512^2$ generation for JiT-L/32 (Figure~\ref{fig:jit_l}), \methodName{} starts at the released model's FID and increases gradually as the budget decreases, whereas the base model and merging baselines increase sharply.
At 50-60\% budgets, their FID is several times ours, with ToMe and FeatSim performing worse than the base model at the same budget. Inception Score shows the same trend: \methodName{}
retains FID scores of a fully dense layout even at a 60\% budget, where the baselines deteriorate substantially. Additional results for JiT-L/16 at $256^2$ are provided in App.~\ref{app:jit} and more qualitative results are presented in App.~\ref{app:ours_jit}. 

\paragraph{Faithfulness across the budget.}

We measure faithfulness, that is, how much a method's image at a reduced budget similar to the image the same method produces at their full budget. In Figure~\ref{fig:t2i}(b), ours stays closest to its dense image on
nearly every metric and budget, while the base model and both baselines significantly diverge from the original image at 70\% budget. These gaps can be attributed to the fact that our layout merges the flattest regions first and the refiner restores their texture, whereas the merging baselines group tokens by feature similarity across the whole frame and the base model receives coarse tokens it was
never trained on. See App.~\ref{app:perc} for the base model and ours on JiT.

\paragraph{Text-to-image generation at $512^2$ and $1024^2$.}
\label{sec:results_t2i}
\label{sec:results_1024}
On MiniT2I-L
(Figure~\ref{fig:t2i}(a); values in Table~\ref{tab:main_t2i512}) GenEval and DPG-Bench \methodName{} achieves similar, stable scores across compute budgets while the base model
degrades significantly, to near-zero scores. A similar trend is also observed in PixelDiT at
$1024^2$ resolution (Table~\ref{tab:main_1024}). In addition, Figure~\ref{fig:qual_minit2i_l} visually demonstrates the images generated by each baseline at every budget, confirming the stable quality of our model across budgets. For more results, see
App.~\ref{app:methods}, App.~\ref{app:ours_jit}, and App.~\ref{app:layouts_pixeldit}.

\paragraph{User study and VLM-as-Judge.}

\begin{figure}
  \begin{minipage}[c]{0.64\textwidth}
    \vspace{-10pt}
    \includegraphics[width=0.98\linewidth]{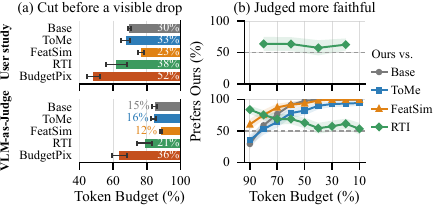}
  \end{minipage}\hfill
  \begin{minipage}[c]{0.35\textwidth}
    \caption{\textbf{User study (top) and VLM-as-Judge (bottom), MiniT2I-L/16} at $512^2$: (a) How many percentage from the compute budget can be cut without a noticeable difference for human and VLM judges. 
    (b) Head-to-head comparison of \methodName{} against baselines by human and VLM judges, showing our method is consistently preferred across all token budgets.}
    \label{fig:vlm}
  \end{minipage}
  \vspace{-2ex}
\end{figure}

The paired metrics say how far an image moved, not when a reader would notice.
In a user study (Figure~\ref{fig:vlm}), 50 people per form compare images on 20 GenEval prompts, and in
VLM-as-Judge a vision--language model does the same on 100. Shown a method's own dense image and a grid of its
budgets, the budget can be cut by 52\,\% for ours before human raters first see a difference, against at most 38\,\%
for the baselines, and by 36\,\% against at most 21\,\% before the VLM judge does (Figure~\ref{fig:vlm}a). When shown the full-budget image and two budget-reduced candidates from two different methods, human raters find ours more faithful than RTI at all four budgets considered. The VLM judge also favors ours than every baseline below 80\,\%  (Figure~\ref{fig:vlm}  b). See App.~\ref{app:vlm} for both protocols.

\paragraph{Wall-clock and speed-up.}

Figure~\ref{fig:t2i}(a) plots every text-to-image
metric against the measured wall-clock speed-up over the full budget inference time. \methodName{} maintains its scores across all five metrics and on the whole speed-up range, the base model starts at the same value but collapses fast. The merging baselines buy less speed, since
the merge itself costs time. 
See App.~\ref{app:wallclock} for the other base models.

%

\begin{table}
    \centering\footnotesize
    \caption{\textbf{Ablation on JiT-B, FID-50k.} \methodName{} with both Patch Aggregator (sec~\ref{sec:encoder}) and Refiner (sec~\ref{sec:decoder}) works best across different compute budgets.}
    \label{tab:abl_components}
    \vspace{-2ex}
    \begin{tabularx}{\textwidth}{llXXXXXXX}
        \toprule
         & Model & 100\% & 90\% & 80\% & 70\% & 60\% & 50\% & 25\% \\
        \midrule
        \multirow{3}{*}{$512^2$} & Base Model & 4.12 & 5.00 & 6.86 & 12.24 & 24.79 & 60.61 & 190.71 \\
         & + Patch Aggregator & 3.96 & 4.52 & 5.26 & 6.67 & 8.56 & 11.58 & 25.07 \\
         & + Patch Refiner & {\bf 3.94} & {\bf 4.35} & {\bf 4.80} & {\bf 5.57} & {\bf 6.59} & {\bf 8.30} & {\bf 16.71} \\
        \midrule
        \multirow{3}{*}{$256^2$} & Base Model & 3.61 & 4.24 & 5.84 & 10.67 & 23.46 & 63.94 & 145.46 \\
         & + Patch Aggregator & {\bf 3.57} & 3.92 & 4.34 & 5.14 & 6.17 & 7.84 & 16.90 \\
         & + Patch Refiner & 3.60 & {\bf 3.89} & {\bf 4.25} & {\bf 4.89} & {\bf 5.73} & {\bf 7.01} & {\bf 13.05} \\
        \bottomrule
    \end{tabularx}
\end{table}

\begin{wrapfigure}{R}{0.5\linewidth}
    \vspace{-1.0\baselineskip}
    \centering
    \includegraphics[width=\linewidth]{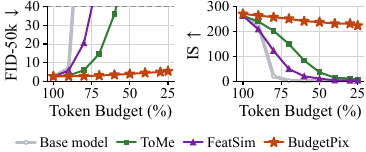}
    \vspace{-16pt}
    \caption{\textbf{\methodName{} on one-step pixel MeanFlow pMF-L/16.} \methodName{} provides a better quality-efficiency trade-off compared with the base model and other token merging baselines.}
    \label{fig:pmf_main}
    \vspace{-1.0\baselineskip}
\end{wrapfigure}

\paragraph{One-step generation with MeanFlow.}
\label{sec:pmf}

\methodName{} reduces the cost per step, rather than the number of steps, making it complementary to inference-efficiency methods such as distillation. We note that \methodName{} is still applicable to one-step generation methods by using the random layout (described in sec~\ref{sec:traininf}) instead of the layout from the warm-up steps. Figure~\ref{fig:pmf_main} shows our results on one-step (1-NFE) pixel MeanFlow
(pMF-L/16;~\citealp{lu2026onestep}), fine-tuned on layouts from training images of each class. While the released model's weights collapse at 80\,\%; \methodName{} maintains FID below 4 down at 50\% budget (see App.~\ref{app:pmf} and~\ref{app:ours_pmf} for more results).

\paragraph{Ablation study.}
\label{sec:qual}

Table~\ref{tab:abl_components} provides the ablation study of \methodName{} components. The aggregator
does most of the work, since it helps coarse tokens carry their region's content. The refiner takes roughly a further third
off, and its contribution grows as the budget falls, since it acts only inside coarse regions. 
See App.~\ref{app:ablations} for other ablations and
App.~\ref{app:layoutrule} for the entropy quadtree against random and oracle
layouts.

\section{Conclusion}
\label{sec:discussion}

\methodName{} shows that a pretrained pixel-space diffusion transformer's token count does not need to remain fixed to its pretraining configuration. After one fine-tuning, a single checkpoint serves budgets spanning from a dense grid to a small fraction of it and degrades gracefully where token-merging baselines collapse, even at 1 NFE (Figure~\ref{fig:pmf_main}). Its limitations are that the layout is derived from the model's intermediate predictions, so it can miss details they do not yet show, and that full-resolution stages, such as PixelDiT's pixel stage, cap the speed-up. Learning where to place tokens, not deriving it from entropy, is a natural next step.

\clearpage   
\bibliography{refs}
\bibliographystyle{preprint}

\appendix
\makeatletter
\def\addcontentsline#1#2#3{\addtocontents{#1}{\protect\contentsline{#2}{#3}{\thepage}{}}}
\makeatother
\setcounter{tocdepth}{2}
\addtocontents{toc}{\protect\setcounter{tocdepth}{2}}
\raggedbottom                                   
\newlength{\spanw}
\newcommand{\spanrule}{\leavevmode\leaders\vrule height 0.62ex depth -0.52ex\hfill\kern0pt}
\newcommand{\spanned}[1]{\settowidth{\spanw}{#1}%
  \ifdim\spanw>\dimexpr\linewidth-3em\relax
    \spanrule\,\parbox[c]{\dimexpr\linewidth-3em\relax}{\centering #1}\,\spanrule
  \else\spanrule\ #1\ \spanrule\fi}

\clearpage
\section*{supplementary material}

\renewcommand{\contentsname}{Contents of the supplementary material}
{\small\tableofcontents}

\clearpage
\section{Additional qualitative results across the budget}
\label{sec:moresamples}
\label{app:gallery}

\subsection{Baseline Comparison (MiniT2I-L/16, $512^2$)}
\label{app:methods}

\begin{figure}[H]
\centering
\includegraphics[width=\linewidth]{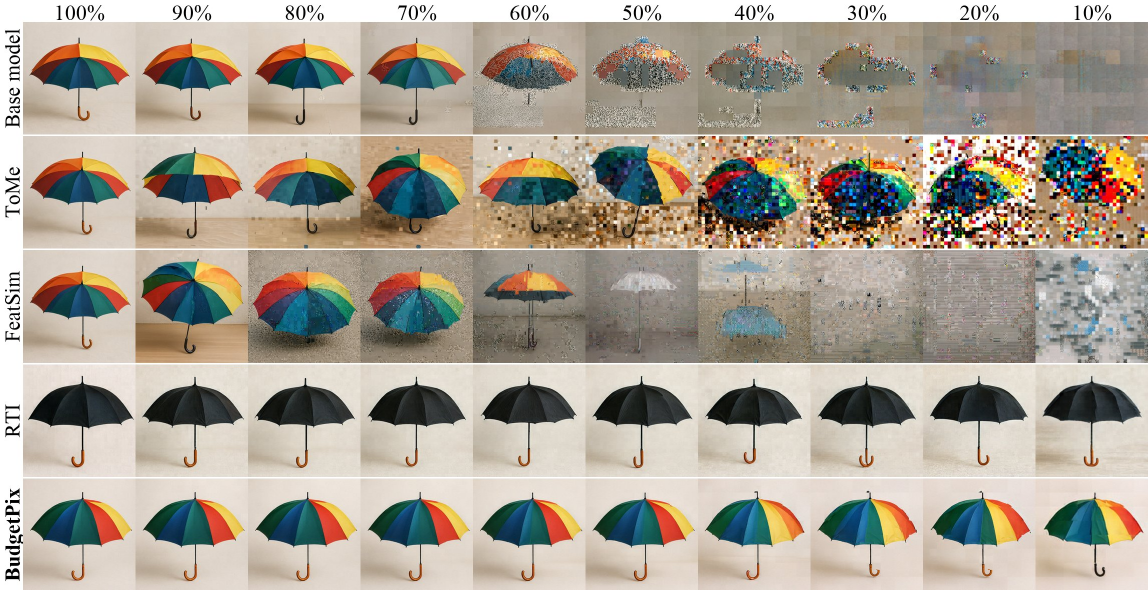}
\caption{\textbf{Every method across the budget: an umbrella.}}
\label{fig:methods_000044}
\end{figure}

\begin{figure}[H]
\centering
\includegraphics[width=\linewidth]{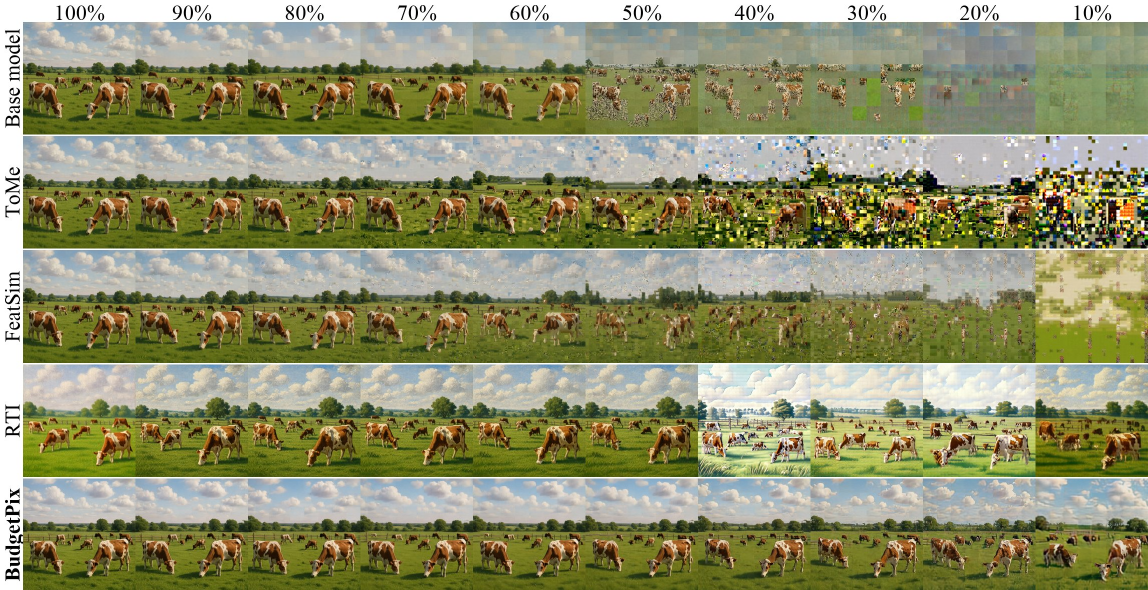}
\caption{\textbf{Every method across the budget: a herd of cows.}}
\label{fig:methods_000099}
\end{figure}

\begin{figure}[H]
\centering
\includegraphics[width=\linewidth]{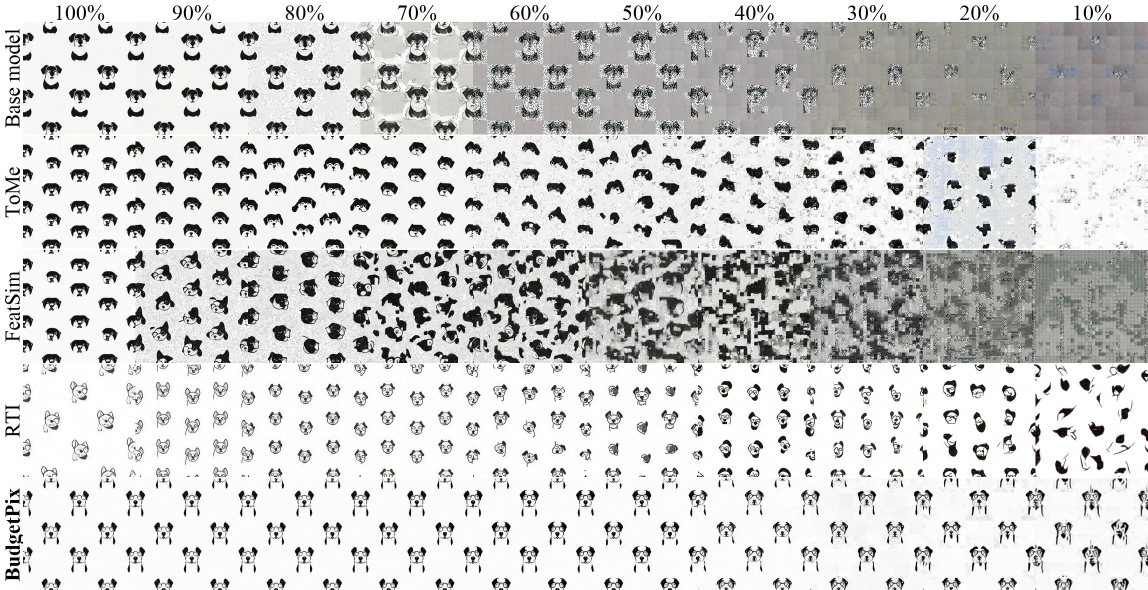}
\caption{\textbf{Every method across the budget: a seamless vector pattern.}}
\label{fig:methods_000031}
\end{figure}

\begin{figure}[H]
\centering
\includegraphics[width=\linewidth]{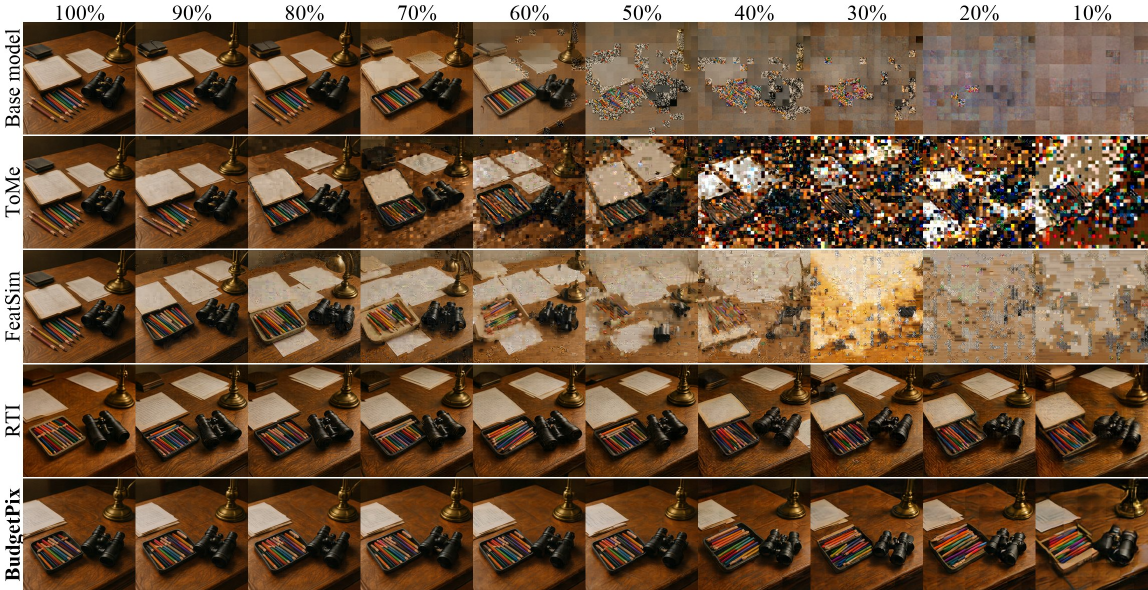}
\caption{\textbf{Every method across the budget: a pencil case and binoculars.}}
\label{fig:methods_000086}
\end{figure}

\begin{figure}[H]
\centering
\includegraphics[width=\linewidth]{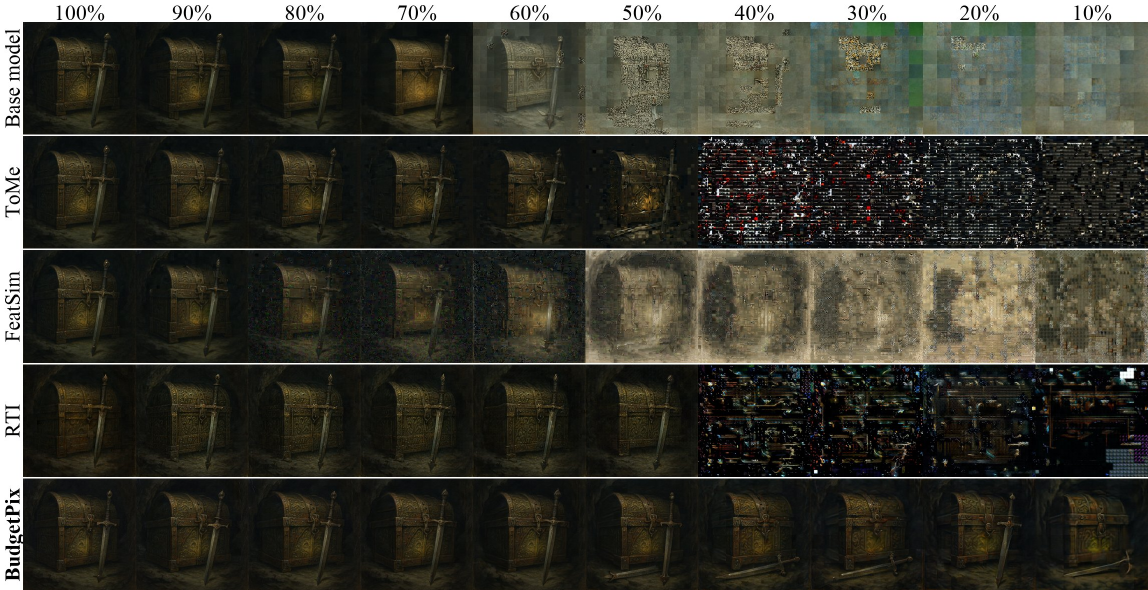}
\caption{\textbf{Every method across the budget: an ornate treasure chest.}}
\label{fig:methods_000076}
\end{figure}

\begin{figure}[H]
\centering
\includegraphics[width=\linewidth]{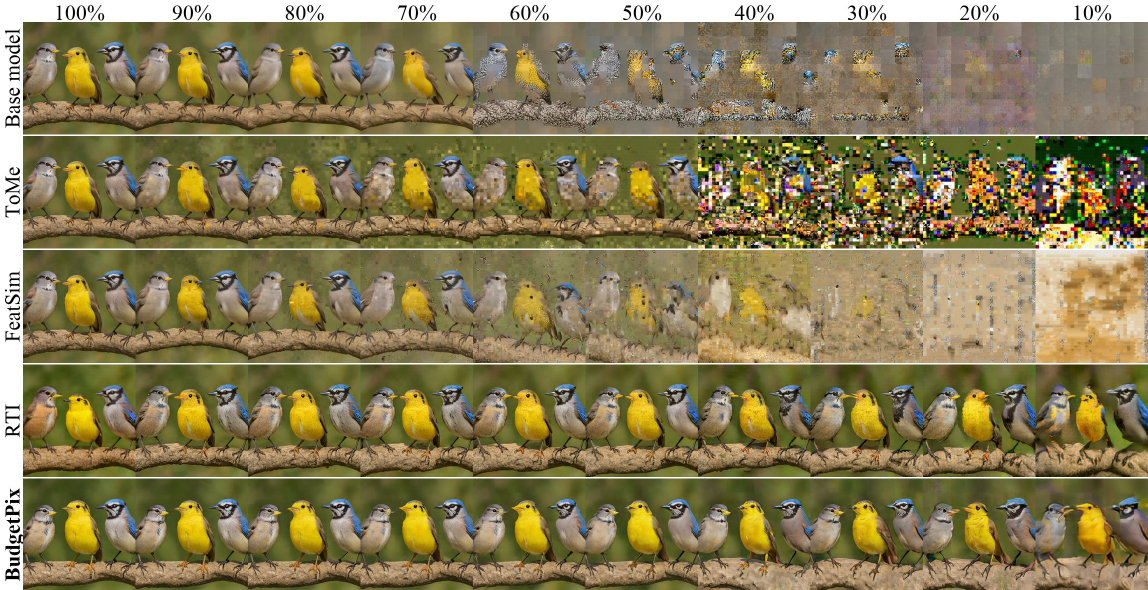}
\caption{\textbf{Every method across the budget: three birds on a branch.}}
\label{fig:methods_000053}
\end{figure}

Figures~\ref{fig:methods_000044}--\ref{fig:methods_000053} show the comparison with the baselines on MiniT2I-L/16 at $512^2$, one prompt per figure. Each row is a method (the base model, ToMe, FeatSim, RTI and ours), and each column a token budget, from 100\,\% to 10\,\%.

\clearpage
\subsection{\methodName{} (JiT-L/32, $512^2$)}
\label{app:ours_jit}

\begin{figure}[H]
\centering
\input{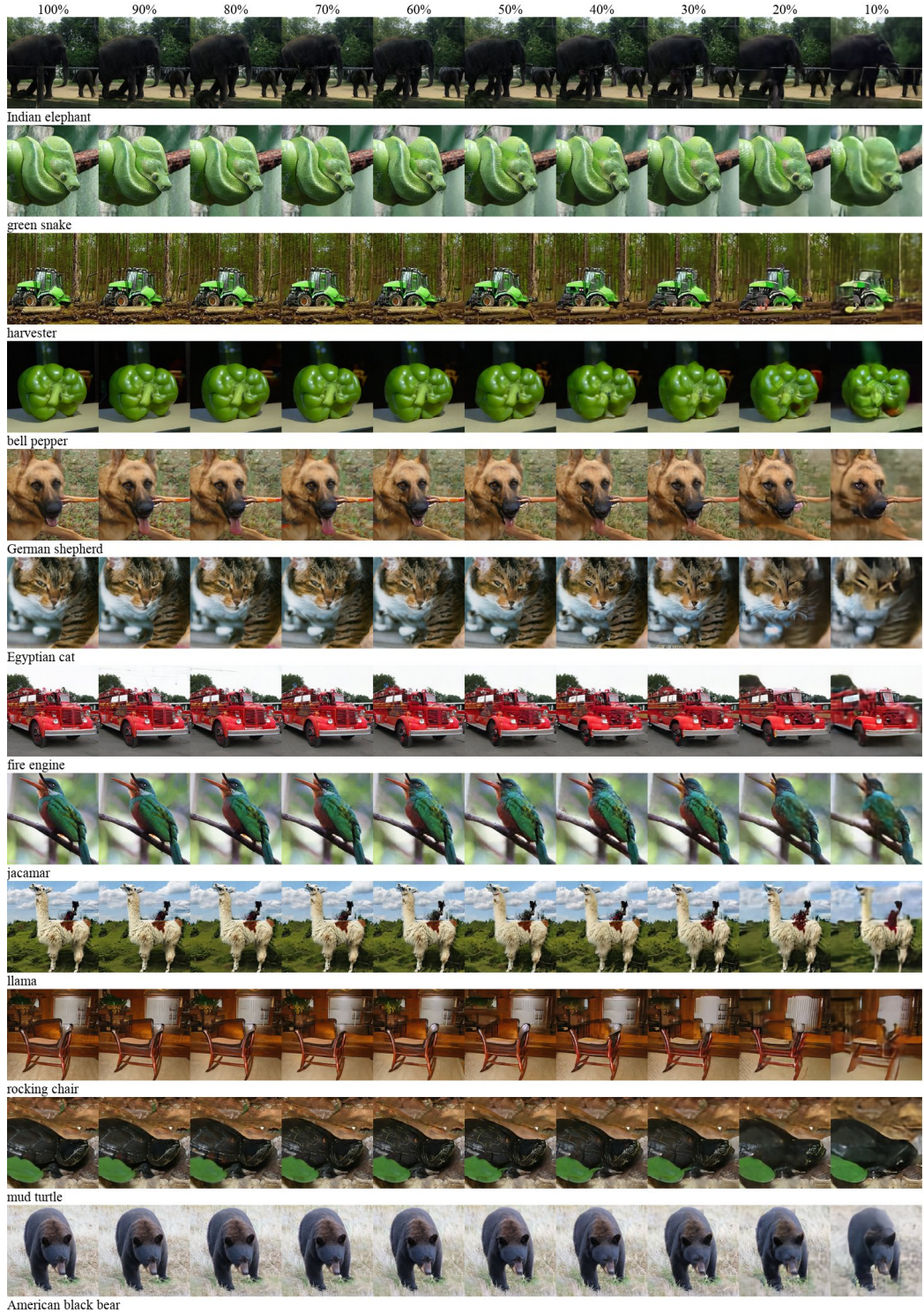}%
\caption{\textbf{\methodName{} on JiT-L/32 at $512^2$, one ImageNet class per row.}}
\label{fig:app_jit_l_1}
\end{figure}

\begin{figure}[H]
\centering
\input{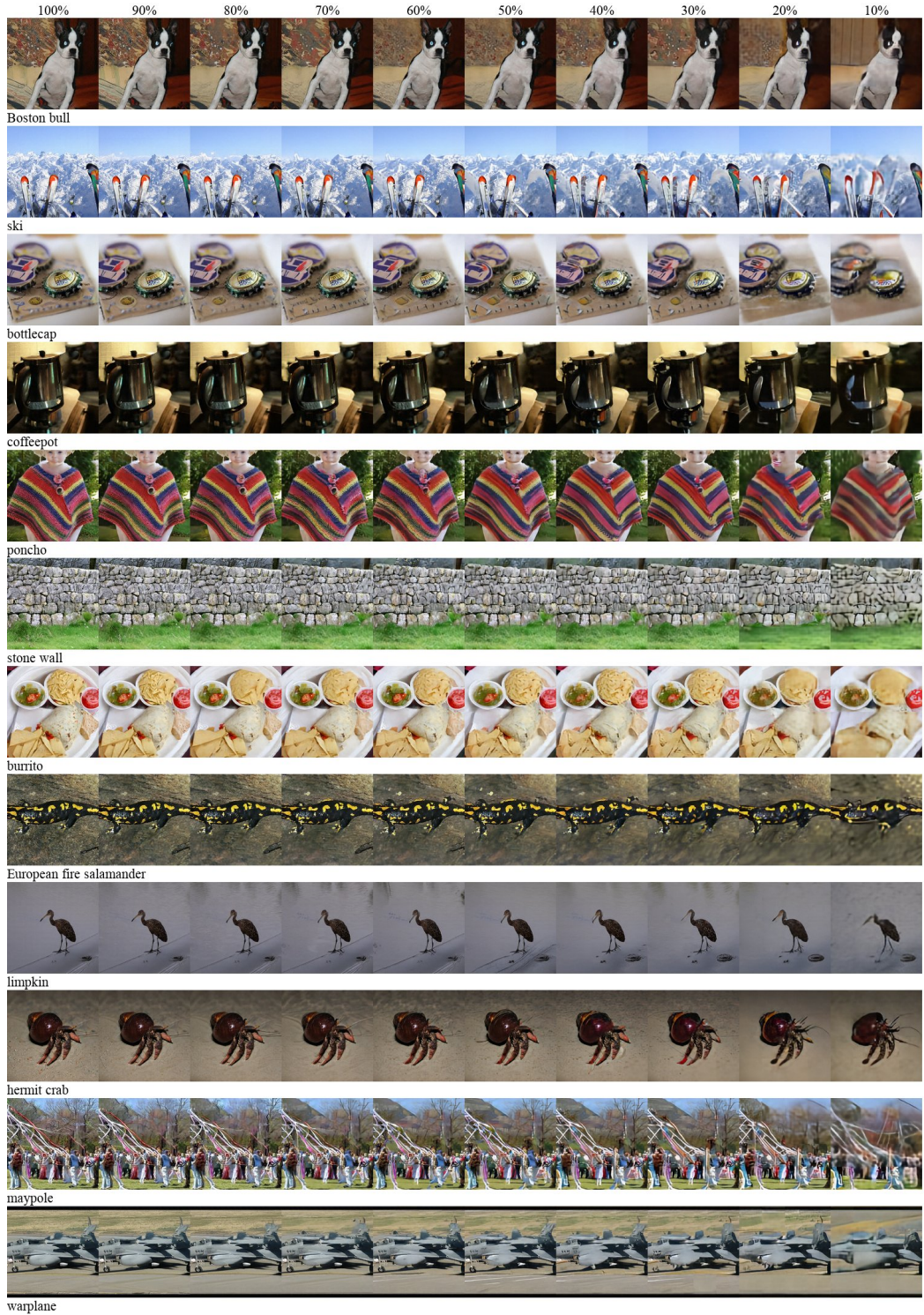}%
\caption{\textbf{\methodName{} on JiT-L/32 at $512^2$, one ImageNet class per row (continued).}}
\label{fig:app_jit_l_2}
\end{figure}

Figures~\ref{fig:app_jit_l_1} and~\ref{fig:app_jit_l_2} show additional qualitative results on JiT-L/32 at $512^2$. Each row is one ImageNet class, and each column a token budget, from 100\,\% to 10\,\%.

\clearpage
\subsection{\methodName{} (MiniT2I-L/16, $512^2$)}
\label{app:ours_minit2i}

\begin{figure}[H]
\centering
\input{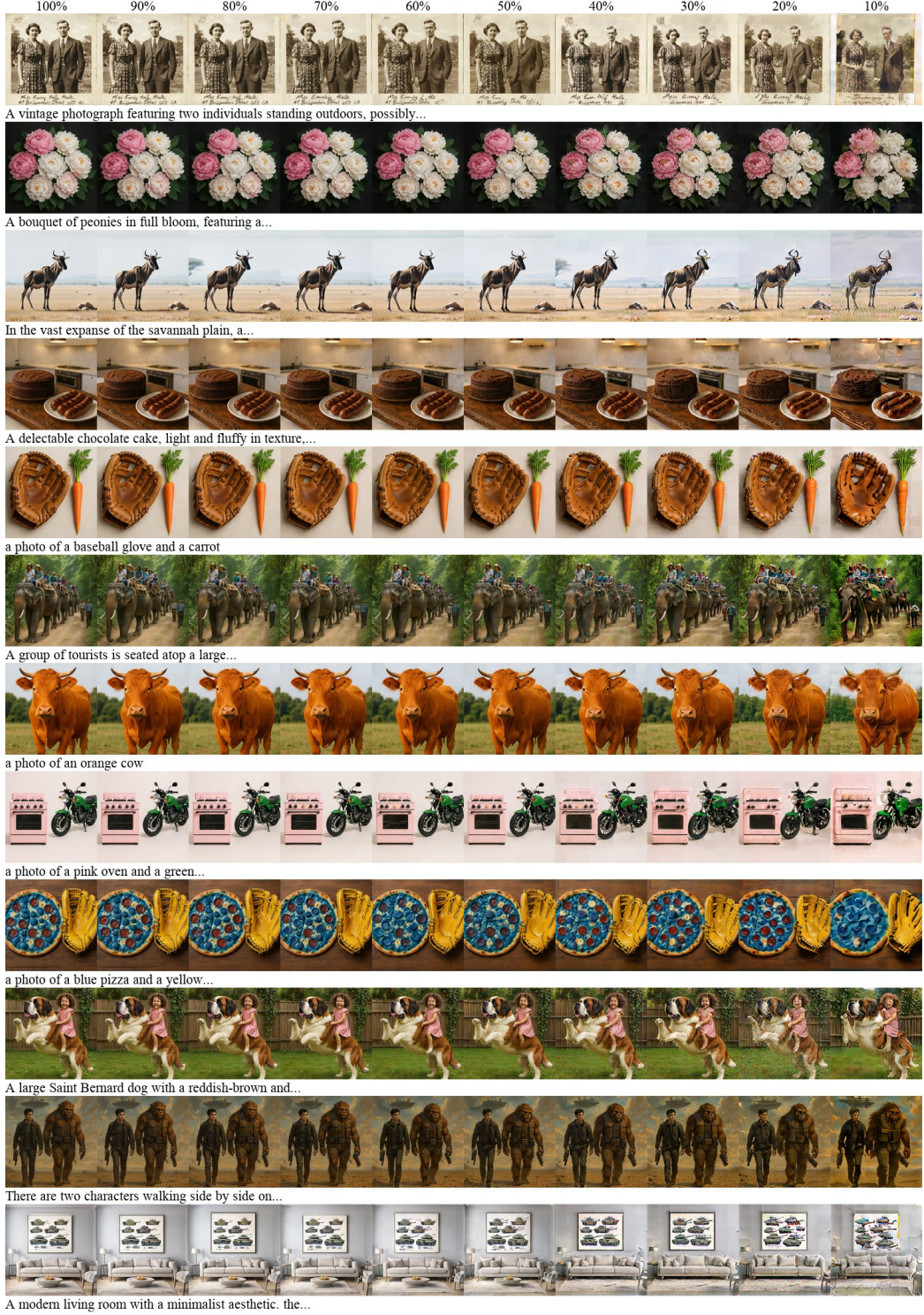}%
\caption{\textbf{\methodName{} on MiniT2I-L/16 at $512^2$, one prompt per row.}}
\label{fig:app_minit2i_l}
\end{figure}

\begin{figure}[H]
\centering
\input{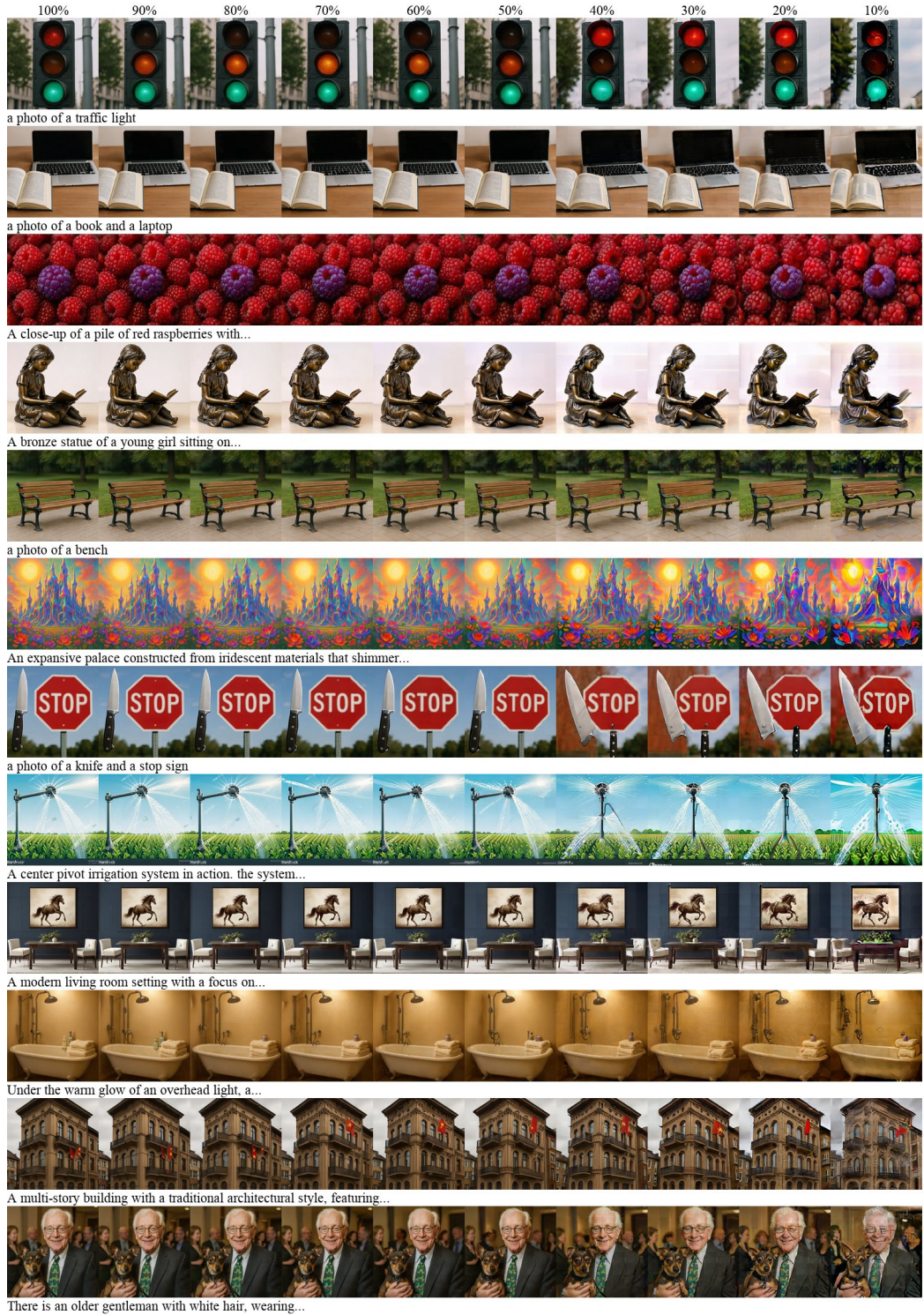}%
\caption{\textbf{\methodName{} on MiniT2I-L/16 at $512^2$, one prompt per row (continued).}}
\label{fig:app_minit2i_l_2}
\end{figure}

Figures~\ref{fig:app_minit2i_l} and~\ref{fig:app_minit2i_l_2} show additional qualitative results on MiniT2I-L/16 at $512^2$. Each row is one prompt, and each column a token budget, from 100\,\% to 10\,\%.

\clearpage
\subsection{\methodName{} (PixelDiT-T2I, $1024^2$)}
\label{app:ours_pixeldit}

\begin{figure}[H]
\centering
\input{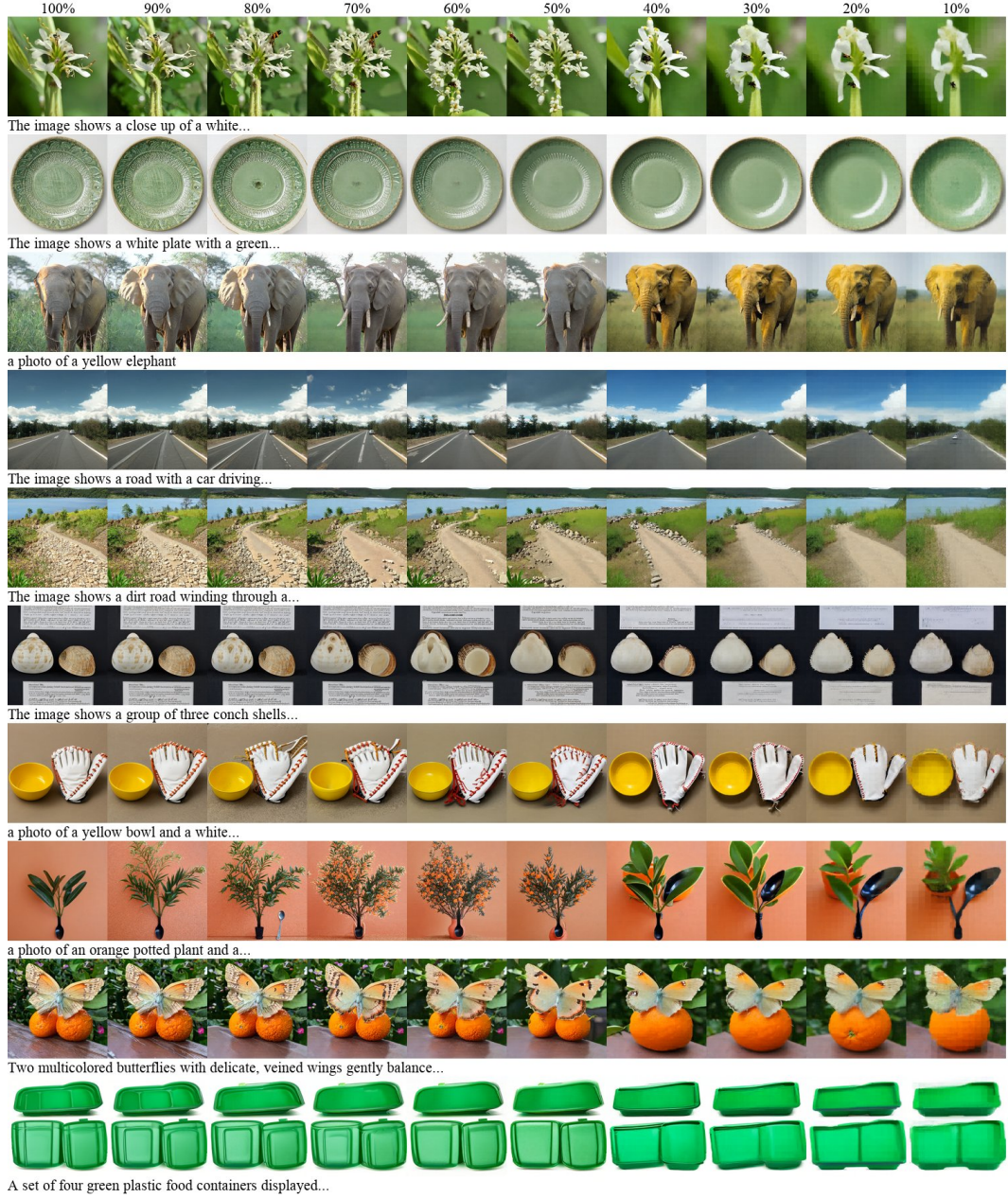}%
\caption{\textbf{\methodName{} on PixelDiT-T2I at $1024^2$, one prompt per row.}}
\label{fig:app_pixeldit}
\end{figure}

\begin{figure}[H]
\centering
\input{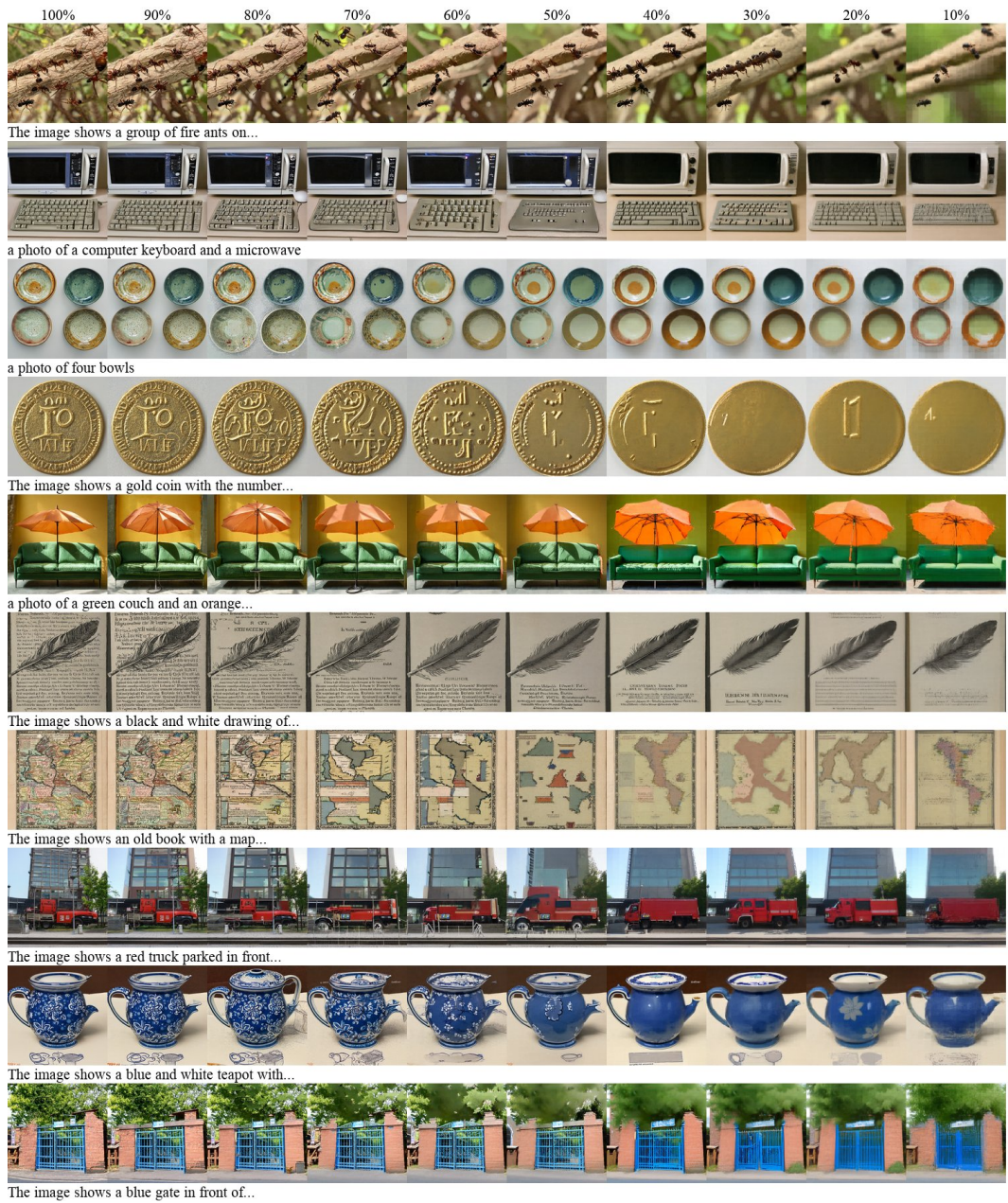}%
\caption{\textbf{\methodName{} on PixelDiT-T2I at $1024^2$, one prompt per row (continued).}}
\label{fig:app_pixeldit_2}
\end{figure}

Figures~\ref{fig:app_pixeldit} and~\ref{fig:app_pixeldit_2} show additional qualitative results on PixelDiT-T2I at $1024^2$. Each row is one prompt, and each column a token budget, from 100\,\% to 10\,\%.

\clearpage
\subsection{\methodName{} (pMF-L/16, $256^2$, 1 NFE)}
\label{app:ours_pmf}

\begin{figure}[H]
\centering
\input{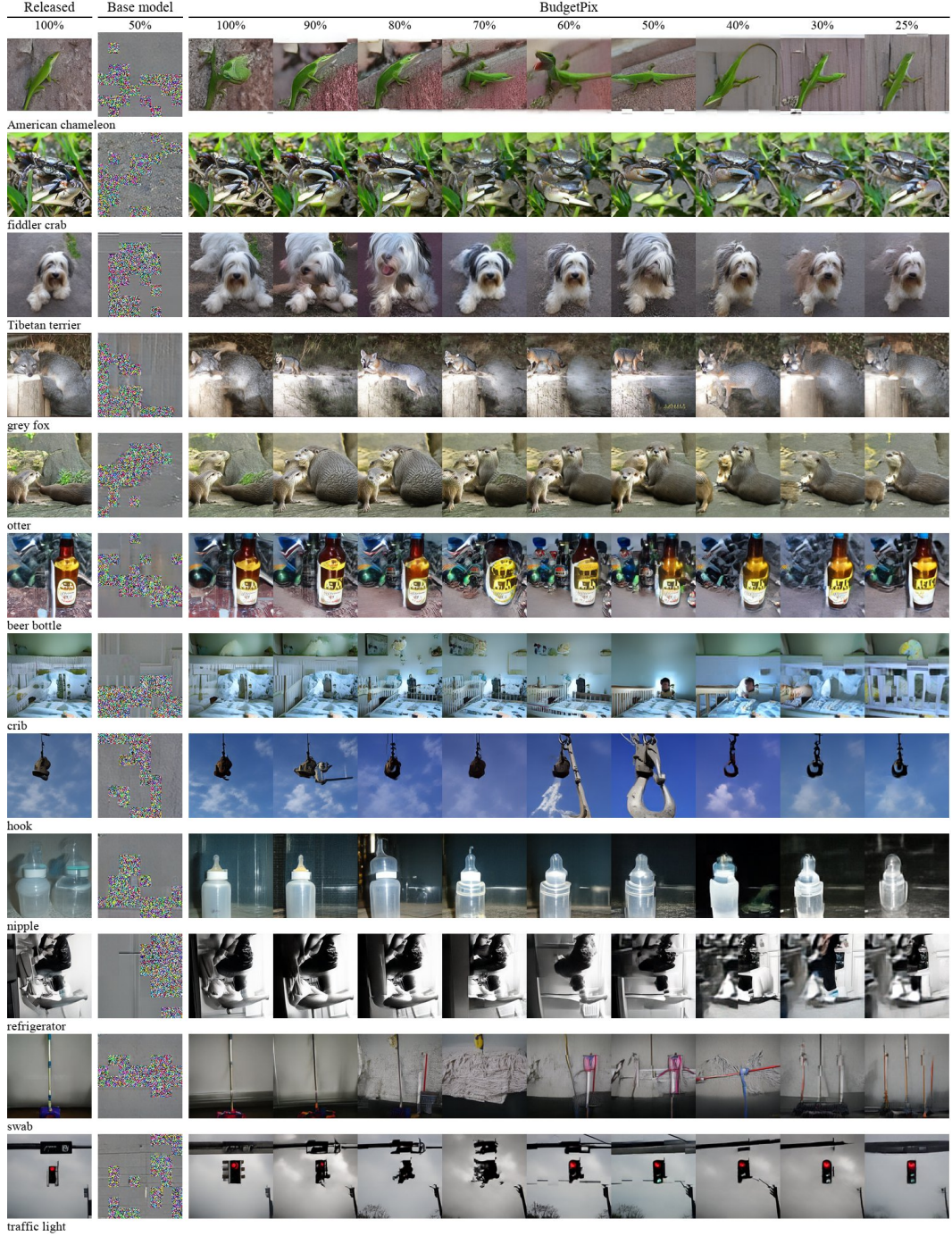}%
\caption{\textbf{One-step samples, pMF-L/16 at $256^2$, one ImageNet class per row.}}
\label{fig:pmf_samples}
\end{figure}

\begin{figure}[H]
\centering
\input{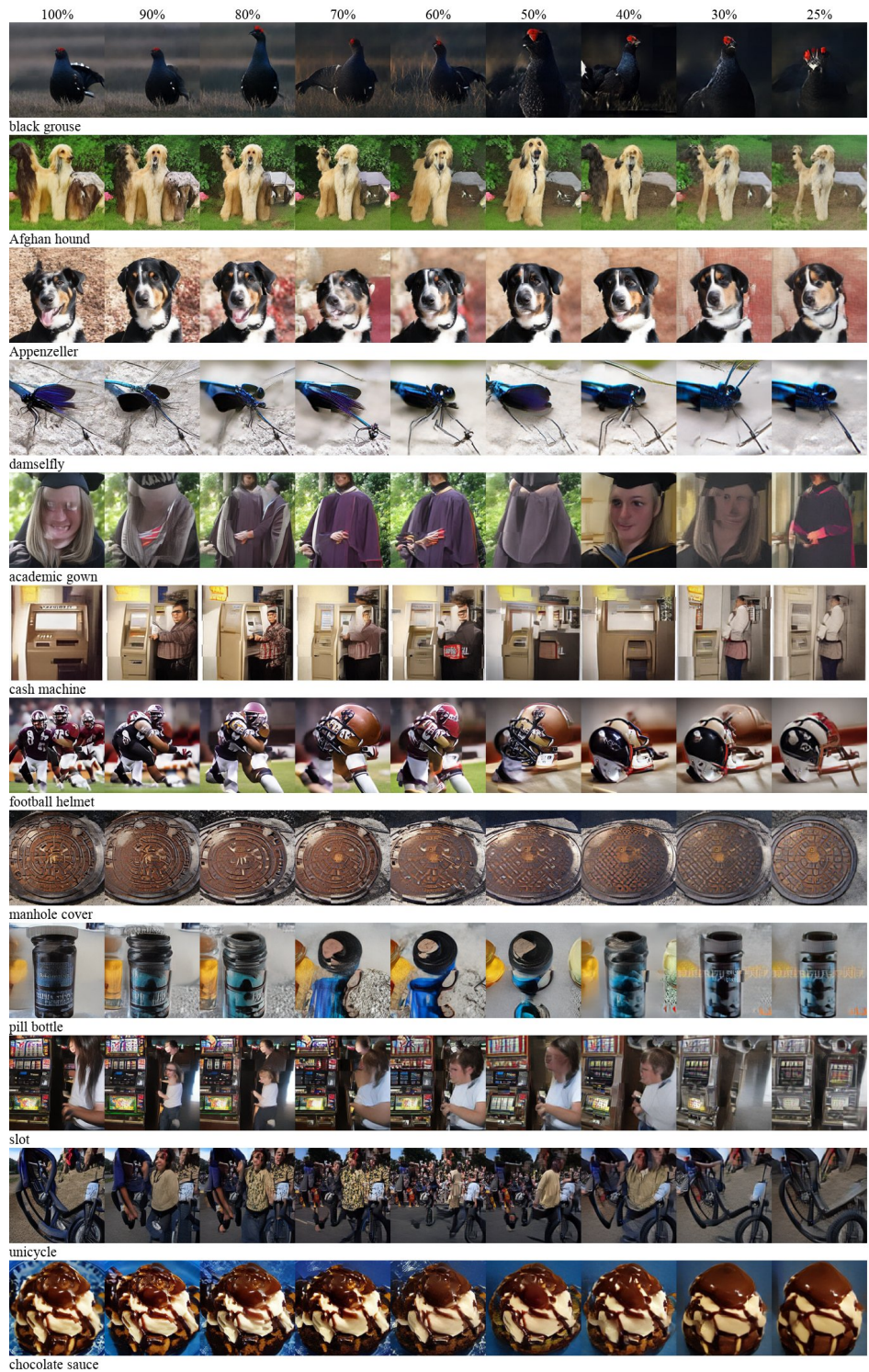}%
\caption{\textbf{\methodName{} on pMF-L/16 at $256^2$, one step, one ImageNet class per row.}}
\label{fig:pmf_ours_2}
\end{figure}

\begin{figure}[H]
\centering
\input{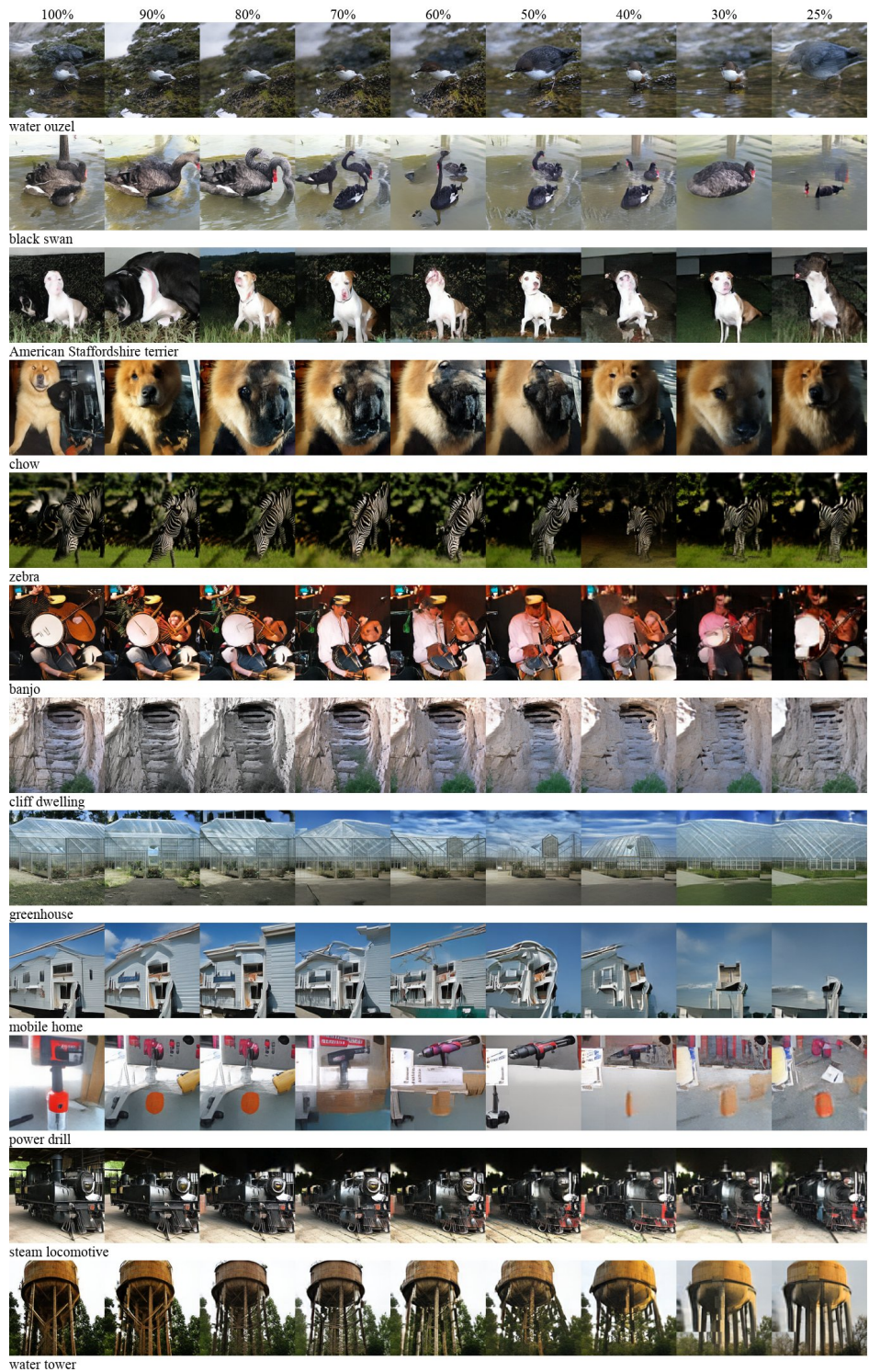}%
\caption{\textbf{\methodName{} on pMF-L/16 at $256^2$, one step, one ImageNet class per row (continued).}}
\label{fig:pmf_ours_3}
\end{figure}

Figures~\ref{fig:pmf_samples}--\ref{fig:pmf_ours_3} show additional qualitative results on the one-step pMF-L/16 at $256^2$. Each row is one ImageNet class, and each column a token budget, from 100\,\% to 25\,\%; Figure~\ref{fig:pmf_samples} first shows the released model at 100\,\% and the base model at 50\,\% (the released weights with our layouts, no training).

\clearpage
\subsection{\methodName{} layouts (JiT-L/32, $512^2$)}
\label{app:layouts_jit}

\begin{figure}[H]
\centering
\input{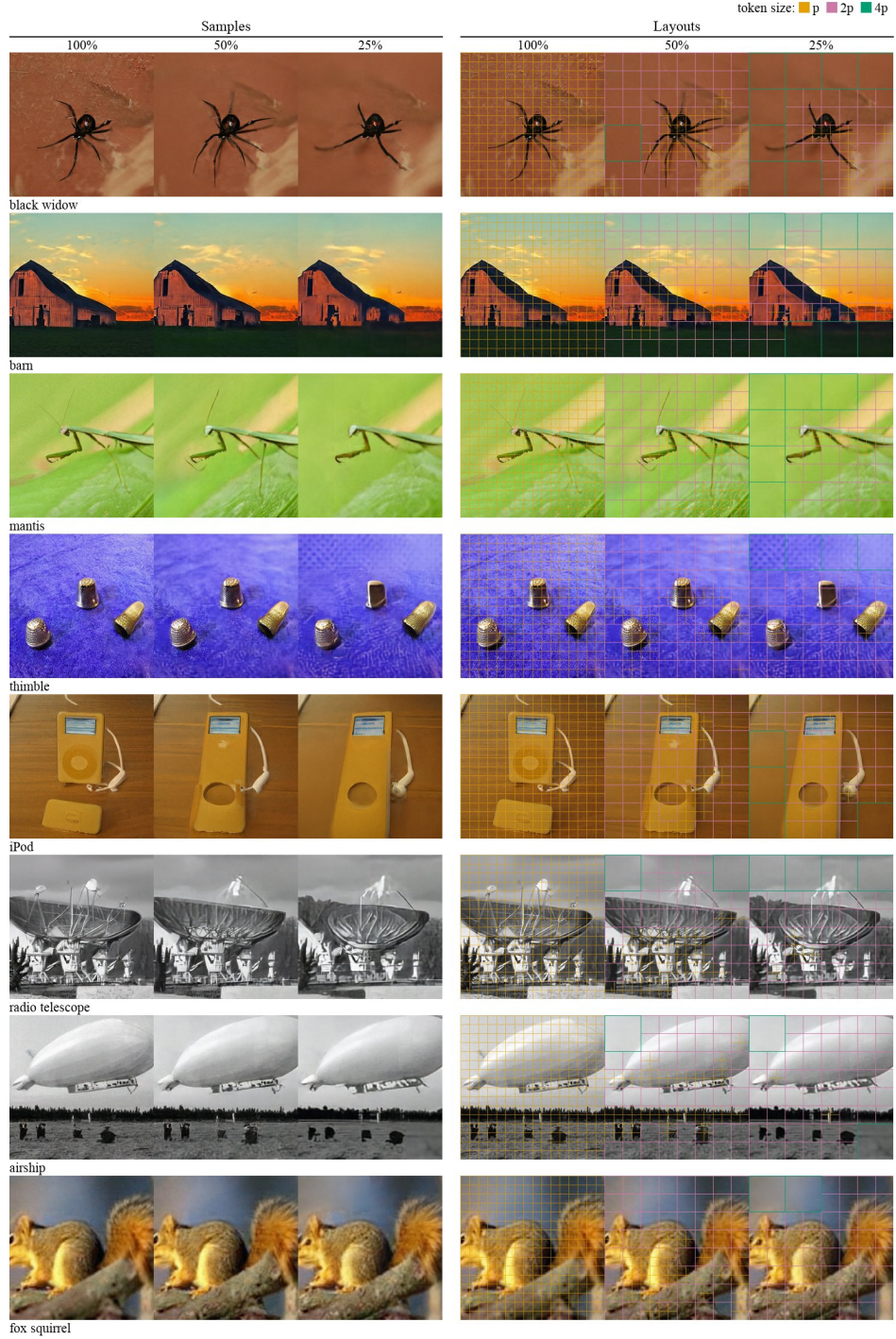}%
\caption{\textbf{\methodName{} on JiT-L/32 at $512^2$: samples and their layouts, one ImageNet class per row.}}
\label{fig:app_layout_jit_1}
\end{figure}

\begin{figure}[H]
\centering
\input{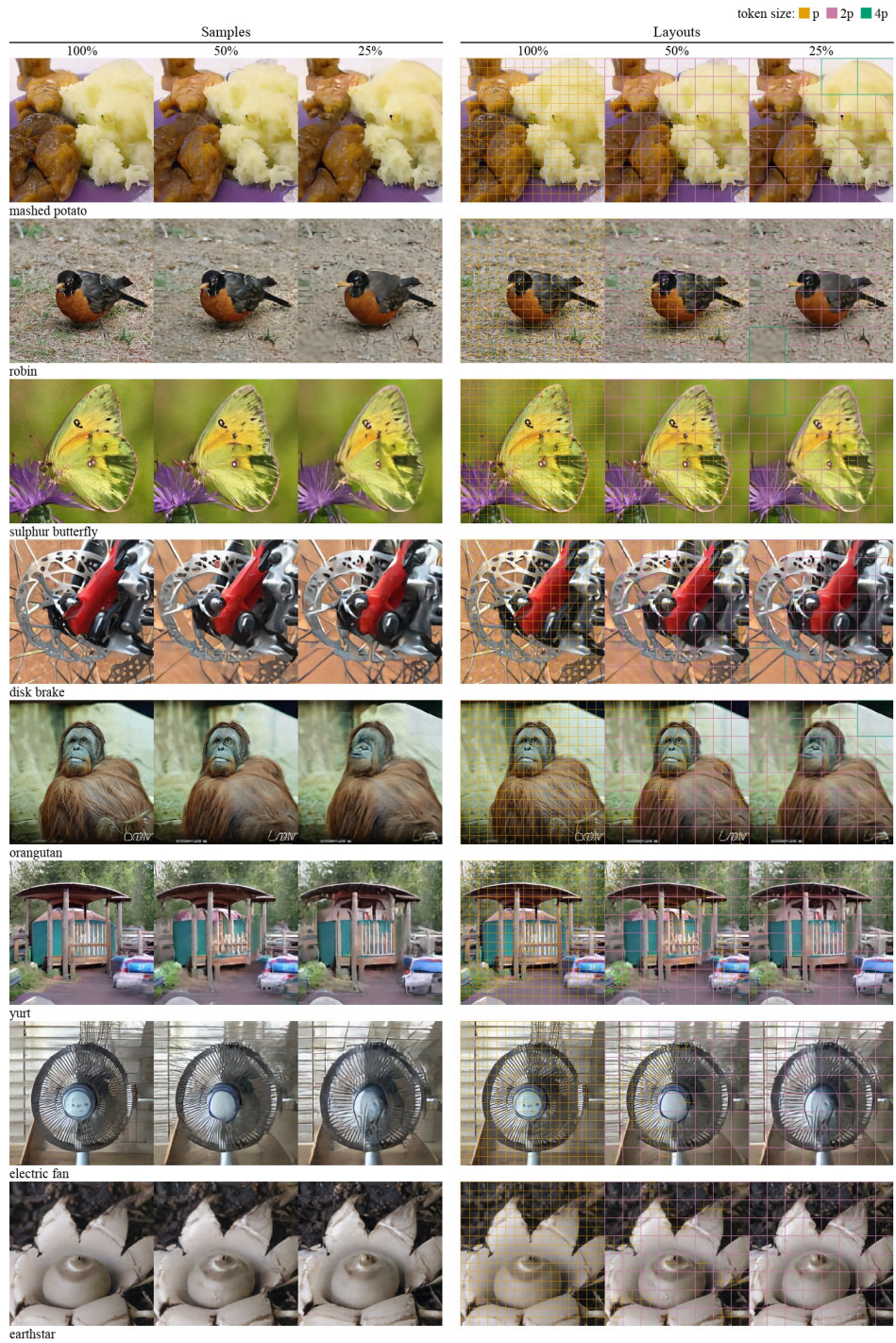}%
\caption{\textbf{\methodName{} on JiT-L/32 at $512^2$: samples and their layouts, one ImageNet class per row (continued).}}
\label{fig:app_layout_jit_2}
\end{figure}

Figures~\ref{fig:app_layout_jit_1} and~\ref{fig:app_layout_jit_2} show additional qualitative results with their layouts on JiT-L/32 at $512^2$. Each row is one ImageNet class; the columns are the samples at 100, 50 and 25\,\% of the grid, then the same images with their layouts drawn over them (each token outlined in the colour of its size).

\clearpage
\subsection{\methodName{} layouts (MiniT2I-L/16, $512^2$)}
\label{app:layouts_minit2i}

\begin{figure}[H]
\centering
\input{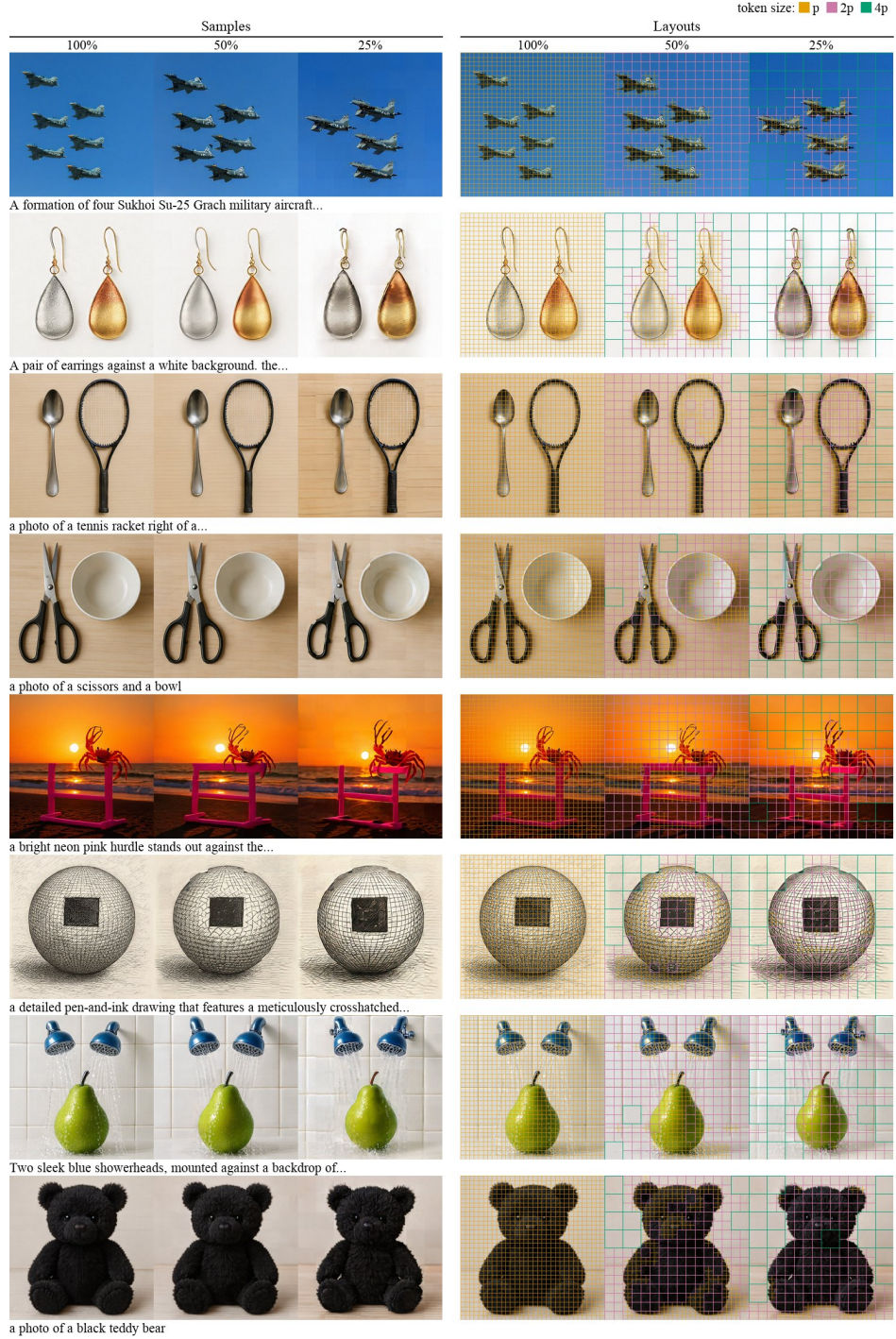}%
\caption{\textbf{\methodName{} on MiniT2I-L/16 at $512^2$: samples and their layouts, one prompt per row.}}
\label{fig:app_layout_minit2i_1}
\end{figure}

\begin{figure}[H]
\centering
\input{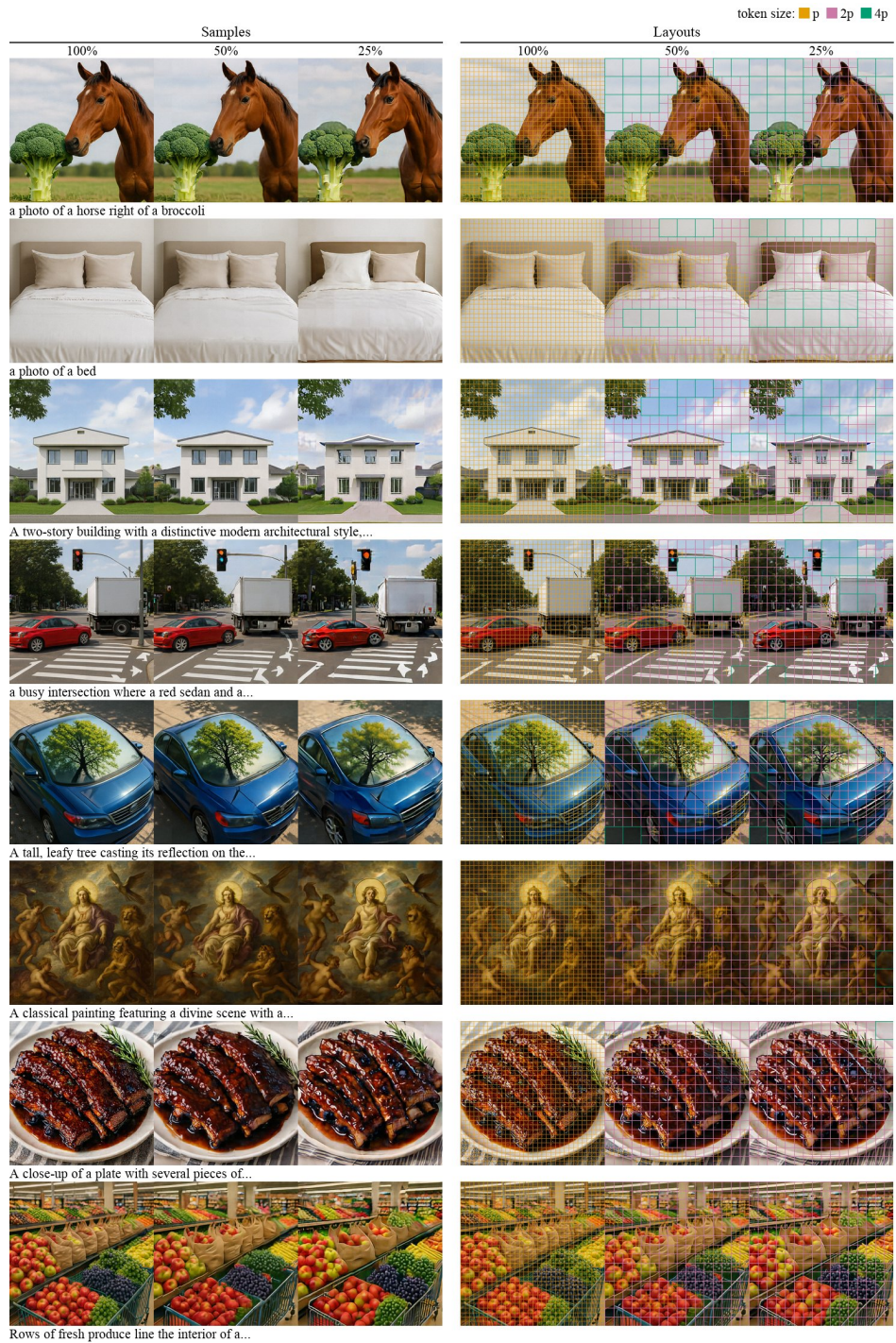}%
\caption{\textbf{\methodName{} on MiniT2I-L/16 at $512^2$: samples and their layouts, one prompt per row (continued).}}
\label{fig:app_layout_minit2i_2}
\end{figure}

Figures~\ref{fig:app_layout_minit2i_1} and~\ref{fig:app_layout_minit2i_2} show additional qualitative results with their layouts on MiniT2I-L/16 at $512^2$. Each row is one prompt; the columns are the samples at 100, 50 and 25\,\% of the grid, then the same images with their layouts drawn over them (each token outlined in the colour of its size).

\clearpage
\subsection{\methodName{} layouts (PixelDiT-T2I, $1024^2$)}
\label{app:layouts_pixeldit}

\begin{figure}[H]
\centering
\input{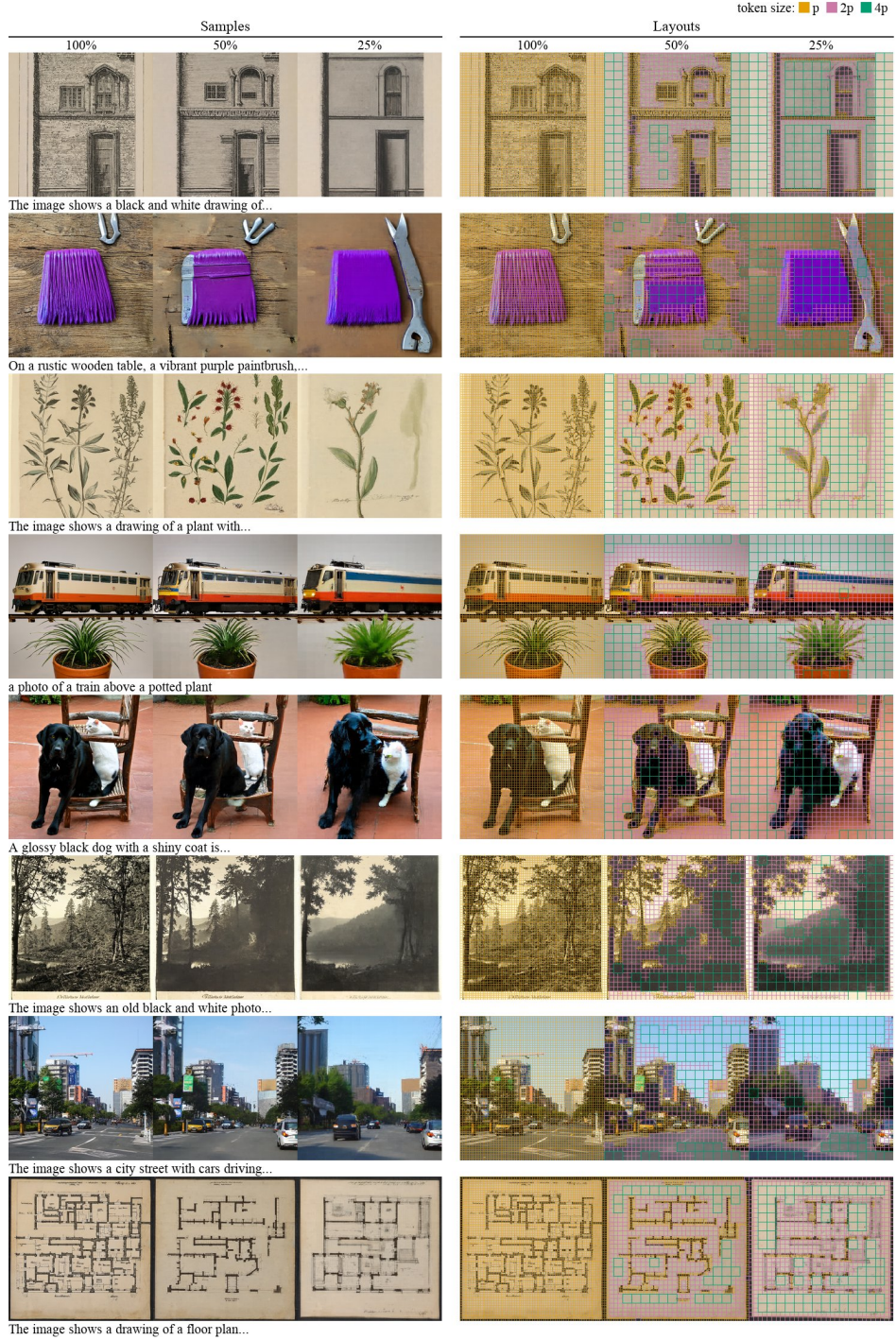}%
\caption{\textbf{\methodName{} on PixelDiT-T2I at $1024^2$: samples and their layouts, one prompt per row.}}
\label{fig:app_layout_pixeldit_1}
\end{figure}

\begin{figure}[H]
\centering
\input{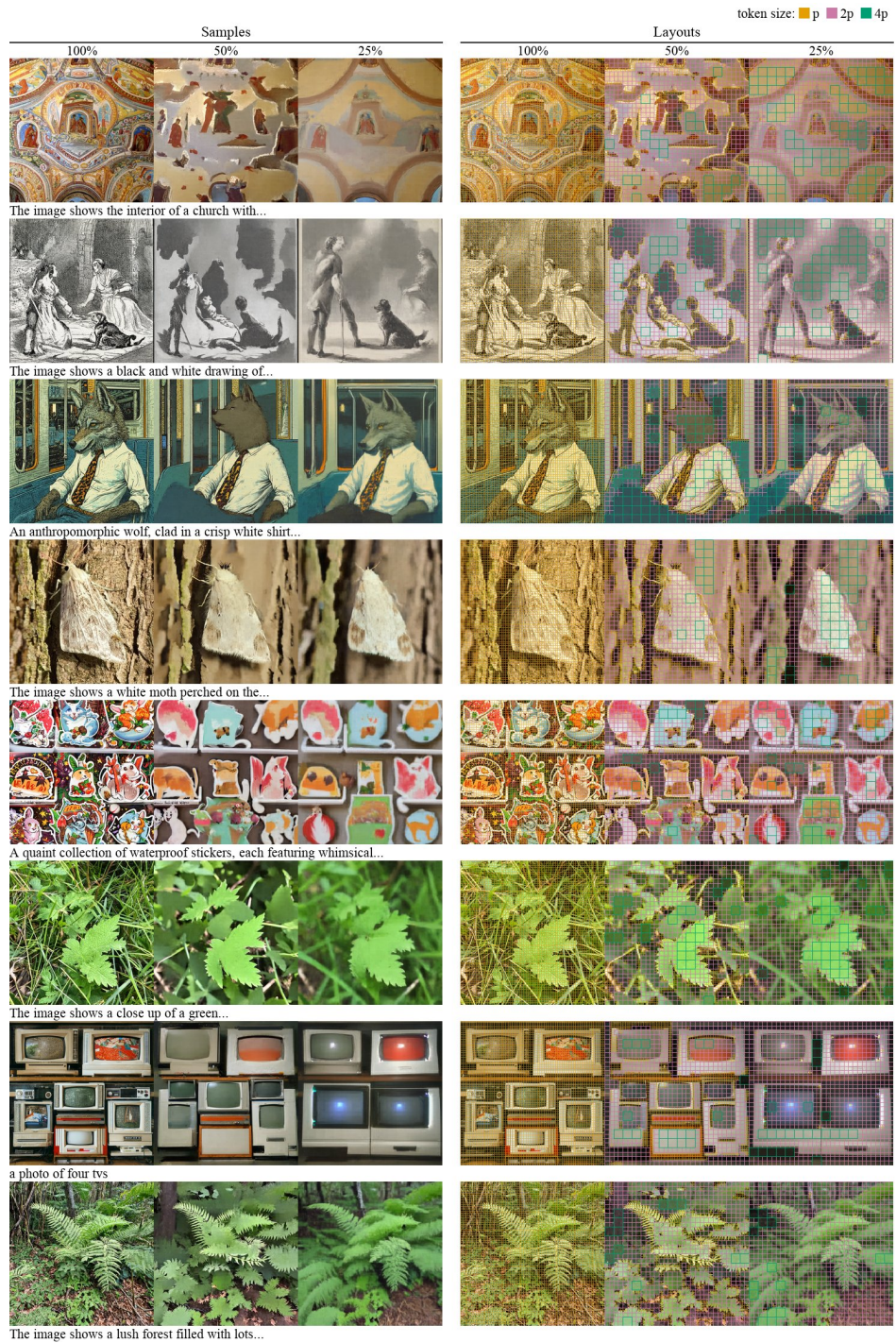}%
\caption{\textbf{\methodName{} on PixelDiT-T2I at $1024^2$: samples and their layouts, one prompt per row (continued).}}
\label{fig:app_layout_pixeldit_2}
\end{figure}

Figures~\ref{fig:app_layout_pixeldit_1} and~\ref{fig:app_layout_pixeldit_2} show additional qualitative results with their layouts on PixelDiT-T2I at $1024^2$. Each row is one prompt; the columns are the samples at 100, 50 and 25\,\% of the grid, then the same images with their layouts drawn over them (each token outlined in the colour of its size).

\clearpage
\section{Results in full}
\label{app:results}
\label{app:prelim}

The tables and figures behind Section~\ref{sec:analysis}, one subsection per result.

\subsection{Class-conditional generation (JiT)}
\label{app:jit}

\begin{figure}[H]
\centering
\includegraphics[width=\linewidth]{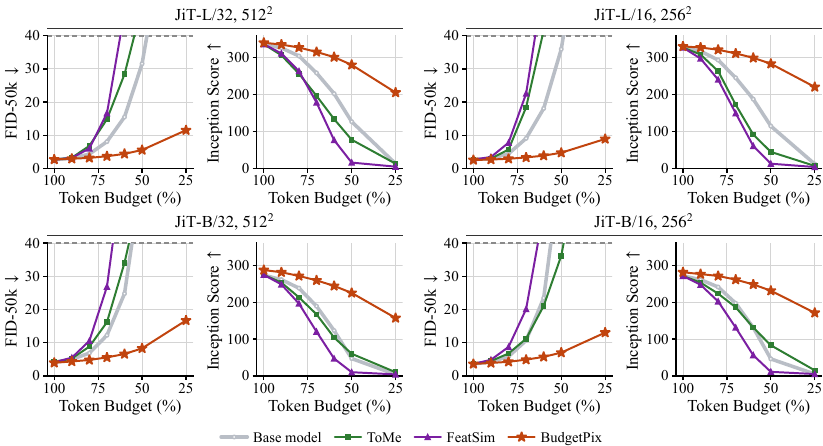}
\caption{\textbf{Class-conditional generation on ImageNet, JiT-L and JiT-B.}}
\label{fig:jit_all}
\end{figure}

FID-50k and Inception score (IS) on ImageNet (see App.~\ref{app:protocol}) for the four JiT parents, JiT-L above
JiT-B: Table~\ref{tab:main_jit} gives every value, and Figure~\ref{fig:jit_all} plots them with FID cut at 40.

\begin{table}[H]
\centering\scriptsize\setlength{\tabcolsep}{1.5pt}
\caption{\textbf{Class-conditional generation on ImageNet, JiT-L and JiT-B: FID-50k and Inception score.}}
\label{tab:main_jit}
\begin{tabular*}{\linewidth}{@{\extracolsep{\fill}}ll rrrrrrr rrrrrrr@{}}
\toprule
& & \multicolumn{7}{c}{FID-50k $\downarrow$} & \multicolumn{7}{c}{Inception score $\uparrow$} \\
\cmidrule(lr){3-9}\cmidrule(l){10-16}
parent & method & 100\% & 90\% & 80\% & 70\% & 60\% & 50\% & 25\% & 100\% & 90\% & 80\% & 70\% & 60\% & 50\% & 25\% \\
\midrule
JiT-L/32, $512^2$ & Base model & \textbf{2.66} & 3.07 & 4.34 & 8.09 & 15.51 & 31.57 & 107.51 & 337 & 327 & 305 & 259 & 203 & 127 & 13 \\
 & ToMe & \textbf{2.66} & 3.25 & 6.82 & 14.83 & 28.40 & 49.16 & 129.69 & 337 & 307 & 257 & 197 & 134 & 78 & 13 \\
 & FeatSim & \textbf{2.66} & 3.20 & 6.04 & 16.62 & 47.19 & 110.60 & 172.31 & 337 & 312 & 264 & 179 & 77 & 16 & 5 \\
 & \methodName{} & 2.71 & \textbf{2.94} & \textbf{3.20} & \textbf{3.69} & \textbf{4.41} & \textbf{5.57} & \textbf{11.54} & \textbf{341} & \textbf{336} & \textbf{328} & \textbf{316} & \textbf{302} & \textbf{281} & \textbf{206} \\
\midrule
JiT-L/16, $256^2$ & Base model & 2.71 & 3.01 & 4.44 & 9.07 & 18.09 & 35.95 & 116.16 & 328 & 318 & 294 & 246 & 188 & 115 & 12 \\
 & ToMe & 2.71 & 2.99 & 5.71 & 18.43 & 41.58 & 70.13 & 159.39 & 328 & 309 & 264 & 173 & 92 & 45 & 7 \\
 & FeatSim & 2.71 & 3.45 & 7.77 & 22.56 & 56.21 & 120.23 & 188.34 & 328 & 298 & 241 & 150 & 61 & 13 & 4 \\
 & \methodName{} & \textbf{2.56} & \textbf{2.70} & \textbf{2.93} & \textbf{3.33} & \textbf{3.87} & \textbf{4.75} & \textbf{8.90} & \textbf{331} & \textbf{328} & \textbf{322} & \textbf{312} & \textbf{300} & \textbf{284} & \textbf{221} \\
\midrule
JiT-B/32, $512^2$ & Base model & 4.12 & 5.00 & 6.86 & 12.24 & 24.79 & 60.61 & 190.71 & 273 & 261 & 238 & 189 & 123 & 47 & 4 \\
 & ToMe & 4.11 & 5.25 & 8.81 & 16.18 & 34.07 & 57.40 & 147.26 & 274 & 253 & 213 & 167 & 104 & 60 & 10 \\
 & FeatSim & 4.11 & 5.48 & 10.55 & 26.88 & 66.38 & 140.23 & 192.26 & 274 & 248 & 196 & 120 & 47 & 10 & 4 \\
 & \methodName{} & \textbf{3.94} & \textbf{4.35} & \textbf{4.80} & \textbf{5.57} & \textbf{6.59} & \textbf{8.30} & \textbf{16.71} & \textbf{287} & \textbf{281} & \textbf{271} & \textbf{259} & \textbf{244} & \textbf{225} & \textbf{157} \\
\midrule
JiT-B/16, $256^2$ & Base model & 3.61 & 4.24 & 5.84 & 10.67 & 23.46 & 63.94 & 145.46 & 271 & 261 & 241 & 197 & 131 & 46 & 5 \\
 & ToMe & \textbf{3.60} & 4.47 & 6.75 & 11.19 & 21.08 & 36.20 & 104.98 & 271 & 253 & 223 & 186 & 131 & 83 & 15 \\
 & FeatSim & \textbf{3.60} & 4.85 & 8.72 & 20.12 & 49.63 & 115.89 & 166.16 & 271 & 247 & 202 & 131 & 56 & 11 & 5 \\
 & \methodName{} & \textbf{3.60} & \textbf{3.89} & \textbf{4.25} & \textbf{4.89} & \textbf{5.73} & \textbf{7.01} & \textbf{13.05} & \textbf{281} & \textbf{276} & \textbf{271} & \textbf{261} & \textbf{248} & \textbf{231} & \textbf{171} \\
\bottomrule
\end{tabular*}

\end{table}

\clearpage
\subsection{Faithfulness to the dense image (JiT)}
\label{app:perc}

\begin{figure}[H]
\centering
\includegraphics[width=\linewidth]{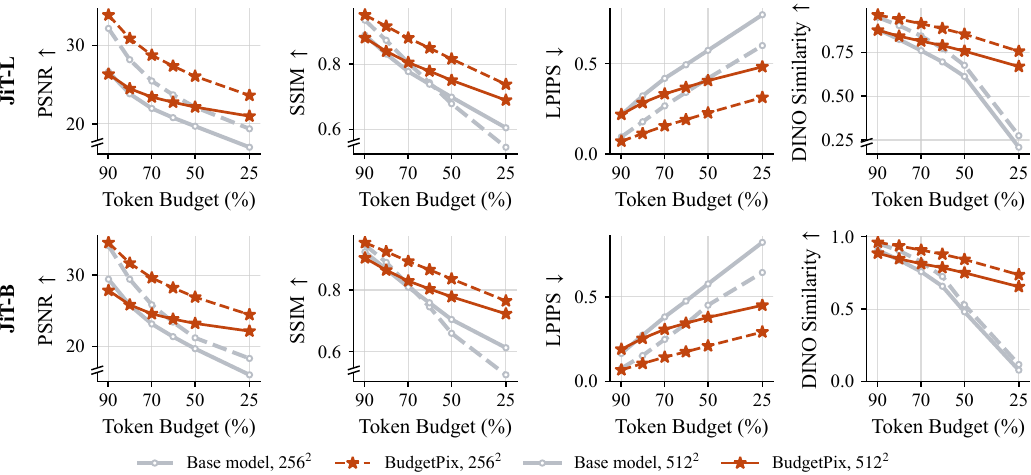}
\caption{\textbf{Paired metrics against the token budget, JiT-L and JiT-B.}}
\label{fig:perc}
\end{figure}

As in Figure~\ref{fig:t2i}(b), each image at a reduced budget is scored against the same model's image at the dense
grid, from the same label and noise (see App.~\ref{app:protocol}): Figure~\ref{fig:perc} for JiT-L and JiT-B, and
Table~\ref{tab:perc_full} for JiT-B, which adds CLIP similarity and splits PSNR between the pixels the layout merged and
those it kept. A break ($/\!/$) marks an axis that does not start at zero.

\begin{table}[H]
\centering\scriptsize\setlength{\tabcolsep}{1.5pt}
\caption{\textbf{Faithfulness to the dense image, JiT-B: paired metrics.}}
\label{tab:perc_full}
\begin{tabular*}{\linewidth}{@{\extracolsep{\fill}}ll rrrrrr rrrrrr@{}}
\toprule
& & \multicolumn{6}{c}{JiT-B/32, $512^2$} & \multicolumn{6}{c}{JiT-B/16, $256^2$} \\
\cmidrule(lr){3-8}\cmidrule(l){9-14}
metric & method & 90\% & 80\% & 70\% & 60\% & 50\% & 25\% & 90\% & 80\% & 70\% & 60\% & 50\% & 25\% \\
\midrule
PSNR $\uparrow$ & Base model & 29.4 & 25.8 & 23.1 & 21.4 & 19.7 & 16.0 & 34.1 & 29.4 & 25.8 & 23.4 & 21.2 & 18.3 \\
 & \methodName{} & 27.8 & 25.8 & 24.6 & 23.8 & 23.2 & 22.1 & 34.5 & 31.7 & 29.6 & 28.2 & 26.9 & 24.4 \\
\midrule
PSNR, merged pixels $\uparrow$ & Base model & 28.5 & 25.7 & 23.4 & 21.7 & 20.0 & 16.1 & 30.5 & 27.4 & 24.8 & 22.9 & 21.2 & 18.4 \\
 & \methodName{} & 28.6 & 26.8 & 25.5 & 24.6 & 23.9 & 22.3 & 32.0 & 30.1 & 28.6 & 27.6 & 26.7 & 24.5 \\
\midrule
PSNR, kept pixels $\uparrow$ & Base model & 30.1 & 26.4 & 23.5 & 21.5 & 19.6 & 15.2 & 35.9 & 31.2 & 27.4 & 24.6 & 21.7 & 17.3 \\
 & \methodName{} & 28.0 & 25.9 & 24.5 & 23.6 & 22.7 & 21.0 & 35.8 & 33.1 & 31.1 & 29.6 & 28.1 & 24.6 \\
\midrule
SSIM $\uparrow$ & Base model & 0.924 & 0.867 & 0.809 & 0.758 & 0.704 & 0.613 & 0.946 & 0.890 & 0.816 & 0.745 & 0.660 & 0.525 \\
 & \methodName{} & 0.904 & 0.864 & 0.829 & 0.803 & 0.778 & 0.723 & 0.954 & 0.924 & 0.893 & 0.866 & 0.836 & 0.764 \\
\midrule
LPIPS $\downarrow$ & Base model & 0.166 & 0.273 & 0.382 & 0.475 & 0.576 & 0.822 & 0.077 & 0.153 & 0.249 & 0.340 & 0.451 & 0.644 \\
 & \methodName{} & 0.190 & 0.254 & 0.308 & 0.345 & 0.379 & 0.450 & 0.068 & 0.106 & 0.144 & 0.175 & 0.210 & 0.292 \\
\midrule
CLIP similarity $\uparrow$ & Base model & 0.953 & 0.922 & 0.882 & 0.833 & 0.753 & 0.589 & 0.972 & 0.943 & 0.901 & 0.851 & 0.770 & 0.614 \\
 & \methodName{} & 0.947 & 0.931 & 0.917 & 0.905 & 0.893 & 0.863 & 0.977 & 0.966 & 0.953 & 0.940 & 0.925 & 0.884 \\
\midrule
DINOv2 similarity $\uparrow$ & Base model & 0.907 & 0.839 & 0.758 & 0.657 & 0.481 & 0.078 & 0.954 & 0.903 & 0.827 & 0.725 & 0.529 & 0.117 \\
 & \methodName{} & 0.885 & 0.847 & 0.813 & 0.786 & 0.750 & 0.654 & 0.960 & 0.936 & 0.907 & 0.879 & 0.843 & 0.737 \\
\bottomrule
\end{tabular*}
\end{table}

\clearpage
\subsection{Text-to-image generation (MiniT2I)}
\label{app:t2i}

Table~\ref{tab:main_t2i512} gives the values behind Figures~\ref{fig:t2i}(a)
and~\ref{fig:speedup_t2i_b} by token budget, for MiniT2I-B/16 and MiniT2I-L/16 at $512^2$. PixelDiT at $1024^2$ is
complete in Table~\ref{tab:main_1024}.

\begin{table}[H]
\centering\scriptsize\setlength{\tabcolsep}{1.5pt}
\caption{\textbf{Text-to-image generation, MiniT2I-B/16 and MiniT2I-L/16 at $512^2$.}}
\label{tab:main_t2i512}
\begin{tabular*}{\linewidth}{@{\extracolsep{\fill}}ll rrrrrrr rrrrrrr@{}}
\toprule
& & \multicolumn{7}{c}{MiniT2I-B/16} & \multicolumn{7}{c}{MiniT2I-L/16} \\
\cmidrule(lr){3-9}\cmidrule(l){10-16}
metric & method & 100\% & 90\% & 80\% & 70\% & 60\% & 50\% & 25\% & 100\% & 90\% & 80\% & 70\% & 60\% & 50\% & 25\% \\
\midrule
GenEval $\uparrow$ & Base model & 0.876 & \textbf{0.884} & 0.872 & 0.865 & 0.506 & 0.040 & 0.006 & 0.882 & 0.877 & 0.868 & 0.854 & 0.773 & 0.183 & 0.017 \\
 & ToMe & \textbf{0.877} & 0.878 & 0.867 & 0.861 & 0.841 & 0.779 & 0.283 & \textbf{0.886} & \textbf{0.891} & \textbf{0.886} & 0.873 & 0.845 & 0.741 & 0.229 \\
 & FeatSim & \textbf{0.877} & 0.882 & 0.871 & 0.865 & 0.830 & 0.610 & 0.106 & \textbf{0.886} & 0.880 & 0.850 & 0.798 & 0.676 & 0.292 & 0.007 \\
 & RTI & 0.850 & 0.864 & 0.870 & 0.869 & 0.870 & 0.868 & 0.856 & 0.863 & 0.873 & 0.867 & 0.870 & 0.868 & 0.871 & 0.859 \\
 & \methodName{} & \textbf{0.877} & \textbf{0.884} & \textbf{0.875} & \textbf{0.879} & \textbf{0.875} & \textbf{0.871} & \textbf{0.866} & 0.880 & 0.880 & 0.875 & \textbf{0.875} & \textbf{0.874} & \textbf{0.880} & \textbf{0.874} \\
\midrule
DPG-Bench $\uparrow$ & Base model & 84.2 & 83.7 & 83.7 & 82.8 & 74.3 & 23.2 & 6.8 & 84.7 & 84.4 & 83.6 & 83.0 & 80.6 & 54.1 & 8.0 \\
 & ToMe & 84.1 & \textbf{84.5} & 84.3 & 84.0 & 82.9 & 80.4 & 62.8 & 84.5 & \textbf{84.8} & 84.3 & 83.4 & 81.2 & 77.0 & 54.6 \\
 & FeatSim & 84.1 & 84.3 & 84.2 & 84.2 & 83.4 & 80.9 & 57.8 & 84.5 & 84.6 & 84.0 & 82.8 & 81.0 & 75.1 & 17.0 \\
 & RTI & 82.3 & 83.8 & 83.8 & 84.0 & 83.8 & 83.7 & 82.5 & 82.9 & 84.0 & 84.1 & 84.4 & 83.8 & 83.9 & 83.0 \\
 & \methodName{} & \textbf{84.3} & 84.3 & \textbf{84.5} & \textbf{84.4} & \textbf{84.1} & \textbf{84.4} & \textbf{83.8} & \textbf{84.8} & 84.7 & \textbf{84.7} & \textbf{84.6} & \textbf{84.7} & \textbf{84.7} & \textbf{84.3} \\
\midrule
CLIP $\uparrow$ & Base model & 28.21 & 28.12 & 28.03 & 27.50 & 23.39 & 17.86 & 15.99 & 28.88 & 28.81 & 28.59 & 28.30 & 27.25 & 22.17 & 16.20 \\
 & ToMe & \textbf{28.61} & \textbf{28.68} & 28.60 & 28.35 & 28.01 & 27.38 & 24.48 & \textbf{28.95} & \textbf{29.09} & \textbf{29.14} & 28.80 & 28.17 & 27.21 & 23.48 \\
 & FeatSim & \textbf{28.61} & 28.67 & \textbf{28.72} & \textbf{28.72} & \textbf{28.60} & 27.61 & 23.08 & \textbf{28.95} & 28.84 & 28.55 & 27.92 & 27.02 & 25.68 & 17.51 \\
 & RTI & 28.35 & 28.10 & 28.17 & 28.18 & 28.22 & \textbf{28.21} & \textbf{28.29} & 28.71 & 28.55 & 28.52 & 28.61 & 28.54 & 28.57 & 28.67 \\
 & \methodName{} & 28.11 & 28.12 & 28.07 & 28.09 & 28.06 & 28.06 & 27.95 & 28.87 & 28.91 & 28.89 & \textbf{28.92} & \textbf{28.96} & \textbf{29.03} & \textbf{29.27} \\
\midrule
PickScore $\uparrow$ & Base model & \textbf{22.77} & \textbf{22.72} & \textbf{22.63} & 22.29 & 20.34 & 18.66 & 18.28 & 22.82 & 22.72 & 22.54 & 22.28 & 21.53 & 19.41 & 18.11 \\
 & ToMe & 22.51 & 22.43 & 22.36 & 22.19 & 21.94 & 21.56 & 20.01 & \textbf{22.83} & 22.72 & 22.61 & 22.37 & 21.95 & 21.36 & 19.72 \\
 & FeatSim & 22.51 & 22.47 & 22.39 & 22.26 & 21.98 & 21.36 & 19.65 & \textbf{22.83} & 22.66 & 22.40 & 21.99 & 21.40 & 20.53 & 18.29 \\
 & RTI & 22.09 & 22.37 & 22.37 & 22.36 & 22.33 & 22.30 & \textbf{22.07} & 22.41 & 22.71 & 22.69 & 22.68 & 22.66 & \textbf{22.64} & \textbf{22.39} \\
 & \methodName{} & 22.50 & 22.49 & 22.46 & \textbf{22.43} & \textbf{22.38} & \textbf{22.31} & 22.00 & 22.82 & \textbf{22.81} & \textbf{22.79} & \textbf{22.76} & \textbf{22.72} & 22.63 & 22.33 \\
\midrule
ImageReward $\uparrow$ & Base model & \textbf{1.16} & 1.14 & 1.12 & 0.96 & -0.45 & -1.95 & -2.23 & \textbf{1.23} & 1.20 & 1.17 & 1.10 & 0.70 & -1.17 & -2.22 \\
 & ToMe & 1.15 & 1.14 & 1.12 & 1.06 & 0.97 & 0.77 & -0.38 & \textbf{1.23} & 1.21 & 1.19 & 1.11 & 0.96 & 0.67 & -0.67 \\
 & FeatSim & 1.15 & 1.15 & 1.12 & 1.08 & 0.99 & 0.67 & -0.91 & \textbf{1.23} & 1.20 & 1.15 & 1.04 & 0.80 & 0.06 & -2.17 \\
 & RTI & 1.03 & 1.12 & 1.12 & 1.12 & 1.11 & 1.10 & 1.07 & 1.14 & 1.21 & 1.21 & \textbf{1.21} & 1.20 & \textbf{1.20} & 1.16 \\
 & \methodName{} & \textbf{1.16} & \textbf{1.17} & \textbf{1.16} & \textbf{1.16} & \textbf{1.14} & \textbf{1.14} & \textbf{1.09} & 1.21 & \textbf{1.22} & \textbf{1.22} & \textbf{1.21} & \textbf{1.21} & \textbf{1.20} & \textbf{1.17} \\
\bottomrule
\end{tabular*}

\end{table}

\clearpage
\subsection{Wall-clock speed-up (every parent)}
\label{app:wallclock}

\begin{figure}[H]
\centering
\includegraphics[width=\linewidth]{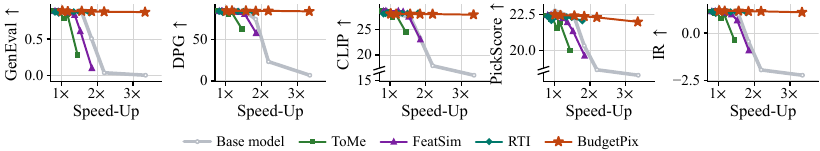}
\caption{\textbf{Five metrics against the measured wall-clock speed-up, MiniT2I-B/16 at $512^2$.}}
\label{fig:speedup_t2i_b}
\end{figure}

\begin{figure}[H]
\centering
\includegraphics[width=\linewidth]{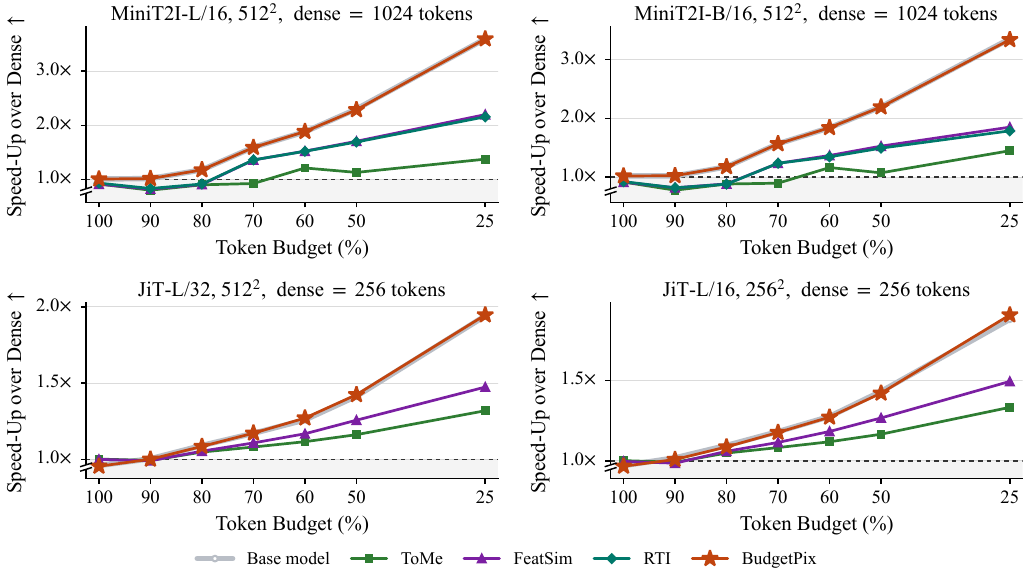}
\caption{\textbf{Wall-clock speed-up against the token budget, four parents.}}
\label{fig:timing}
\end{figure}

\begin{figure}[H]
\centering
\includegraphics[width=\linewidth]{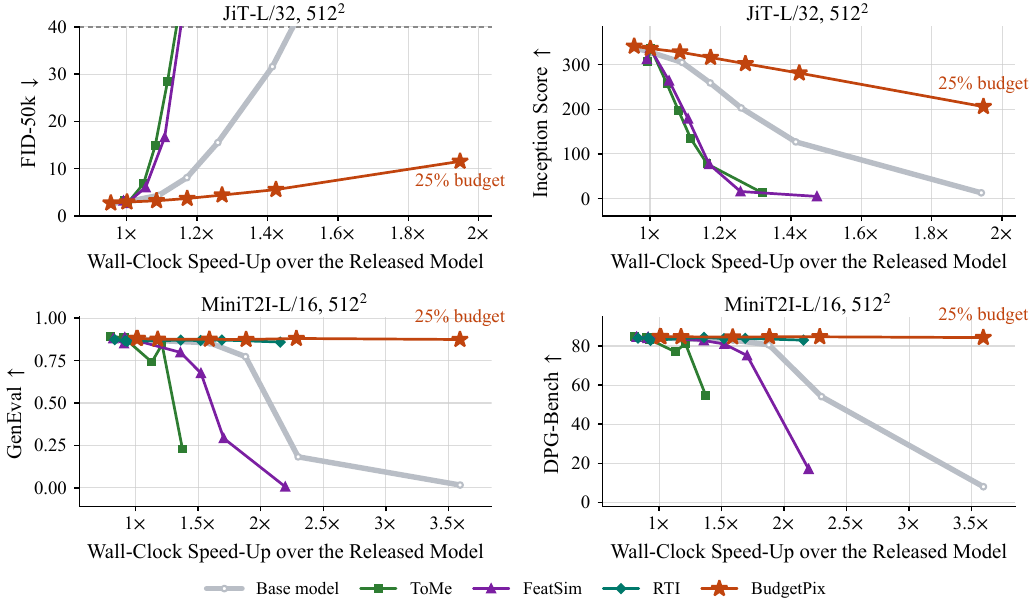}
\caption{\textbf{Quality against measured speed-up, JiT-L/32 and MiniT2I-L/16 at $512^2$.}}
\label{fig:pareto}
\end{figure}

 Speed-up is measured wall-clock time over the released model at its dense grid (see
App.~\ref{app:timing}). Figure~\ref{fig:speedup_t2i_b} is Figure~\ref{fig:t2i}(a) for MiniT2I-B/16 (DPG:
DPG-Bench; IR: ImageReward). Figure~\ref{fig:timing} (dashed: the released model, $1\times$) and
Table~\ref{tab:speedup} give the speed-up along the budget, Figure~\ref{fig:pareto} quality against it (FID cut at
40), and Tables~\ref{tab:time_minit2i_l_512}--\ref{tab:time_jit_b_256} the seconds per image and the speed-up of every method.
PixelDiT was timed for the base model and ours only. A break ($/\!/$) marks an axis that does not start at zero.

\begin{table}[H]
\centering\scriptsize\setlength{\tabcolsep}{1.5pt}
\caption{\textbf{Wall-clock speed-up over the released model, every parent.}}
\label{tab:speedup}
\begin{tabular*}{\linewidth}{@{\extracolsep{\fill}}ll rrrrrrr@{}}
\toprule
parent & method & 100\% & 90\% & 80\% & 70\% & 60\% & 50\% & 25\% \\
\midrule
MiniT2I-L/16, $512^2$ & Base model & 1.01$\times$ & 1.02$\times$ & 1.18$\times$ & 1.59$\times$ & 1.88$\times$ & 2.30$\times$ & 3.59$\times$ \\
 & ToMe & 0.91$\times$ & 0.80$\times$ & 0.90$\times$ & 0.92$\times$ & 1.21$\times$ & 1.13$\times$ & 1.37$\times$ \\
 & FeatSim & 0.91$\times$ & 0.82$\times$ & 0.91$\times$ & 1.36$\times$ & 1.52$\times$ & 1.70$\times$ & 2.20$\times$ \\
 & RTI & 0.93$\times$ & 0.83$\times$ & 0.92$\times$ & 1.36$\times$ & 1.52$\times$ & 1.69$\times$ & 2.16$\times$ \\
 & \methodName{} & 1.01$\times$ & 1.01$\times$ & 1.18$\times$ & 1.59$\times$ & 1.88$\times$ & 2.29$\times$ & 3.59$\times$ \\
\midrule
MiniT2I-B/16, $512^2$ & Base model & 1.02$\times$ & 1.02$\times$ & 1.17$\times$ & 1.57$\times$ & 1.84$\times$ & 2.19$\times$ & 3.35$\times$ \\
 & ToMe & 0.92$\times$ & 0.78$\times$ & 0.88$\times$ & 0.90$\times$ & 1.16$\times$ & 1.07$\times$ & 1.45$\times$ \\
 & FeatSim & 0.91$\times$ & 0.82$\times$ & 0.88$\times$ & 1.23$\times$ & 1.37$\times$ & 1.53$\times$ & 1.85$\times$ \\
 & RTI & 0.92$\times$ & 0.82$\times$ & 0.88$\times$ & 1.24$\times$ & 1.34$\times$ & 1.49$\times$ & 1.79$\times$ \\
 & \methodName{} & 1.02$\times$ & 1.02$\times$ & 1.18$\times$ & 1.56$\times$ & 1.84$\times$ & 2.19$\times$ & 3.34$\times$ \\
\midrule
PixelDiT-T2I, $1024^2$ & Base model & 0.99$\times$ & 0.79$\times$ & 0.88$\times$ & 0.99$\times$ & 1.12$\times$ & 1.29$\times$ & 1.94$\times$ \\
 & \methodName{} & 1.00$\times$ & 0.79$\times$ & 0.88$\times$ & 1.00$\times$ & 1.12$\times$ & 1.29$\times$ & 1.95$\times$ \\
\midrule
JiT-L/32, $512^2$ & Base model & 0.96$\times$ & 1.00$\times$ & 1.09$\times$ & 1.17$\times$ & 1.26$\times$ & 1.41$\times$ & 1.94$\times$ \\
 & ToMe & 1.00$\times$ & 0.99$\times$ & 1.05$\times$ & 1.08$\times$ & 1.12$\times$ & 1.16$\times$ & 1.32$\times$ \\
 & FeatSim & 1.00$\times$ & 0.99$\times$ & 1.05$\times$ & 1.11$\times$ & 1.17$\times$ & 1.26$\times$ & 1.47$\times$ \\
 & \methodName{} & 0.95$\times$ & 1.00$\times$ & 1.08$\times$ & 1.17$\times$ & 1.27$\times$ & 1.42$\times$ & 1.95$\times$ \\
\midrule
JiT-B/32, $512^2$ & Base model & 0.95$\times$ & 0.92$\times$ & 0.96$\times$ & 1.02$\times$ & 1.08$\times$ & 1.15$\times$ & 1.39$\times$ \\
 & ToMe & 1.00$\times$ & 0.92$\times$ & 0.96$\times$ & 0.99$\times$ & 1.02$\times$ & 1.07$\times$ & 1.20$\times$ \\
 & FeatSim & 1.01$\times$ & 0.97$\times$ & 1.01$\times$ & 1.05$\times$ & 1.09$\times$ & 1.15$\times$ & 1.28$\times$ \\
 & \methodName{} & 0.94$\times$ & 0.92$\times$ & 0.97$\times$ & 1.00$\times$ & 1.06$\times$ & 1.14$\times$ & 1.36$\times$ \\
\midrule
JiT-L/16, $256^2$ & Base model & 0.97$\times$ & 1.02$\times$ & 1.09$\times$ & 1.18$\times$ & 1.27$\times$ & 1.43$\times$ & 1.88$\times$ \\
 & ToMe & 1.00$\times$ & 0.99$\times$ & 1.05$\times$ & 1.08$\times$ & 1.12$\times$ & 1.17$\times$ & 1.33$\times$ \\
 & FeatSim & 1.00$\times$ & 0.98$\times$ & 1.06$\times$ & 1.11$\times$ & 1.18$\times$ & 1.27$\times$ & 1.49$\times$ \\
 & \methodName{} & 0.97$\times$ & 1.01$\times$ & 1.09$\times$ & 1.18$\times$ & 1.27$\times$ & 1.42$\times$ & 1.91$\times$ \\
\midrule
JiT-B/16, $256^2$ & Base model & 0.97$\times$ & 0.98$\times$ & 1.03$\times$ & 1.11$\times$ & 1.19$\times$ & 1.32$\times$ & 1.63$\times$ \\
 & ToMe & 0.99$\times$ & 0.90$\times$ & 0.96$\times$ & 0.98$\times$ & 1.04$\times$ & 1.06$\times$ & 1.21$\times$ \\
 & FeatSim & 1.00$\times$ & 0.96$\times$ & 1.01$\times$ & 1.04$\times$ & 1.09$\times$ & 1.16$\times$ & 1.28$\times$ \\
 & \methodName{} & 0.95$\times$ & 0.96$\times$ & 1.02$\times$ & 1.09$\times$ & 1.18$\times$ & 1.28$\times$ & 1.58$\times$ \\
\bottomrule
\end{tabular*}

\end{table}

\begin{table}[H]
\centering\scriptsize\setlength{\tabcolsep}{1.5pt}
\caption{\textbf{Wall-clock per image and speed-up, MiniT2I-L/16 at $512^2$.}}
\label{tab:time_minit2i_l_512}
\begin{tabular*}{\linewidth}{@{\extracolsep{\fill}}ll rrrrrrr@{}}
\toprule
measure & method & 100\% & 90\% & 80\% & 70\% & 60\% & 50\% & 25\% \\
\midrule
seconds per image & Base model & 4.912 & 4.870 & 4.205 & 3.105 & 2.634 & 2.151 & 1.376 \\
 & ToMe & 5.413 & 6.171 & 5.493 & 5.353 & 4.082 & 4.382 & 3.600 \\
 & FeatSim & 5.419 & 6.023 & 5.452 & 3.643 & 3.249 & 2.903 & 2.252 \\
 & RTI & 5.305 & 5.947 & 5.402 & 3.636 & 3.257 & 2.926 & 2.294 \\
 & \methodName{} & 4.913 & 4.872 & 4.206 & 3.113 & 2.626 & 2.164 & 1.376 \\
\midrule
speed-up & Base model & 1.01$\times$ & 1.02$\times$ & 1.18$\times$ & 1.59$\times$ & 1.88$\times$ & 2.30$\times$ & 3.59$\times$ \\
 & ToMe & 0.91$\times$ & 0.80$\times$ & 0.90$\times$ & 0.92$\times$ & 1.21$\times$ & 1.13$\times$ & 1.37$\times$ \\
 & FeatSim & 0.91$\times$ & 0.82$\times$ & 0.91$\times$ & 1.36$\times$ & 1.52$\times$ & 1.70$\times$ & 2.20$\times$ \\
 & RTI & 0.93$\times$ & 0.83$\times$ & 0.92$\times$ & 1.36$\times$ & 1.52$\times$ & 1.69$\times$ & 2.16$\times$ \\
 & \methodName{} & 1.01$\times$ & 1.01$\times$ & 1.18$\times$ & 1.59$\times$ & 1.88$\times$ & 2.29$\times$ & 3.59$\times$ \\
\bottomrule
\end{tabular*}
\end{table}

\begin{table}[H]
\centering\scriptsize\setlength{\tabcolsep}{1.5pt}
\caption{\textbf{Wall-clock per image and speed-up, MiniT2I-B/16 at $512^2$.}}
\label{tab:time_minit2i_b_512}
\begin{tabular*}{\linewidth}{@{\extracolsep{\fill}}ll rrrrrrr@{}}
\toprule
measure & method & 100\% & 90\% & 80\% & 70\% & 60\% & 50\% & 25\% \\
\midrule
seconds per image & Base model & 1.782 & 1.784 & 1.549 & 1.161 & 0.988 & 0.828 & 0.543 \\
 & ToMe & 1.986 & 2.340 & 2.061 & 2.029 & 1.564 & 1.695 & 1.255 \\
 & FeatSim & 1.988 & 2.222 & 2.057 & 1.472 & 1.329 & 1.191 & 0.983 \\
 & RTI & 1.970 & 2.223 & 2.067 & 1.471 & 1.353 & 1.218 & 1.018 \\
 & \methodName{} & 1.785 & 1.776 & 1.547 & 1.161 & 0.988 & 0.829 & 0.545 \\
\midrule
speed-up & Base model & 1.02$\times$ & 1.02$\times$ & 1.17$\times$ & 1.57$\times$ & 1.84$\times$ & 2.19$\times$ & 3.35$\times$ \\
 & ToMe & 0.92$\times$ & 0.78$\times$ & 0.88$\times$ & 0.90$\times$ & 1.16$\times$ & 1.07$\times$ & 1.45$\times$ \\
 & FeatSim & 0.91$\times$ & 0.82$\times$ & 0.88$\times$ & 1.23$\times$ & 1.37$\times$ & 1.53$\times$ & 1.85$\times$ \\
 & RTI & 0.92$\times$ & 0.82$\times$ & 0.88$\times$ & 1.24$\times$ & 1.34$\times$ & 1.49$\times$ & 1.79$\times$ \\
 & \methodName{} & 1.02$\times$ & 1.02$\times$ & 1.18$\times$ & 1.56$\times$ & 1.84$\times$ & 2.19$\times$ & 3.34$\times$ \\
\bottomrule
\end{tabular*}
\end{table}

\begin{table}[H]
\centering\scriptsize\setlength{\tabcolsep}{1.5pt}
\caption{\textbf{Wall-clock per image and speed-up, PixelDiT-T2I at $1024^2$.}}
\label{tab:time_pixeldit_1024}
\begin{tabular*}{\linewidth}{@{\extracolsep{\fill}}ll rrrrrrr@{}}
\toprule
measure & method & 100\% & 90\% & 80\% & 70\% & 60\% & 50\% & 25\% \\
\midrule
seconds per image & Base model & 2.992 & 3.760 & 3.383 & 2.983 & 2.652 & 2.306 & 1.527 \\
 & \methodName{} & 2.974 & 3.754 & 3.378 & 2.973 & 2.651 & 2.307 & 1.526 \\
\midrule
speed-up & Base model & 0.99$\times$ & 0.79$\times$ & 0.88$\times$ & 0.99$\times$ & 1.12$\times$ & 1.29$\times$ & 1.94$\times$ \\
 & \methodName{} & 1.00$\times$ & 0.79$\times$ & 0.88$\times$ & 1.00$\times$ & 1.12$\times$ & 1.29$\times$ & 1.95$\times$ \\
\bottomrule
\end{tabular*}
\end{table}

\begin{table}[H]
\centering\scriptsize\setlength{\tabcolsep}{1.5pt}
\caption{\textbf{Wall-clock per image and speed-up, JiT-L/32 at $512^2$.}}
\label{tab:time_jit_l_512}
\begin{tabular*}{\linewidth}{@{\extracolsep{\fill}}ll rrrrrrr@{}}
\toprule
measure & method & 100\% & 90\% & 80\% & 70\% & 60\% & 50\% & 25\% \\
\midrule
seconds per image & Base model & 0.394 & 0.376 & 0.346 & 0.323 & 0.300 & 0.267 & 0.195 \\
 & ToMe & 0.377 & 0.381 & 0.361 & 0.350 & 0.339 & 0.325 & 0.287 \\
 & FeatSim & 0.378 & 0.381 & 0.359 & 0.341 & 0.324 & 0.301 & 0.256 \\
 & \methodName{} & 0.396 & 0.377 & 0.348 & 0.322 & 0.297 & 0.265 & 0.194 \\
\midrule
speed-up & Base model & 0.96$\times$ & 1.00$\times$ & 1.09$\times$ & 1.17$\times$ & 1.26$\times$ & 1.41$\times$ & 1.94$\times$ \\
 & ToMe & 1.00$\times$ & 0.99$\times$ & 1.05$\times$ & 1.08$\times$ & 1.12$\times$ & 1.16$\times$ & 1.32$\times$ \\
 & FeatSim & 1.00$\times$ & 0.99$\times$ & 1.05$\times$ & 1.11$\times$ & 1.17$\times$ & 1.26$\times$ & 1.47$\times$ \\
 & \methodName{} & 0.95$\times$ & 1.00$\times$ & 1.08$\times$ & 1.17$\times$ & 1.27$\times$ & 1.42$\times$ & 1.95$\times$ \\
\bottomrule
\end{tabular*}
\end{table}

\begin{table}[H]
\centering\scriptsize\setlength{\tabcolsep}{1.5pt}
\caption{\textbf{Wall-clock per image and speed-up, JiT-B/32 at $512^2$.}}
\label{tab:time_jit_b_512}
\begin{tabular*}{\linewidth}{@{\extracolsep{\fill}}ll rrrrrrr@{}}
\toprule
measure & method & 100\% & 90\% & 80\% & 70\% & 60\% & 50\% & 25\% \\
\midrule
seconds per image & Base model & 0.131 & 0.134 & 0.129 & 0.122 & 0.115 & 0.108 & 0.089 \\
 & ToMe & 0.124 & 0.134 & 0.129 & 0.125 & 0.122 & 0.116 & 0.103 \\
 & FeatSim & 0.122 & 0.127 & 0.123 & 0.118 & 0.113 & 0.108 & 0.097 \\
 & \methodName{} & 0.132 & 0.135 & 0.128 & 0.123 & 0.116 & 0.109 & 0.091 \\
\midrule
speed-up & Base model & 0.95$\times$ & 0.92$\times$ & 0.96$\times$ & 1.02$\times$ & 1.08$\times$ & 1.15$\times$ & 1.39$\times$ \\
 & ToMe & 1.00$\times$ & 0.92$\times$ & 0.96$\times$ & 0.99$\times$ & 1.02$\times$ & 1.07$\times$ & 1.20$\times$ \\
 & FeatSim & 1.01$\times$ & 0.97$\times$ & 1.01$\times$ & 1.05$\times$ & 1.09$\times$ & 1.15$\times$ & 1.28$\times$ \\
 & \methodName{} & 0.94$\times$ & 0.92$\times$ & 0.97$\times$ & 1.00$\times$ & 1.06$\times$ & 1.14$\times$ & 1.36$\times$ \\
\bottomrule
\end{tabular*}
\end{table}

\begin{table}[H]
\centering\scriptsize\setlength{\tabcolsep}{1.5pt}
\caption{\textbf{Wall-clock per image and speed-up, JiT-L/16 at $256^2$.}}
\label{tab:time_jit_l_256}
\begin{tabular*}{\linewidth}{@{\extracolsep{\fill}}ll rrrrrrr@{}}
\toprule
measure & method & 100\% & 90\% & 80\% & 70\% & 60\% & 50\% & 25\% \\
\midrule
seconds per image & Base model & 0.383 & 0.365 & 0.339 & 0.314 & 0.291 & 0.259 & 0.197 \\
 & ToMe & 0.371 & 0.374 & 0.354 & 0.343 & 0.332 & 0.319 & 0.279 \\
 & FeatSim & 0.372 & 0.377 & 0.350 & 0.333 & 0.314 & 0.293 & 0.249 \\
 & \methodName{} & 0.384 & 0.368 & 0.342 & 0.315 & 0.292 & 0.261 & 0.195 \\
\midrule
speed-up & Base model & 0.97$\times$ & 1.02$\times$ & 1.09$\times$ & 1.18$\times$ & 1.27$\times$ & 1.43$\times$ & 1.88$\times$ \\
 & ToMe & 1.00$\times$ & 0.99$\times$ & 1.05$\times$ & 1.08$\times$ & 1.12$\times$ & 1.17$\times$ & 1.33$\times$ \\
 & FeatSim & 1.00$\times$ & 0.98$\times$ & 1.06$\times$ & 1.11$\times$ & 1.18$\times$ & 1.27$\times$ & 1.49$\times$ \\
 & \methodName{} & 0.97$\times$ & 1.01$\times$ & 1.09$\times$ & 1.18$\times$ & 1.27$\times$ & 1.42$\times$ & 1.91$\times$ \\
\bottomrule
\end{tabular*}
\end{table}

\begin{table}[H]
\centering\scriptsize\setlength{\tabcolsep}{1.5pt}
\caption{\textbf{Wall-clock per image and speed-up, JiT-B/16 at $256^2$.}}
\label{tab:time_jit_b_256}
\begin{tabular*}{\linewidth}{@{\extracolsep{\fill}}ll rrrrrrr@{}}
\toprule
measure & method & 100\% & 90\% & 80\% & 70\% & 60\% & 50\% & 25\% \\
\midrule
seconds per image & Base model & 0.121 & 0.119 & 0.113 & 0.106 & 0.098 & 0.089 & 0.072 \\
 & ToMe & 0.118 & 0.130 & 0.122 & 0.119 & 0.113 & 0.111 & 0.097 \\
 & FeatSim & 0.117 & 0.122 & 0.116 & 0.112 & 0.107 & 0.101 & 0.091 \\
 & \methodName{} & 0.123 & 0.122 & 0.115 & 0.108 & 0.099 & 0.091 & 0.074 \\
\midrule
speed-up & Base model & 0.97$\times$ & 0.98$\times$ & 1.03$\times$ & 1.11$\times$ & 1.19$\times$ & 1.32$\times$ & 1.63$\times$ \\
 & ToMe & 0.99$\times$ & 0.90$\times$ & 0.96$\times$ & 0.98$\times$ & 1.04$\times$ & 1.06$\times$ & 1.21$\times$ \\
 & FeatSim & 1.00$\times$ & 0.96$\times$ & 1.01$\times$ & 1.04$\times$ & 1.09$\times$ & 1.16$\times$ & 1.28$\times$ \\
 & \methodName{} & 0.95$\times$ & 0.96$\times$ & 1.02$\times$ & 1.09$\times$ & 1.18$\times$ & 1.28$\times$ & 1.58$\times$ \\
\bottomrule
\end{tabular*}
\end{table}

\clearpage
\subsection{Budget mode against threshold mode (JiT-L/32)}
\label{app:layoutstats}

\begin{figure}[H]
\centering
\includegraphics[width=0.38\linewidth]{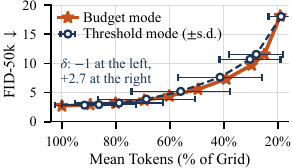}
\caption{\textbf{Budget mode against threshold mode, FID-50k on JiT-L/32 at $512^2$.}}
\label{fig:delta_mode}
\end{figure}

\begin{figure}[H]
\centering
\input{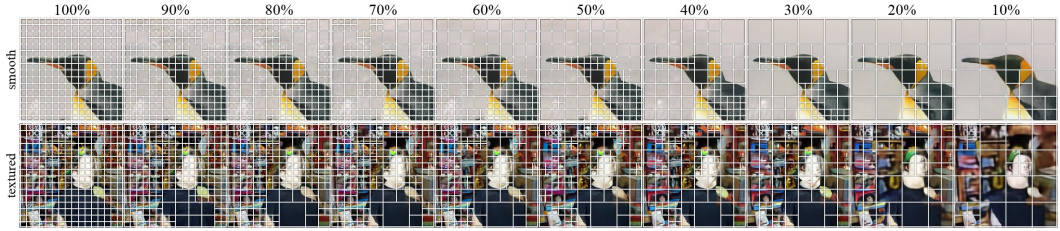}%
\caption{\textbf{The layout as the budget falls, JiT-L/32 at $512^2$, budget mode.}}
\label{fig:layout_budget}
\end{figure}

In budget mode $\delta$ is bisected so that every image gets at most the requested token count; in
threshold mode $\delta$ is fixed and each image takes the count its content gives. Threshold
mode sweeps $\delta$ from $-1$ to $+4$ on JiT-L/32 at $512^2$, with the images, noise and protocol of
Table~\ref{tab:main_jit}; budget mode is that table's points, with 40, 30, 20 and 10\,\% added.
Figure~\ref{fig:delta_mode} draws both down to a fifth of the grid, with bars of one standard deviation of the token
count across images, and Table~\ref{tab:delta_all} lists every point. Read at the same mean token count the two modes
lie on nearly one curve. What threshold mode gives up is the per-image guarantee, which is why every other number in
the paper is in budget mode; see App.~\ref{app:pool} for a pooled budget. Figure~\ref{fig:layout_budget} shows the layout as the budget falls for
a penguin before a plain wall (top) and a textured toyshop (bottom).

\begin{table}[H]
\centering\scriptsize\setlength{\tabcolsep}{1.5pt}
\caption{\textbf{Budget mode against threshold mode, JiT-L/32 at $512^2$: every point of Figure~\ref{fig:delta_mode}.}}
\label{tab:delta_all}
\begin{tabular*}{\linewidth}{@{\extracolsep{\fill}}l rrrrrrrrrrr@{}}
\toprule
budget mode, budget & 100\% & 90\% & 80\% & 70\% & 60\% & 50\% & 40\% & 30\% & 25\% & 20\% & 10\% \\
\midrule
tokens (\% of grid) & 100.0 & 89.5 & 80.1 & 69.5 & 60.2 & 49.6 & 39.1 & 29.7 & 25.0 & 19.1 & 9.8 \\
FID-50k $\downarrow$ & 2.71 & 2.94 & 3.20 & 3.69 & 4.41 & 5.57 & 7.33 & 9.73 & 11.54 & 18.12 & 44.62 \\
Inception score $\uparrow$ & 341 & 336 & 328 & 316 & 302 & 281 & 256 & 228 & 206 & 164 & 76 \\
\midrule
threshold mode, $\delta$ & $-$1 & $-$0.5 & 0 & $+$0.5 & $+$1 & $+$1.5 & $+$1.9 & $+$2 & $+$2.7 & $+$3 & $+$4 \\
\midrule
tokens (\% of grid) & 91.6 & 86.3 & 78.8 & 68.7 & 56.0 & 41.4 & 30.3 & 28.0 & 18.8 & 17.1 & 9.2 \\
s.d.\ across images (points) & 10.9 & 13.7 & 16.2 & 18.0 & 18.2 & 15.6 & 11.3 & 10.0 & 4.7 & 4.9 & 3.5 \\
FID-50k $\downarrow$ & 2.85 & 2.95 & 3.21 & 3.88 & 5.21 & 7.63 & 10.71 & 11.55 & 18.04 & 21.33 & 47.13 \\
Inception score $\uparrow$ & 337 & 334 & 322 & 308 & 285 & 251 & 215 & 206 & 158 & 140 & 64 \\
\bottomrule
\end{tabular*}
\end{table}

\clearpage
\subsection{One-step generation (pixel MeanFlow)}
\label{app:pmf}

Table~\ref{tab:pmf_full} completes Figure~\ref{fig:pmf_main}; see
App.~\ref{app:ours_pmf} for samples. The parent is the one-step pixel MeanFlow model~\citep{lu2026onestep} pMF-L/16,
class-conditional on ImageNet at $256^2$ (256 tokens; token scales of 16, 32 and 64 pixels), fine-tuned with its own
objective (see App.~\ref{app:settings}). One step leaves no trajectory to read a layout from,
so each image takes the entropy quadtree of one of 20 training images of its class, picked by the image index and
bisected to the budget; guidance is the released model's own at every budget. FID-50k follows pMF, 50 images per class against JiT's reference statistics at $256^2$ (the
released L/16 reads 2.54 here and 2.52 in its paper). The base model is the released weights on our layouts, without
training; ToMe and FeatSim are the JiT implementations ported to pMF.

\begin{table}[H]
\centering\scriptsize\setlength{\tabcolsep}{1.5pt}
\caption{\textbf{One-step generation, pMF-L/16 at $256^2$: FID-50k and Inception score.}}
\label{tab:pmf_full}
\begin{tabular*}{\linewidth}{@{\extracolsep{\fill}}llrrrrrrrrr@{}}
\toprule
metric & method & 100\% & 90\% & 80\% & 70\% & 60\% & 50\% & 40\% & 30\% & 25\% \\
& \textit{tokens} & \textit{256} & \textit{230} & \textit{205} & \textit{179} & \textit{154} & \textit{128} & \textit{102} & \textit{77} & \textit{64} \\
\midrule
FID-50k $\downarrow$ & Base model & \textbf{2.54} & 6.91 & 118.76 & 242.62 & 257.20 & 244.95 & 236.82 & 253.49 & 233.64 \\
 & ToMe & \textbf{2.54} & 3.06 & 5.93 & 14.66 & 36.04 & 70.62 & 109.40 & 122.59 & 149.41 \\
 & FeatSim & \textbf{2.54} & 5.71 & 20.44 & 57.37 & 100.26 & 134.35 & 180.26 & 194.43 & 192.94 \\
 & \methodName{} & 2.65 & \textbf{2.65} & \textbf{2.77} & \textbf{3.06} & \textbf{3.46} & \textbf{3.94} & \textbf{4.50} & \textbf{4.96} & \textbf{5.44} \\
\midrule
Inception score $\uparrow$ & Base model & 263 & 226 & 20 & 2 & 2 & 2 & 3 & 3 & 3 \\
 & ToMe & 263 & 240 & 203 & 150 & 84 & 36 & 14 & 11 & 6 \\
 & FeatSim & 263 & 209 & 123 & 49 & 20 & 11 & 4 & 3 & 3 \\
 & \methodName{} & \textbf{271} & \textbf{264} & \textbf{257} & \textbf{251} & \textbf{241} & \textbf{238} & \textbf{232} & \textbf{231} & \textbf{224} \\
\bottomrule
\end{tabular*}

\end{table}

\clearpage
\section{Details of the method and the experiments}
\label{app:setup}
\label{app:checks}

Details that Sections~\ref{sec:method} and~\ref{sec:setup} leave out.

\subsection{Architecture of the added modules}
\label{app:arch}

Table~\ref{tab:arch} lists how each family embeds and decodes a coarse token of side $np$ ($n = 2, 4$; $p$ is the
released patch and $D$ the backbone width), as built in the code; Sections~\ref{sec:encoder} and~\ref{sec:decoder}
describe the modules. Base tokens ($n = 1$) use the released embedding and head unchanged, and each coarse scale has its
own Patch Aggregator. PixelDiT's released pixel stage decodes every pixel at every budget, which sets its cost floor, so
instead of a pixel-space refiner, CellRefine corrects each cell's conditioning of that stage.

\begin{table}[H]
\centering\scriptsize\setlength{\tabcolsep}{3pt}
\caption{\textbf{Architecture of the added modules, every family.}}
\label{tab:arch}
\begin{tabular}{>{\raggedright\arraybackslash}p{0.16\linewidth}>{\centering\arraybackslash}p{0.19\linewidth}>{\centering\arraybackslash}p{0.19\linewidth}>{\centering\arraybackslash}p{0.19\linewidth}>{\centering\arraybackslash}p{0.19\linewidth}}
\toprule
 & JiT & pMF-L/16 & MiniT2I & PixelDiT-T2I \\
 & \textit{$256^2$ and $512^2$} & \textit{$256^2$} & \textit{$512^2$} & \textit{$1024^2$} \\
\midrule
width $D$ & 768 (B), 1024 (L) & 1024 & 768 (B), 1248 (L) & 1536 \\
cell embedding (released) & \multicolumn{3}{>{\centering\arraybackslash}p{\dimexpr 0.57\linewidth+4\tabcolsep\relax}}{\spanned{$p\times p$ convolution to 128 channels, then $1\times1$ convolution to $D$}} & linear map of the flattened $16\times16\times3$ cell \\
\midrule
downscale path & released embedding of the patch resized to $p\times p$ & \multicolumn{3}{>{\centering\arraybackslash}p{\dimexpr 0.57\linewidth+4\tabcolsep\relax}}{\spanned{none}} \\
split path: scale mixer & $n^2$ cell embeddings $+$ learned positions; $1\times1$ conv to 256, GELU, depthwise $n\times n$ conv, $1\times1$ conv, GELU, linear to $D$ & \multicolumn{3}{>{\centering\arraybackslash}p{\dimexpr 0.57\linewidth+4\tabcolsep\relax}}{\spanned{$n^2$ cell embeddings $+$ learned positions; $1\times1$ conv to 256, depthwise $n\times n$ conv, linear to $D$; no activation}} \\
coarse token & downscale path $+$ scale mixer $+$ scale embedding & \multicolumn{3}{>{\centering\arraybackslash}p{\dimexpr 0.57\linewidth+4\tabcolsep\relax}}{\spanned{mean of the $n^2$ cell embeddings $+$ scale mixer}} \\
position & sin--cos embedding and RoPE at the token centre & mean of the covered cells' learned positions; RoPE at the token centre & sin--cos embedding and RoPE at the token centre & RoPE at the token centre \\
\midrule
upscale path & \multicolumn{3}{>{\centering\arraybackslash}p{\dimexpr 0.57\linewidth+4\tabcolsep\relax}}{\spanned{released head's $p\times p$ prediction, bilinearly upsampled to $np\times np$}} & the token's state copied to its $n^2$ cells \\
refiner tokens & $4\times4$-pixel pieces of the $p\times p$ prediction: 16 ($p = 16$) or 64 ($p = 32$) & \multicolumn{2}{>{\centering\arraybackslash}p{\dimexpr 0.38\linewidth+2\tabcolsep\relax}}{\spanned{16 pieces of $4\times4$ pixels}} & one learned slot per cell (4 or 16) \\
refiner blocks & \multicolumn{4}{>{\centering\arraybackslash}p{\dimexpr 0.76\linewidth+6\tabcolsep\relax}}{\spanned{2 transformer blocks, width 64, 4 heads, adaLN on the token's final hidden state only}} \\
refiner output & \multicolumn{3}{>{\centering\arraybackslash}p{\dimexpr 0.57\linewidth+4\tabcolsep\relax}}{\spanned{linear to $4\cdot4\cdot3\,n^2$ values per piece, pixel shuffle by $n$: a residual on the upscale path}} & linear to a correction of each cell's 1536-d conditioning, then the released pixel stage \\
one refiner per & coarse scale & coarse scale and branch ($u$; $v$ in training) & coarse scale & coarse scale \\
\midrule
zero-initialised & last layer of the mixer, scale embedding, refiner adaLN and output & \multicolumn{3}{>{\centering\arraybackslash}p{\dimexpr 0.57\linewidth+4\tabcolsep\relax}}{\spanned{last layer of the mixer, refiner adaLN and output}} \\
added parameters & 2.39M (B/16), 2.40M (B/32), 3.06M (L/16, L/32) & 4.77M & 2.26M (B), 3.50M (L) & 4.37M \\
\bottomrule
\end{tabular}
\end{table}

\clearpage
\subsection{Training objective}
\label{app:loss}

The fine-tuning objective of Section~\ref{sec:loss}, $w(t)\lVert x - \hat{x}\rVert_2^2$, keeps the weighting each
parent was pretrained with, so no comparison gains from a change of loss. The parents regress velocity; with
$z_t = t\,x + (1-t)\,\boldsymbol{\varepsilon}$, the target and predicted velocities are $(x - z_t)$ and
$(\hat{x} - z_t)$ over the same factor, so
\begin{equation}
\mathcal{L}(\theta) = \mathbb{E}\Big\lVert
\frac{x - z_t}{\max(1 - t,\, t_\epsilon)} - \frac{\hat{x} - z_t}{\max(1 - t,\, t_\epsilon)}
\Big\rVert_2^2
= \mathbb{E}\Big[\frac{1}{\max(1 - t,\, t_\epsilon)^2}\,\lVert x - \hat{x}\rVert_2^2\Big],
\end{equation}
and $w(t) = \max(1 - t,\, t_\epsilon)^{-2}$, where each parent's own clip $t_\epsilon$ keeps the weight finite as
$t \to 1$. PixelDiT regresses $\boldsymbol{\varepsilon} - x$ at noise level $\sigma = 1 - t$: the same objective with
$w = \sigma^{-2}$, and its own sampling of $\sigma$ (logit-normal, shift 4.0).

\clearpage
\subsection{Zero-initialisation check}
\label{app:zeroinit}

Every added path starts at zero, so training starts from the released model
(Section~\ref{sec:impl}). Table~\ref{tab:sk_zeroinit} checks this on the seven parents: max $|\Delta|$ is the largest
absolute difference from the released model's output on the same inputs, and FID is dense-grid FID-50k on ImageNet for
JiT and held-out FID for MiniT2I and PixelDiT (see App.~\ref{app:protocol}).

\begin{table}[H]
\centering\scriptsize\setlength{\tabcolsep}{1.5pt}
\caption{\textbf{Zero-initialisation check, every parent.}}
\label{tab:sk_zeroinit}
\begin{tabular*}{\linewidth}{@{\extracolsep{\fill}}l rrrrrrr@{}}
\toprule
quantity & JiT-B/16 & JiT-L/16 & JiT-B/32 & JiT-L/32 & MiniT2I-B/16 & MiniT2I-L/16 & PixelDiT-T2I \\
& \textit{$256^2$} & \textit{$256^2$} & \textit{$512^2$} & \textit{$512^2$} & \textit{$512^2$} & \textit{$512^2$} & \textit{$1024^2$} \\
\midrule
max $|\Delta|$ at initialisation & $1.9\times10^{-6}$ & $1.6\times10^{-6}$ & $2.5\times10^{-6}$ & $1.9\times10^{-6}$ & $2.0\times10^{-6}$ & $4.5\times10^{-6}$ & $9.6\times10^{-4}$ \\
\midrule
FID, released model & 3.61 & 2.71 & 4.12 & 2.66 & 24.6 & 24.8 & 43.7 \\
FID, adapted model at step 0 & 3.61 & 2.71 & 4.12 & 2.66 & 24.6 & 24.8 & 43.7 \\
FID, adapted model after training & 3.60 & 2.56 & 3.94 & 2.71 & 17.1 & 22.9 & 36.2 \\
\bottomrule
\end{tabular*}
\end{table}

\clearpage
\subsection{Fine-tuning and sampling settings}
\label{app:settings}

Tables~\ref{tab:hp_jit} and~\ref{tab:hp_t2i} give each parent's fine-tuning recipe, read from its run, and the sampler
behind its benchmark scores. Every parent starts from its released checkpoint
(for JiT, its EMA weights), all weights are fine-tuned, and one checkpoint per parent serves every budget. MiniT2I-L's
120K mix is BLIP3o-60k (58,859 images), DALL-E~3 (19,016) and ShareGPT-4o-Image (41,355), with 1,024 held out. The
held-out FIDs use 50 steps (see App.~\ref{app:protocol}), and the wall-clock runs the settings of
Table~\ref{tab:time_conditions}.

\paragraph{Shared by the seven parents of Section~\ref{sec:impl}.} \emph{Layout:} the entropy quadtree on a 512-bin luma
histogram (entropy in $[0, 9]$ bits), with $\tau_{2p} = 6.0$ and $\tau_{4p} = 4.0$ bits; the offset
$\delta \sim \mathcal{U}(0, 2)$ is drawn per image and step; $a = 0.15$ of the images train on the dense grid, and
$b = 0.10$ of the rest (8.5\,\% of all images) on the random layout, whose cell entropies are drawn uniformly in
$[0, 9]$ bits (Section~\ref{sec:layoutdist}). No image augmentation: a centre crop and a resize fix the geometry.
\emph{Optimisation:} AdamW without weight decay, a linear warm-up then a constant learning rate, the added modules at
$8\times$ the backbone rate, EMA 0.9999 in fp32 (the evaluated weights), bf16 autocast with fp32 master weights, seed 0,
four GPUs. \emph{Sampling:} the first $n_w$ steps run on the dense grid, and the layout is recomputed
every $m = 5$ steps.

\begin{table}[H]
\centering\scriptsize\setlength{\tabcolsep}{3pt}
\caption{\textbf{Fine-tuning and sampling settings, JiT parents (ImageNet-1k).}}
\label{tab:hp_jit}
\begin{tabular}{>{\raggedright\arraybackslash}p{0.22\linewidth}>{\centering\arraybackslash}p{0.17\linewidth}>{\centering\arraybackslash}p{0.17\linewidth}>{\centering\arraybackslash}p{0.17\linewidth}>{\centering\arraybackslash}p{0.17\linewidth}}
\toprule
setting & JiT-B/16 & JiT-B/32 & JiT-L/16 & JiT-L/32 \\
 & \textit{$256^2$} & \textit{$512^2$} & \textit{$256^2$} & \textit{$512^2$} \\
\midrule
patch / token scales & 16 / 16, 32, 64 & 32 / 32, 64, 128 & 16 / 16, 32, 64 & 32 / 32, 64, 128 \\
parameters, base + added & 131.3M + 2.39M & 133.4M + 2.40M & 459.1M + 3.06M & 461.8M + 3.06M \\
\midrule
steps / epochs / images seen & \multicolumn{2}{>{\centering\arraybackslash}p{\dimexpr 0.34\linewidth+2\tabcolsep\relax}}{\spanned{142,987 / 100 / 128.1M}} & \multicolumn{2}{>{\centering\arraybackslash}p{\dimexpr 0.34\linewidth+2\tabcolsep\relax}}{\spanned{71,494 / 50 / 64.1M}} \\
global batch (per GPU $\times$ GPUs $\times$ accum.) & \multicolumn{3}{>{\centering\arraybackslash}p{\dimexpr 0.51\linewidth+4\tabcolsep\relax}}{\spanned{896 ($112 \times 4 \times 2$)}} & 896 ($56 \times 4 \times 4$) \\
learning rate, backbone / added & \multicolumn{4}{>{\centering\arraybackslash}p{\dimexpr 0.68\linewidth+6\tabcolsep\relax}}{\spanned{$3.5\times10^{-5}$ / $2.8\times10^{-4}$}} \\
learning-rate warm-up & \multicolumn{4}{>{\centering\arraybackslash}p{\dimexpr 0.68\linewidth+6\tabcolsep\relax}}{\spanned{1k steps}} \\
optimiser & \multicolumn{4}{>{\centering\arraybackslash}p{\dimexpr 0.68\linewidth+6\tabcolsep\relax}}{\spanned{AdamW $(0.9, 0.95)$, clip 3.0}} \\
noise scale & 1.0 & 2.0 & 1.0 & 2.0 \\
timestep $t$ & \multicolumn{4}{>{\centering\arraybackslash}p{\dimexpr 0.68\linewidth+6\tabcolsep\relax}}{\spanned{logit-normal$(-0.8, 0.8)$}} \\
label dropout & \multicolumn{4}{>{\centering\arraybackslash}p{\dimexpr 0.68\linewidth+6\tabcolsep\relax}}{\spanned{0.1}} \\
\midrule
sampler, guidance & \multicolumn{4}{>{\centering\arraybackslash}p{\dimexpr 0.68\linewidth+6\tabcolsep\relax}}{\spanned{50 Heun, 3.0 on $t \in [0.1, 1]$}} \\
dense warm-up steps $n_w$ & 10 & 5 & 10 & 5 \\
FID samples & \multicolumn{4}{>{\centering\arraybackslash}p{\dimexpr 0.68\linewidth+6\tabcolsep\relax}}{\spanned{50k, class-balanced, against training-set statistics}} \\
\bottomrule
\end{tabular}
\end{table}

\begin{table}[H]
\centering\scriptsize\setlength{\tabcolsep}{3pt}
\caption{\textbf{Fine-tuning and sampling settings, text-to-image parents.}}
\label{tab:hp_t2i}
\begin{tabular}{>{\raggedright\arraybackslash}p{0.22\linewidth}>{\centering\arraybackslash}p{0.22\linewidth}>{\centering\arraybackslash}p{0.22\linewidth}>{\centering\arraybackslash}p{0.24\linewidth}}
\toprule
setting & MiniT2I-B/16 & MiniT2I-L/16 & PixelDiT-T2I \\
 & \multicolumn{2}{>{\centering\arraybackslash}p{\dimexpr 0.44\linewidth+2\tabcolsep\relax}}{\spanned{\textit{$512^2$}}} & \textit{$1024^2$} \\
\midrule
patch / token scales & \multicolumn{3}{>{\centering\arraybackslash}p{\dimexpr 0.68\linewidth+4\tabcolsep\relax}}{\spanned{16 / 16, 32, 64}} \\
parameters, base + added & 258.1M + 2.26M & 911.8M + 3.50M & 1302.5M + 4.37M \\
training data & CC12M, LLaVA recaptions, 1M images & CC12M (1M), then the 120K mix & PD12M, 200K images \\
\midrule
steps / images seen & 100k / 25.6M & 40k (24k CC12M, then 16k mix) / 10.2M & 30k / 0.96M \\
global batch (per GPU $\times$ GPUs $\times$ accum.) & 256 ($16 \times 4 \times 4$) & 256 ($4 \times 4 \times 16$) & 32 ($4 \times 4 \times 2$) \\
learning rate, backbone / added & \multicolumn{2}{>{\centering\arraybackslash}p{\dimexpr 0.44\linewidth+2\tabcolsep\relax}}{\spanned{$10^{-5}$ / $8\times10^{-5}$}} & $2\times10^{-5}$ / $1.6\times10^{-4}$ \\
learning-rate warm-up & \multicolumn{2}{>{\centering\arraybackslash}p{\dimexpr 0.44\linewidth+2\tabcolsep\relax}}{\spanned{1k steps}} & 2k steps \\
optimiser & \multicolumn{2}{>{\centering\arraybackslash}p{\dimexpr 0.44\linewidth+2\tabcolsep\relax}}{\spanned{AdamW $(0.9, 0.95)$, $\epsilon\,10^{-8}$, clip 3.0}} & AdamW $(0.9, 0.999)$, $\epsilon\,10^{-15}$, clip 0.1 \\
diffusion law & \multicolumn{2}{>{\centering\arraybackslash}p{\dimexpr 0.44\linewidth+2\tabcolsep\relax}}{\spanned{$x$-prediction, noise scale 2.0, $t \sim$ logit-normal$(-0.8, 0.8)$}} & flow matching, velocity target, shift 4.0, $\sigma \sim$ logit-normal$(0, 1)$ \\
caption dropout & \multicolumn{2}{>{\centering\arraybackslash}p{\dimexpr 0.44\linewidth+2\tabcolsep\relax}}{\spanned{0.1, attention mask zeroed}} & 0.1, null embedding \\
text encoder (frozen) & \multicolumn{2}{>{\centering\arraybackslash}p{\dimexpr 0.44\linewidth+2\tabcolsep\relax}}{\spanned{flan-T5-large, 256 tokens}} & Gemma-2-2B-it, 300 tokens \\
memory & \multicolumn{2}{>{\centering\arraybackslash}p{\dimexpr 0.44\linewidth+2\tabcolsep\relax}}{\spanned{none}} & gradient checkpointing, EMA on CPU \\
\midrule
sampler, guidance & \multicolumn{2}{>{\centering\arraybackslash}p{\dimexpr 0.44\linewidth+2\tabcolsep\relax}}{\spanned{100 Euler, 5.0}} & 25 DPM-Solver++(2M)~\citep{lu2025dpm}, 4.5 \\
dense warm-up steps $n_w$ & \multicolumn{3}{>{\centering\arraybackslash}p{\dimexpr 0.68\linewidth+4\tabcolsep\relax}}{\spanned{5}} \\
held-out set & \multicolumn{2}{>{\centering\arraybackslash}p{\dimexpr 0.44\linewidth+2\tabcolsep\relax}}{\spanned{5k CC12M prompts}} & 2k PD12M prompts \\
\bottomrule
\end{tabular}
\end{table}

\paragraph{Pixel MeanFlow.} pMF-L/16 (see App.~\ref{app:pmf}) is fine-tuned with pMF's own objective, ported
from its code (the MeanFlow identity with its forward-mode derivative, $x$-prediction with the guidance in the target, the
auxiliary $v$-loss, and LPIPS and ConvNeXt-V2 perceptual terms for $t < 0.8$, in pMF's time, where $t = 1$ is pure
noise), with one layout per image held fixed across the step's network calls. Training layouts come from the
clean image: the dense grid with probability 0.30, otherwise the entropy quadtree at
$\delta \sim \mathcal{U}(0, 2)$, with random entropy maps for 10\,\% of those (7\,\% of all images). A merged token is the mean of its
base-patch embeddings plus a zero-initialised mixer, and each coarse scale has a zero-initialised pixel refiner on each output branch ($u$, and $v$ in training; 4.77M
added parameters in all), so training starts from the released model. AdamW $(0.9, 0.95)$ without weight decay,
learning rate $10^{-5}$ ($8\times$ on the added modules), 500 warm-up steps then constant, gradient clipping at 20, EMA
0.9995. It trains for 64{,}000 steps at a global batch of 91 (7 GPUs $\times$ 13; 52 on 4 GPUs for steps
40{,}000--58{,}000), 5.1M images, with the learning rate cooled linearly to zero over the last 8{,}000. Sampling keeps the
released guidance, 7.0 on $t \in [0.2, 0.7]$.

\clearpage
\subsection{Evaluation protocol}
\label{app:protocol}
\label{app:fid50k}

\paragraph{Methods compared.} The \emph{released model} is the parent as shipped, at its dense grid; the \emph{base
model} is the released weights run at the same budget through our sampler, without training. ToMe merges tokens by
bipartite soft matching on the image keys before RoPE, averaged over heads (size-weighted, with proportional attention and
the same $r$ in every block; condition tokens, text or JiT's in-context class tokens, are never merged). FeatSim makes one similarity merge down to $B$ tokens early in the network (block 3 on
MiniT2I). Both wrap the released weights, keep the sampler on the dense grid, and on MiniT2I write the merged tokens
back to the dense grid for the last two blocks. RTI runs its released adapter (40k steps), which exists for MiniT2I-B
and MiniT2I-L only. On PixelDiT ours is compared with the base model. Every method of a parent shares the prompts or
labels, the noise, the solver, the steps, the guidance and the attention arithmetic.

\paragraph{Class-conditional.} FID and Inception score follow JiT: Inception-v3 from torch-fidelity on uint8 images at
native resolution, against JiT's ImageNet-1k~\citep{deng2009imagenet} training-set statistics at $256^2$ and $512^2$;
the Inception score uses 10 splits. FID-50k uses 50 images per class; the ablations (see App.~\ref{app:extra}) use
FID-10k, 10 per class. Each FID is a single run.

\paragraph{Text-to-image.} GenEval (553 prompts, 4 images each) counts an image as correct only if every element of its
prompt is rendered, averaged over six tasks; DPG-Bench (1,065 prompts, 4 images each) is the mean mPLUG-large answer;
CLIP score (ViT-L/14), PickScore and ImageReward are means over the 1,632 PartiPrompts, one image each. Held-out FID, used at the dense grid, compares 5k
generated with 5k held-out real images for CC12M~\citep{changpinyo2021conceptual} and 2k with 2k for PD12M, at native
resolution, with 50 steps at guidance 3.0 (MiniT2I-B), 5.0 (MiniT2I-L) and 2.75 (PixelDiT).

\paragraph{Paired metrics.} Each image at a reduced budget is scored against the same method's image at the dense
grid, from the same prompt or label and noise, by PSNR, SSIM, LPIPS and the cosine similarity of CLIP and DINOv2
embeddings. On MiniT2I-L each cell is the mean over the 553 GenEval prompts, one image each, with its 95\,\% interval;
on JiT it is the mean over 2,000 class-balanced images at guidance 3.0 and 50 steps, and PSNR is also split between the
pixels the layout merged into coarse tokens and those it kept.

\clearpage
\subsection{User study and VLM-as-Judge}
\label{app:vlm}

\paragraph{Images.} Both studies use the images of Figure~\ref{fig:t2i}(b): MiniT2I-L/16 at $512^2$, five methods at
ten budgets, and the released model. Image $j$ comes from prompt $j$ and noise seeded by $j$, so two images in one
comparison differ by method and budget alone. The prompts are 100 GenEval prompts drawn at random (22 colours, 21
two-object, 20 colour-attribution, 14 position, 13 single-object, 10 counting); the user study uses 20 of them, chosen
from the released model's images alone. Both studies are scored alike: ``no visible difference'' counts as 5\,\%, a tie
as one half, and 95\,\% intervals are bootstrapped over prompts. Table~\ref{tab:user_study} gives every number of
Figure~\ref{fig:vlm}, with the intervals in brackets.

\paragraph{User study.}\label{app:human} Nine anonymous online forms of 20 questions, one image per question, in
random order; each form counts its first 50 participants. The instructions ask for faithfulness only, not to prefer a
sharper or prettier image that has changed what the reference shows.

\begin{figure}[H]
\centering
\includegraphics[width=0.8\linewidth]{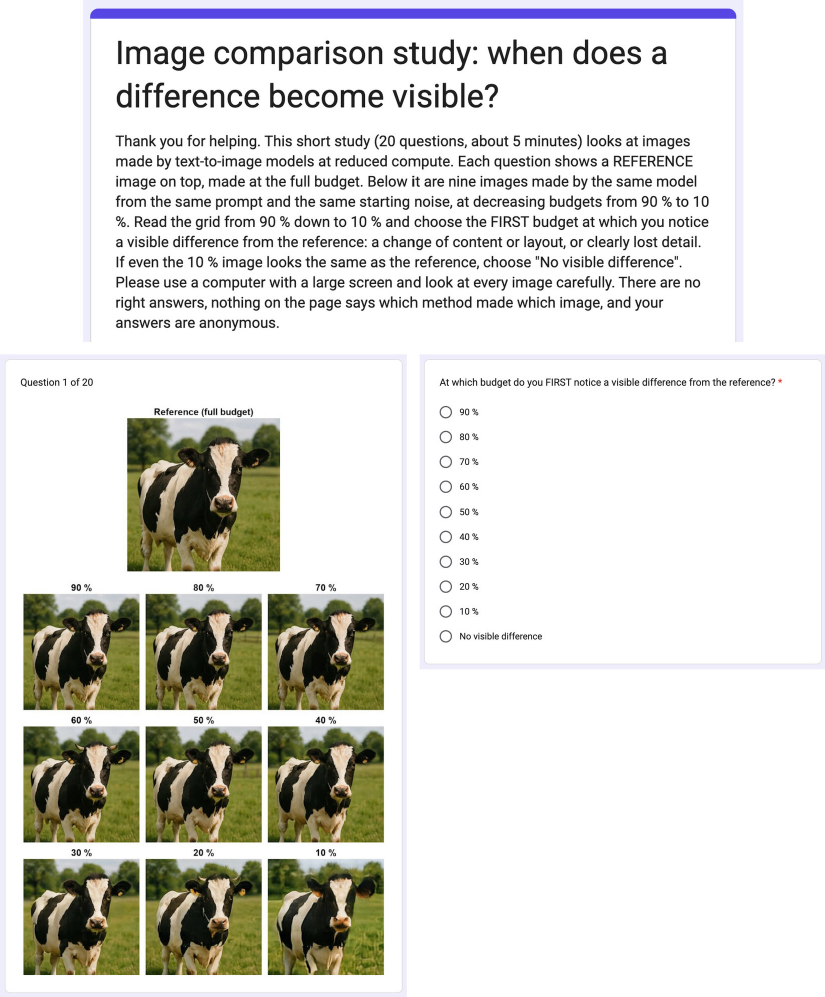}
\caption{\textbf{User-study form on where a difference is first seen: its start and first question.}}
\label{fig:form}
\end{figure}

\paragraph{User study: where the drop is first seen (Figure~\ref{fig:vlm}(a)).} Five forms, one per method, show the
method's own image at the dense grid above its budgets from 90 to 10\,\%, and ask for the first budget with a visible
difference. Figure~\ref{fig:form} shows the start of one of them and its first question.

\begin{figure}[H]
\centering
\includegraphics[width=0.8\linewidth]{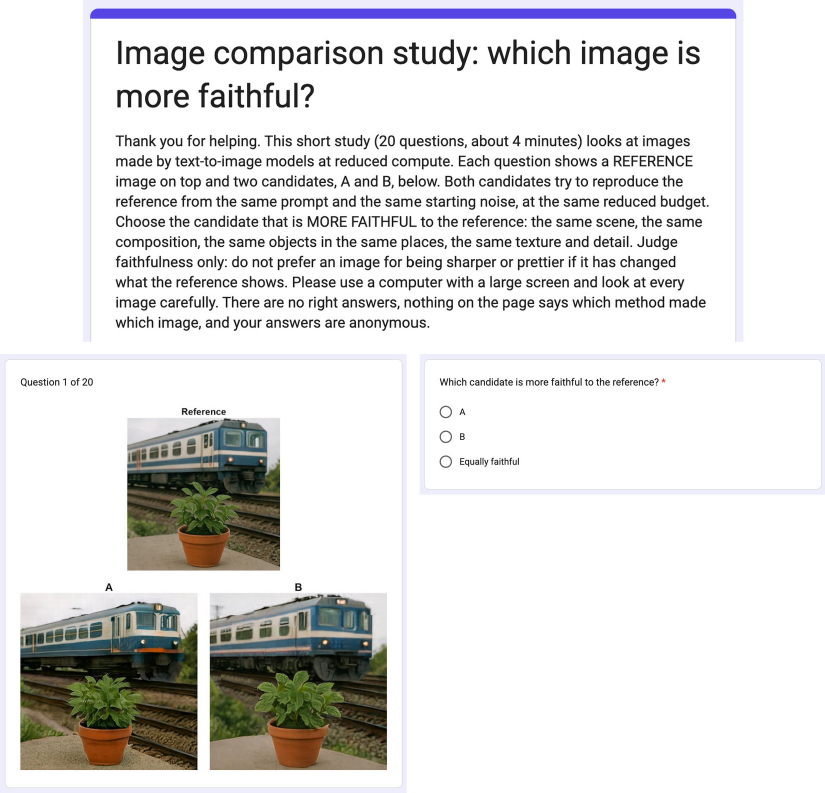}
\caption{\textbf{User-study form on which image is more faithful: its start and first question.}}
\label{fig:form_faithful}
\end{figure}

\paragraph{User study: which is more faithful (Figure~\ref{fig:vlm}(b)).} Four forms, one per budget (80, 60, 40 and
20\,\%), show the released image above ours and RTI as A and B, ours on the left in half the questions, and ask which is
more faithful, or ``equally faithful''. Figure~\ref{fig:form_faithful} shows the start of one of them and its first
question.

\paragraph{User study: reading the numbers.} Participants chose A in 48 to 54\,\% of the answers that picked a side, so
position did not drive them. Dropping the 17 who gave one answer to 18 or more of their 20 questions moves no cut by
more than 2.4 points and no share at all.

\paragraph{VLM-as-Judge.} The judge is Gemini~3.8~Flash at temperature 0, thinking level low, one image set per
call, answering with a single number or letter.

\paragraph{VLM-as-Judge: where the drop is first seen (Figure~\ref{fig:vlm}(a)).} The judge sees a method's own image
at the dense grid, then one picture tiling its ten budgets in order, after this text:
\begin{quote}\small
The first image is a REFERENCE, generated by this same method at its full token budget from the prompt: ``$\langle$\textit{prompt}$\rangle$''

The second image is a grid of ten images from the same prompt and the same starting noise, generated by the SAME method at decreasing token budgets. The budgets are printed above each tile and run from 100\% (top left) down to 10\% (bottom right), in order; the 100\% tile is the reference itself.

Reading the grid in order from 100\% downwards, at which budget do you FIRST notice a visible difference from the reference (a change of content, layout, or clearly lost detail)? If even the 10\% tile is indistinguishable from the reference, answer 0.

Reply with exactly one number from this list and nothing else: 100, 90, 80, 70, 60, 50, 40, 30, 20, 10, 0.
\end{quote}
With the method's own dense image as reference, the answer is what the budget costs, not what the method already
changed at 100\,\%. One call per method and prompt: 500
judgments.

\paragraph{VLM-as-Judge: which is more faithful (Figure~\ref{fig:vlm}(b)).} The judge sees the released image, then
ours and one baseline at the same budget, after this text:
\begin{quote}\small
The first image is a REFERENCE, generated by an image model at its full token budget from the prompt: ``$\langle$\textit{prompt}$\rangle$''

Two candidate images follow, A then B. Each was generated by a different method at a reduced token budget from the same prompt and the same starting noise, trying to reproduce the reference.

Which candidate is MORE FAITHFUL to the reference: the same scene, the same composition, the same objects in the same places, the same texture and detail? Judge faithfulness to the reference only. Do not reward an image for being sharper or more pleasing than the reference if it has changed what the reference shows.

Reply with exactly one character and nothing else: A, B, or T if the two are equally faithful.
\end{quote}
The three images are resized and JPEG-encoded identically. Each pair is judged in
both left/right orders, as independent calls: 7,193 judgments over four baselines, nine budgets and 100 prompts.

\paragraph{VLM-as-Judge: reading the numbers.} Position bias, the share of pairs answered on the same side in both
orders, is 8.7\,\%, below the 15\,\% rule of thumb. The judge never answered T.

\begin{table}[H]
\centering\scriptsize\setlength{\tabcolsep}{1.5pt}
\caption{\textbf{User study and VLM-as-Judge, MiniT2I-L/16 at $512^2$: the numbers of Figure~\ref{fig:vlm}.}}
\label{tab:user_study}
\begin{tabular*}{\linewidth}{@{\extracolsep{\fill}}lrrrrr@{}}
\toprule
cut before a difference is seen (\%) & Base model & ToMe & FeatSim & RTI & \methodName{} \\
\midrule
people, 20 prompts & 30.3 {\tiny[29.2, 31.4]} & 32.5 {\tiny[30.1, 35.1]} & 23.3 {\tiny[21.5, 25.3]} & 38.0 {\tiny[31.8, 44.0]} & \textbf{51.8} {\tiny[48.0, 55.7]} \\
judge, the same 20 prompts & 18.5 {\tiny[14.5, 22.5]} & 20.0 {\tiny[16.5, 24.0]} & 12.5 {\tiny[11.0, 14.5]} & 24.0 {\tiny[14.5, 35.0]} & \textbf{43.5} {\tiny[35.0, 51.0]} \\
judge, all 100 prompts & 15.4 {\tiny[14.1, 16.9]} & 15.9 {\tiny[14.3, 17.5]} & 11.7 {\tiny[11.0, 12.4]} & 21.1 {\tiny[16.9, 25.8]} & \textbf{36.2} {\tiny[31.9, 40.4]} \\
\bottomrule
\end{tabular*}

\vspace{5pt}

\begin{tabular*}{\linewidth}{@{\extracolsep{\fill}}lrrrr@{}}
\toprule
prefers ours to RTI (\%) & 80\% & 60\% & 40\% & 20\% \\
\midrule
people, 20 prompts & 63.5 {\tiny[52.7, 74.7]} & 63.4 {\tiny[52.2, 74.2]} & 57.2 {\tiny[45.3, 69.5]} & 62.3 {\tiny[52.8, 72.2]} \\
\quad answers for ours / RTI / equal & 54 / 27 / 19 & 56 / 29 / 16 & 50 / 36 / 14 & 54 / 29 / 17 \\
judge, the same 20 prompts & 77.5 {\tiny[60.0, 95.0]} & 72.5 {\tiny[55.0, 87.5]} & 50.0 {\tiny[30.0, 70.0]} & 75.0 {\tiny[57.5, 90.0]} \\
judge, all 100 prompts & 75.5 {\tiny[67.5, 83.0]} & 69.0 {\tiny[61.0, 76.5]} & 54.5 {\tiny[46.0, 63.0]} & 61.5 {\tiny[53.5, 69.5]} \\
\bottomrule
\end{tabular*}

\end{table}

\clearpage
\subsection{Wall-clock measurement}
\label{app:timing}

\paragraph{Timing.} Every cell was timed on one H100, with the overhead of the measurement rather than of the method
removed: per-forward GPU-to-CPU synchronisations, layout work repeated on every forward, and kernel-launch cost, with
one CUDA graph per layout. The images are bit-identical to those of each method's released code; ours and the base model
also resample merged tokens with two fp32 matrix products (identical images on MiniT2I, ${\sim}55$\,dB PSNR on JiT).
Speed-up is over the released model at its dense grid, timed the same way. Table~\ref{tab:time_conditions} lists the
conditions, which every method of a parent shares.

\paragraph{Two checks.} ToMe, FeatSim and RTI are not deterministic: two runs of their released code agree only to 20,
27 and 34\,dB PSNR, so for them the images match up to this noise. PixelDiT does not replay correctly as a
CUDA graph (23\,dB against its own output), so it is timed without graphs, with bit-identical images.


\begin{table}[H]
\centering\scriptsize\setlength{\tabcolsep}{3pt}
\caption{\textbf{Conditions of the wall-clock measurements, every parent.}}
\label{tab:time_conditions}
\begin{tabular}{>{\raggedright\arraybackslash}p{0.17\linewidth}>{\centering\arraybackslash}p{0.19\linewidth}>{\centering\arraybackslash}p{0.16\linewidth}>{\centering\arraybackslash}p{0.14\linewidth}>{\centering\arraybackslash}p{0.1\linewidth}>{\centering\arraybackslash}p{0.14\linewidth}}
\toprule
setting & JiT-B/32, L/32 ($512^2$) & JiT-B/16, L/16 ($256^2$) & MiniT2I-B/16 & MiniT2I-L/16 & PixelDiT-T2I \\
\midrule
resolution & $512\times512$ & $256\times256$ & \multicolumn{2}{>{\centering\arraybackslash}p{\dimexpr 0.24\linewidth+2\tabcolsep\relax}}{\spanned{$512\times512$}} & $1024\times1024$ \\
dense grid & 256 tokens (32~px) & 256 tokens (16~px) & \multicolumn{2}{>{\centering\arraybackslash}p{\dimexpr 0.24\linewidth+2\tabcolsep\relax}}{\spanned{1024 tokens (16~px)}} & 4096 tokens (16~px) \\
token scales of the layout & 32~/ 64~/ 128~px & \multicolumn{4}{>{\centering\arraybackslash}p{\dimexpr 0.54\linewidth+6\tabcolsep\relax}}{\spanned{16~/ 32~/ 64~px}} \\
budgets 100, 90, 80, 70, 60, 50, 25\,\% & \multicolumn{2}{>{\centering\arraybackslash}p{\dimexpr 0.35\linewidth+2\tabcolsep\relax}}{\spanned{256, 230, 205, 179, 154, 128, 64 tokens}} & \multicolumn{2}{>{\centering\arraybackslash}p{\dimexpr 0.24\linewidth+2\tabcolsep\relax}}{\spanned{1024, 922, 819, 717, 614, 512, 256 tokens}} & 4096, 3686, 3277, 2867, 2458, 2048, 1024 tokens \\
conditioning & \multicolumn{2}{>{\centering\arraybackslash}p{\dimexpr 0.35\linewidth+2\tabcolsep\relax}}{\spanned{class label, random}} & \multicolumn{3}{>{\centering\arraybackslash}p{\dimexpr 0.38\linewidth+4\tabcolsep\relax}}{\spanned{text, pre-encoded}} \\
batch (images per sampler call) & \multicolumn{2}{>{\centering\arraybackslash}p{\dimexpr 0.35\linewidth+2\tabcolsep\relax}}{\spanned{50}} & 32 & 20 & 8 \\
warm-up, discarded & \multicolumn{5}{>{\centering\arraybackslash}p{\dimexpr 0.73\linewidth+8\tabcolsep\relax}}{\spanned{one call per cell for the released model and \methodName{}, one per process for the others}} \\
timed calls per cell & released, \methodName{}, all of JiT-B/32: 3 (median); others: 2 & released, \methodName{}: 3 (median); others: 2 & \multicolumn{2}{>{\centering\arraybackslash}p{\dimexpr 0.24\linewidth+2\tabcolsep\relax}}{\spanned{released, \methodName{}: 2 (median); others: 1}} & released, \methodName{}: 2 (median); base model: 1 \\
solver, steps & \multicolumn{2}{>{\centering\arraybackslash}p{\dimexpr 0.35\linewidth+2\tabcolsep\relax}}{\spanned{Heun, 50 (last step Euler)}} & \multicolumn{2}{>{\centering\arraybackslash}p{\dimexpr 0.24\linewidth+2\tabcolsep\relax}}{\spanned{Euler, 100}} & DPM-Solver++(2M), 50 \\
guidance scale & \multicolumn{2}{>{\centering\arraybackslash}p{\dimexpr 0.35\linewidth+2\tabcolsep\relax}}{\spanned{3.0 on $t \in [0.1, 1]$}} & 3.0 & 5.0 & 4.5 \\
network forwards per step & \multicolumn{2}{>{\centering\arraybackslash}p{\dimexpr 0.35\linewidth+2\tabcolsep\relax}}{\spanned{4 (2 Heun stages $\times$ cond./uncond.)}} & \multicolumn{3}{>{\centering\arraybackslash}p{\dimexpr 0.38\linewidth+4\tabcolsep\relax}}{\spanned{2 (cond., uncond.)}} \\
dense warm-up steps $n_w$ & 5, +1 forward (21 of 199 forwards) & 10, +1 forward (41 of 199 forwards) & \multicolumn{2}{>{\centering\arraybackslash}p{\dimexpr 0.24\linewidth+2\tabcolsep\relax}}{\spanned{5, +1 forward for the $\hat{x}$ estimate}} & 5, +1 forward \\
layout & \multicolumn{5}{>{\centering\arraybackslash}p{\dimexpr 0.73\linewidth+8\tabcolsep\relax}}{\spanned{built once, then frozen}} \\
numerics & \multicolumn{2}{>{\centering\arraybackslash}p{\dimexpr 0.35\linewidth+2\tabcolsep\relax}}{\spanned{bf16 weights, bf16 autocast}} & \multicolumn{2}{>{\centering\arraybackslash}p{\dimexpr 0.24\linewidth+2\tabcolsep\relax}}{\spanned{bf16 weights, bf16 autocast; RTI in pure bf16, as it ships}} & bf16 weights, bf16 autocast \\
attention arithmetic (all methods) & \multicolumn{2}{>{\centering\arraybackslash}p{\dimexpr 0.35\linewidth+2\tabcolsep\relax}}{\spanned{the model's own}} & \multicolumn{2}{>{\centering\arraybackslash}p{\dimexpr 0.24\linewidth+2\tabcolsep\relax}}{\spanned{released: einsum + softmax}} & fused SDPA, masked when a reduced budget pads the layout \\
methods timed & \multicolumn{2}{>{\centering\arraybackslash}p{\dimexpr 0.35\linewidth+2\tabcolsep\relax}}{\spanned{released, base model, \methodName{}, ToMe, FeatSim}} & \multicolumn{2}{>{\centering\arraybackslash}p{\dimexpr 0.24\linewidth+2\tabcolsep\relax}}{\spanned{released, base model, \methodName{}, ToMe, FeatSim, RTI}} & released, base model, \methodName{} \\
\bottomrule
\end{tabular}
\end{table}

\clearpage
\section{Ablations on JiT-B}
\label{app:extra}
\label{app:ablations}

Each ablation changes one factor at a time on JiT-B, on the development model (see
App.~\ref{app:reference}--\ref{app:ceiling}) or on the final JiT-B/32 (see App.~\ref{app:sampler}--\ref{app:pool}). All
are FID-10k (see App.~\ref{app:protocol}), not comparable with the FID-50k of the main tables. The reported results use
none of the budget-tuned schedules or guidance scales found here. The component ablation is
Table~\ref{tab:abl_components}.

\subsection{Reference runs (development JiT-B)}
\label{app:reference}

Table~\ref{tab:three} is the reference for the development runs (see
App.~\ref{app:pixelstream}--\ref{app:guidance}): the 50-epoch model on JiT-B/32 with the default sampler (uniform
schedule, 50 steps, guidance 3.0). JiT-B/16 had no adapted model yet, so only its base model is shown.

\begin{table}[H]
\centering\scriptsize\setlength{\tabcolsep}{1.5pt}
\caption{\textbf{Reference runs, JiT-B: FID-10k.}}
\label{tab:three}
\begin{tabular*}{\linewidth}{@{\extracolsep{\fill}}ll rrrrrrrr@{}}
\toprule
parent & method & 100\% & 90\% & 80\% & 70\% & 60\% & 50\% & 40\% & 30\% \\
& \textit{tokens} & \textit{256} & \textit{230} & \textit{205} & \textit{179} & \textit{154} & \textit{128} & \textit{102} & \textit{77} \\
\midrule
JiT-B/32, $512^2$ & Base model & 6.53 & 7.44 & 9.45 & 14.96 & 27.61 & 63.91 & 118.04 & 156.99 \\
 & \methodName{}, 50 epochs & \textbf{6.47} & \textbf{6.93} & \textbf{7.36} & \textbf{8.33} & \textbf{9.36} & \textbf{11.18} & \textbf{13.89} & \textbf{17.33} \\
\midrule
JiT-B/16, $256^2$ & Base model & 6.02 & 6.67 & 8.30 & 13.26 & 26.34 & 67.38 & 116.44 & 152.40 \\
\bottomrule
\end{tabular*}
\end{table}

\subsection{A parallel pixel-stream decoder (development JiT-B/32)}
\label{app:pixelstream}

Before the refiner, the decoder was a pixel stream running in parallel with the token stream.
Table~\ref{tab:v5snap} follows it from 10 to 50 epochs.

\begin{table}[H]
\centering\scriptsize\setlength{\tabcolsep}{1.5pt}
\caption{\textbf{Pixel-stream decoder across training length, JiT-B/32 at $512^2$: FID-10k.}}
\label{tab:v5snap}
\begin{tabular*}{\linewidth}{@{\extracolsep{\fill}}ll rrrrrrrr@{}}
\toprule
parent & training & 100\% & 90\% & 80\% & 70\% & 60\% & 50\% & 40\% & 30\% \\
& \textit{tokens} & \textit{256} & \textit{230} & \textit{205} & \textit{179} & \textit{154} & \textit{128} & \textit{102} & \textit{77} \\
\midrule
JiT-B/32, $512^2$ & 10 epochs & 7.48 & 8.16 & 9.33 & 11.30 & 13.87 & 18.19 & 23.95 & 29.94 \\
 & 20 epochs & 7.45 & 8.15 & 9.10 & 10.91 & 13.08 & 16.27 & 21.80 & 28.29 \\
 & 30 epochs & 7.40 & 8.05 & 9.05 & 10.72 & 12.43 & 15.95 & 20.50 & 26.78 \\
 & 50 epochs & 7.33 & 8.01 & 8.94 & 10.36 & 12.15 & 15.17 & 19.41 & 24.96 \\
\bottomrule
\end{tabular*}
\end{table}

\subsection{Sampling schedule and step count (development JiT-B/32)}
\label{app:schedule}

Table~\ref{tab:sched} varies the timestep schedule and the step count at 128 tokens (half the grid),
guidance 6.0.

\begin{table}[H]
\centering\scriptsize\setlength{\tabcolsep}{1.5pt}
\caption{\textbf{Timestep schedule against step count, JiT-B/32 at $512^2$: FID-10k.}}
\label{tab:sched}
\begin{tabular*}{\linewidth}{@{\extracolsep{\fill}}ll rrrr@{}}
\toprule
parent & schedule & 10 steps & 15 steps & 25 steps & 50 steps \\
\midrule
JiT-B/32, $512^2$ & cosine & 10.02 & 9.15 & 9.01 & 8.82 \\
 & shift-3 & 11.71 & 11.21 & 10.61 & 10.10 \\
 & uniform & 12.14 & 11.88 & 10.42 & 9.10 \\
\bottomrule
\end{tabular*}
\end{table}

\subsection{Guidance scale along the budget (development JiT-B/32)}
\label{app:guidance}

Table~\ref{tab:cfg} sweeps guidance from 3.0 to 6.0 along the budget, with the cosine schedule at 50
steps; the last row is the default sampler (uniform schedule, guidance 3.0).

\begin{table}[H]
\centering\scriptsize\setlength{\tabcolsep}{1.5pt}
\caption{\textbf{Guidance scale along the budget, JiT-B/32 at $512^2$: FID-10k.}}
\label{tab:cfg}
\begin{tabular*}{\linewidth}{@{\extracolsep{\fill}}ll rrrrr@{}}
\toprule
parent & sampler & 90\% & 80\% & 70\% & 60\% & 40\% \\
& \textit{tokens} & \textit{230} & \textit{205} & \textit{179} & \textit{154} & \textit{102} \\
\midrule
JiT-B/32, $512^2$ & cosine, guidance 3.0 & 7.19 & 7.75 & 8.75 & 10.17 & 14.50 \\
 & cosine, guidance 4.0 & 7.21 & 7.31 & 7.62 & 8.37 & 11.14 \\
 & cosine, guidance 5.0 & 8.03 & 7.87 & 7.92 & 8.17 & 9.96 \\
 & cosine, guidance 6.0 & 8.93 & 8.55 & 8.46 & 8.45 & 9.66 \\
 & uniform, guidance 3.0 (default) & 6.93 & 7.36 & 8.33 & 9.36 & 13.89 \\
\bottomrule
\end{tabular*}
\end{table}

\subsection{Upper bound on a better decoder (development JiT-B/32)}
\label{app:ceiling}

Two arms change only the decoded output of the 50-epoch JiT-B/32 (Table~\ref{tab:three}).
Oracle-fine gives the coarse regions their ground-truth fine detail, an upper bound on any better decoder;
downsample-lanczos resamples every output to $256^2$ and back, removing all finer detail.

\begin{table}[H]
\centering\scriptsize\setlength{\tabcolsep}{1.5pt}
\caption{\textbf{Upper bound on a better decoder, JiT-B/32 at $512^2$: FID-10k.}}
\label{tab:ceiling}
\begin{tabular*}{\linewidth}{@{\extracolsep{\fill}}ll rrr@{}}
\toprule
parent & arm & 80\% & 50\% & 30\% \\
& \textit{tokens} & \textit{205} & \textit{128} & \textit{77} \\
& \textit{guidance} & \textit{3.0} & \textit{6.0} & \textit{6.0} \\
\midrule
JiT-B/32, $512^2$ & reference & 7.36 & 9.10 & 11.69 \\
 & downsample-lanczos & 7.42 & 8.92 & 11.14 \\
 & oracle-fine & 7.38 & 9.34 & 11.56 \\
\bottomrule
\end{tabular*}
\end{table}

\subsection{Layout and sampler settings (final JiT-B/32)}
\label{app:sampler}
\label{app:layoutrule}

Table~\ref{tab:sampler} varies one setting at a time on the final JiT-B/32 ($512^2$, 100 epochs),
from the sampler of every reported result (\emph{default}: 50 Heun steps, guidance 3.0, five dense warm-up steps, a
layout rebuilt every five steps), on the same 10k images and noise. Random placement is the entropy quadtree on
permuted entropy maps, which keeps the token count; the oracle is the quadtree of a 50-step dense draft from the same
noise. The noise-seed rows repeat the default with shifted noise.

Table~\ref{tab:sampler_gs} varies guidance and the timestep schedule the same way. Time shift 3.0 spends
more steps at high noise, the Karras schedule more at low noise, cosine more at both ends, and two-phase only 8 of the
50 at $t \ge 0.7$; 50 Euler steps make half the network evaluations of the others. 

\begin{table}[H]
\centering\scriptsize\setlength{\tabcolsep}{1.5pt}
\caption{\textbf{Layout and sampler settings, JiT-B/32 at $512^2$: FID-10k.}}
\label{tab:sampler}
\begin{tabular*}{\linewidth}{@{\extracolsep{\fill}}ll rr@{}}
\toprule
& & 50\% & 25\% \\
setting & value & \textit{128} & \textit{64} \\
\midrule
layout signal & entropy quadtree (default) & \textbf{10.99} & 19.79 \\
 & random placement & 19.51 & 23.41 \\
 & oracle, 50-step draft & 11.30 & \textbf{19.58} \\
\midrule
dense warm-up steps & 0 & 12.44 & 20.41 \\
 & 2 & 11.20 & 20.33 \\
 & 5 (default) & 10.99 & 19.79 \\
 & 10 & 10.36 & 18.09 \\
 & 20 & \textbf{9.68} & \textbf{15.73} \\
\midrule
re-layout interval & every step & \textbf{10.96} & 19.76 \\
 & every 5 steps (default) & 10.99 & 19.79 \\
 & every 10 steps & 11.00 & \textbf{19.70} \\
 & every 25 steps & 11.04 & 19.73 \\
 & once, after the warm-up & \textbf{10.96} & 19.76 \\
\midrule
Heun steps & 25 & \textbf{10.88} & \textbf{18.02} \\
 & 50 (default) & 10.99 & 19.79 \\
 & 100 & 10.92 & 19.92 \\
\midrule
noise seed & 0 (default) & 10.99 & 19.79 \\
 & 1 & 10.81 & 19.36 \\
 & 2 & 10.74 & 19.38 \\
\bottomrule
\end{tabular*}
\end{table}

\begin{table}[H]
\centering\scriptsize\setlength{\tabcolsep}{1.5pt}
\caption{\textbf{Guidance scale and timestep schedule, JiT-B/32 at $512^2$: FID-10k.}}
\label{tab:sampler_gs}
\begin{tabular*}{\linewidth}{@{\extracolsep{\fill}}ll rr@{}}
\toprule
 &  & 50\% & 25\% \\
setting & value & \textit{128} & \textit{64} \\
\midrule
guidance & 2.0 & 19.09 & 31.73 \\
 & 2.5 & 13.77 & 24.27 \\
 & 3.0 (default) & 10.99 & 19.79 \\
 & 4.0 & 8.96 & 15.13 \\
 & 5.0 & \textbf{8.67} & 13.50 \\
 & 6.0 & 8.91 & \textbf{13.20} \\
 & 7.0 & 9.30 & 13.67 \\
\midrule
schedule & uniform, Heun (default) & 10.99 & 19.79 \\
 & time shift 3.0 & 14.54 & 24.01 \\
 & Karras, $\rho = 7$ & 11.50 & \textbf{17.72} \\
 & cosine & 11.65 & 20.40 \\
 & two-phase & 11.13 & 19.48 \\
 & Euler, 100 steps & \textbf{10.72} & 19.01 \\
 & Euler, 50 steps & 11.03 & 18.46 \\
\bottomrule
\end{tabular*}
\end{table}

\subsection{Sensitivity to the entropy thresholds (final JiT-B/32)}
\label{app:thr}

The thresholds $\tau_{2p} = 6.0$ and $\tau_{4p} = 4.0$ bits are taken from
APT~\citep{choudhury2026faster}, not tuned. Table~\ref{tab:thr} shifts $\tau_{2p}$ by $+\kappa$ and $\tau_{4p}$ by
$-\kappa$ and then bisects $\delta$ to the budget, so the count stays near 128 tokens (124 at $\kappa = -1.5$) and mainly
the split between $2p$ and $4p$ tokens moves; a negative $\kappa$ favours $4p$. The model, sampler and images are those
of Table~\ref{tab:sampler}.

\begin{table}[H]
\centering\scriptsize\setlength{\tabcolsep}{1.5pt}
\caption{\textbf{Entropy-threshold skew, JiT-B/32 at $512^2$: FID-10k and Inception score at 128 tokens.}}
\label{tab:thr}
\begin{tabular*}{\linewidth}{@{\extracolsep{\fill}}l rrrrrrr@{}}
\toprule
skew $\kappa$ & $-1.5$ & $-1$ & $-0.5$ & 0 (default) & $+0.5$ & $+1$ & $+1.5$ \\
\midrule
FID-10k $\downarrow$ & 15.79 & 13.02 & 11.42 & 10.99 & \textbf{10.95} & \textbf{10.95} & 10.97 \\
Inception score $\uparrow$ & 152 & 165 & 173 & 175 & 176 & \textbf{177} & \textbf{177} \\
\bottomrule
\end{tabular*}
\end{table}

\subsection{Layout evolution during sampling (final JiT-B/32)}
\label{app:dynamics}

\begin{figure}[H]
\centering
\includegraphics[width=\linewidth]{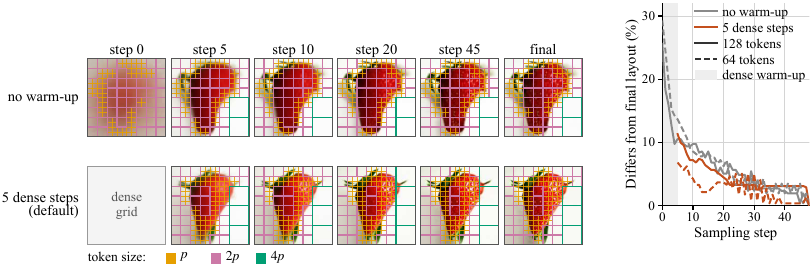}
\caption{\textbf{Layout evolution during sampling, JiT-B/32 at $512^2$.}}
\label{fig:dynamics}
\end{figure}

Eight ImageNet classes from fixed noise, 50 Heun steps at guidance 3.0 with the layout rebuilt at
every step, at 128 and 64 tokens, with no warm-up (the first layout is read off the pure-noise estimate) or with the
default five dense steps. Figure~\ref{fig:dynamics} shows one sample's estimate and layout at 128 tokens (left; top
without warm-up, bottom with it) and, over the eight images, the share of the image whose token size still differs from
the final layout (right; grey band: the default's warm-up).

\subsection{Pooled budget across a batch (final JiT-B/32)}
\label{app:pool}

\begin{figure}[H]
\centering
\includegraphics[width=\linewidth]{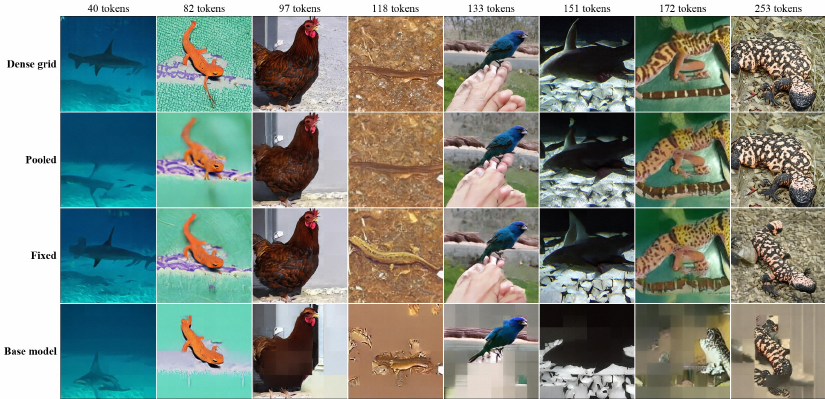}
\caption{\textbf{Pooled budget, JiT-B/32 at $512^2$, a mean of 128 tokens.}}
\label{fig:pool}
\end{figure}

Every reported result gives each image the same budget. The pooled arm instead sets one offset per
batch, so the batch meets the budget on average and each image takes what its content asks for (16 to 253 tokens at a
mean of 128); the fixed arm is the default. Table~\ref{tab:pool} gives FID-10k as in Table~\ref{tab:sampler}, and PSNR
(mean, with 10th / 90th percentiles) and LPIPS of 2,000 images against the dense image of the same noise, with the base
model at the same token count as a control. Figure~\ref{fig:pool} shows eight samples at a mean of 128 tokens, ordered by
the tokens the pool gave them (above each column): the dense image, the pooled and fixed arms, and the base model.

\begin{table}[H]
\centering\scriptsize\setlength{\tabcolsep}{1.5pt}
\caption{\textbf{Pooled against fixed budget, JiT-B/32 at $512^2$: FID-10k and paired fidelity.}}
\label{tab:pool}
\begin{tabular*}{\linewidth}{@{\extracolsep{\fill}}ll rrr@{}}
\toprule
& & 70\% & 50\% & 25\% \\
metric & arm & \textit{179} & \textit{128} & \textit{64} \\
\midrule
FID-10k $\downarrow$ & fixed (default) & 8.20 & 10.99 & \textbf{19.79} \\
 & pooled & \textbf{7.79} & \textbf{10.82} & 20.39 \\
\midrule
PSNR $\uparrow$ & fixed (default) & 24.57 (20.86 / 28.40) & 23.22 (19.83 / 26.84) & 22.14 (18.96 / 25.59) \\
 & pooled & 26.35 (22.05 / 29.54) & 23.49 (20.47 / 26.56) & 22.08 (19.18 / 25.27) \\
 & base model & 20.58 (17.44 / 24.17) & 18.68 (15.95 / 21.59) & 16.02 (13.75 / 18.52) \\
\midrule
LPIPS $\downarrow$ & fixed (default) & 0.308 & 0.379 & \textbf{0.450} \\
 & pooled & \textbf{0.271} & \textbf{0.370} & 0.457 \\
 & base model & 0.487 & 0.639 & 0.877 \\
\bottomrule
\end{tabular*}
\end{table}

\end{document}